\documentclass[10pt,twocolumn,letterpaper]{article}

\usepackage{iccv}
\usepackage{times}
\usepackage{epsfig}
\usepackage{graphicx}
\usepackage{amsmath}
\usepackage{amssymb}

\usepackage{booktabs}
\usepackage{multirow}
\usepackage{comment}
\usepackage{color}
\usepackage{bm}
\usepackage{algorithm}
\usepackage{algorithmic}
\usepackage{xr-hyper}
\usepackage{amssymb}
\usepackage{amsmath}
\usepackage{pifont}%
\usepackage{wrapfig}

\usepackage{enumerate}

\newcommand{\xmark}{\ding{55}}%

\usepackage{array}
\newcolumntype{H}{>{\setbox0=\hbox\bgroup}c<{\egroup}@{}}

\usepackage{subcaption}
\usepackage[pagebackref=true,breaklinks=true,letterpaper=true,colorlinks,bookmarks=false]{hyperref}

\usepackage[capitalize]{cleveref}
\crefname{section}{Sec.}{Secs.}
\Crefname{section}{Section}{Sections}
\Crefname{table}{Table}{Tables}
\crefname{table}{Tab.}{Tabs.}

\usepackage{authblk}

\iccvfinalcopy 

\def\iccvPaperID{6478} 

\ificcvfinal\fi

\begin{document}

\title{ROME: Robustifying Memory-Efficient NAS via Topology Disentanglement and Gradient  Accumulation}

\author[1,2]{Xiaoxing Wang$^{\dagger}$\thanks{Work done as an intern at Meituan. $^{\dagger}$ Equal contribution}}
\author[2]{Xiangxiang Chu$^{\dagger}$}
\author[1,2]{Yuda Fan$^{*}$}
\author[1]{Zhexi Zhang}
\author[2]{Bo Zhang}
\author[1]{Xiaokang Yang}
\author[1]{Junchi Yan}

\affil[1]{Shanghai Jiao Tong University}
\affil[2]{Meituan}


\maketitle
\ificcvfinal\thispagestyle{empty}\fi

\begin{abstract}
Albeit being a prevalent architecture searching approach, differentiable architecture search (DARTS) is largely hindered by its substantial memory cost since the entire supernet resides in the memory. This is where the single-path DARTS comes in, which only chooses a single-path submodel at each step. While being memory-friendly, it also comes with low computational costs. Nonetheless, we discover a critical issue of single-path DARTS that has \textbf{not} been primarily noticed. Namely, it also suffers from severe \emph{performance collapse} since too many parameter-free operations like skip connections are derived, just like DARTS does. In this paper, we propose a new algorithm called RObustifying Memory-Efficient NAS (ROME) to give a cure. First, we disentangle the topology search from the operation search to make searching and evaluation consistent. We then adopt Gumbel-Top2 reparameterization and gradient accumulation to robustify the unwieldy bi-level optimization. We verify ROME extensively across 15 benchmarks to demonstrate its effectiveness and robustness.
\end{abstract}


\section{Introduction}
Despite the fast development of neural architecture search (NAS) \cite{zoph2016neural} to aid network design in vision tasks like classification~\cite{tan2019efficientnet,chu2021darts,gibbsnas,nasflow}, object detection~\cite{ghiasi2019fpn,eautodet}, and segmentation~\cite{liu2019auto}, there has been an urging demand for faster searching algorithms. Early methods based on the evaluation of a huge number of candidate models \cite{zoph2016neural,tan2018mnasnet,howard2019searching} require unaffordable costs (typically 2k GPU days). In the light of weight-sharing mechanism introduced in SMASH \cite{brock2017smash}, a variety of low-cost approaches have emerged \cite{bender2018understanding,pham2018efficient,liu2018darts}. DARTS \cite{liu2018darts} has taken the dominance with a myriad of follow-up works \cite{wu2018fbnet,cai2018proxylessnas,xie2018snas,dong2019searching,chen2019progressive,zela2019understanding}. In this paper, we investigate a single-path based variation of DARTS, typically GDAS \cite{dong2019searching},  for its fast speed and low GPU memory.

Unlike many DARTS variants that have to perform all candidate operations, single-path methods~\cite{stamoulis2019single,dong2019searching,xie2018snas}, also termed as memory-efficient NAS\footnote{We interchangeably use the term `single-path' and `memory-efficient'.}, are developed to sample and activate only a subset of operations in the supernet. For differentiable search, Gumbel-Softmax reparameterization tricks~\cite{jang2016categorical,maddison2016concrete} are generally employed \cite{wu2018fbnet,dong2019searching,xie2018snas}. 

In this paper, we show that single-path methods also suffer from severe \emph{performance collapse} where many parameterless operations accumulate, akin to that of DARTS as broadly discussed in \cite{zela2020understanding,chen2020stabilizing,xie2018snas,chu2019fair,liang2019darts+}. 
We propose a robust algorithm called ROME to resolve this issue. 

\begin{figure*}[tb!]
\centering
\includegraphics[width=0.8\textwidth]{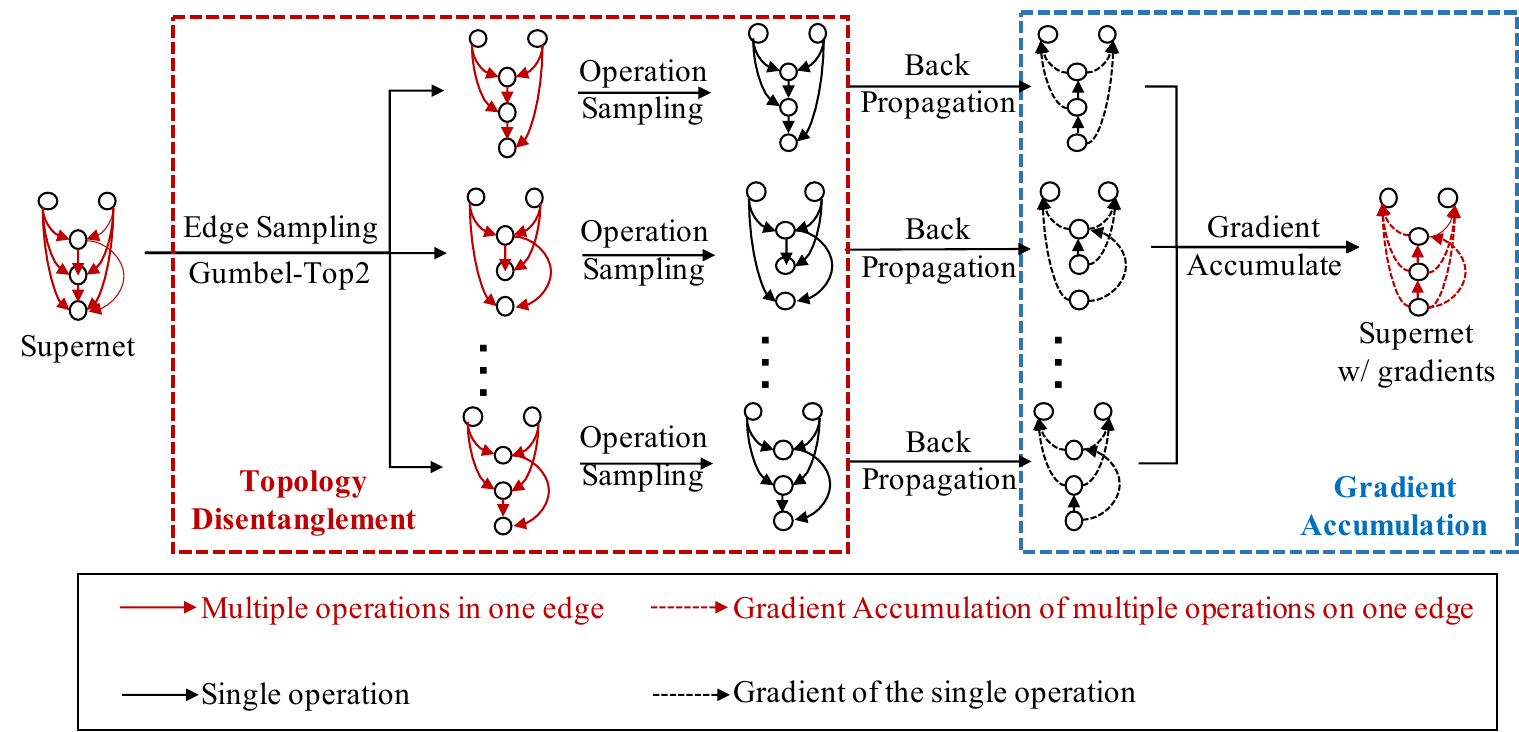}
\caption{ROME (v2): Gumbel-Top2 is devised to sample edges to satisfy the restriction that each node has in-degree 2. Subnetworks apply forward and backward independently. Gradients are accumulated to update the supernet weights at once.}
\label{fig:framework}
\vskip -0.15 in
\end{figure*}

We observe that single-path methods usually intertwine topology search with operation search, which creates inconsistency between searching and evaluation.  We first disentangle the two from each other. Specifically, in addition to the existing architectural parameters ($\bm{\alpha}$) that represent the significance of each operation, we involve topology parameters ($\bm{\beta}$) to denote the relative importance of edges. A single-path architecture can then be induced by both $\bm{\alpha}$ and $\bm{\beta}$. Moreover, to further robustify the searching process, we propose a \emph{gradient accumulation} strategy during the bi-level optimization.  We sketch the framework of ROME in Fig.~\ref{fig:framework}. In a nutshell, our contributions are,

\textbf{1) Revealing performance collapse in single-path differentiable NAS.} 
Similar to the performance collapse problem in DARTS, the architectures searched by single-path based methods can also be dominated by parameterless operations, especially skip connections. 
However, this issue hasn’t been deeply explored, which motivates us to propose a new robust, lower memory cost and search cost method.

\textbf{2) Consistent search and evaluation by disentangling topology search from operation selection.} 
We introduce independent topology parameters to unwind topology from operations, which avoids  structure inconsistency between the search and evaluation stage. We further devise Gumbel-Top2  re-parameterization to make our method differentiable and provide its theoretical validity. To our best knowledge, this is the first work to achieve consistent search and evaluation for single-path differentiable NAS. 

\textbf{3) Robustifying bi-level optimization via gradient accumulation.} We devise two gradient accumulation techniques to address the aforementioned issue. One helps fair training for each candidate operations. The other instead reduces the estimation variance on architectural weights.  

\textbf{4) Strong performance while maintaining low memory cost.}
Tested on the popular settings\footnote{Independently search under different random seeds instead of multiple training of a single best-searched model.} for NAS \cite{zela2020understanding,lindauer2020best,chu2021darts},
our approach achieves state-of-the-art on various search spaces and datasets across \textbf{15} benchmarks.  Our approach is fast, robust, and memory efficient. Compared with GDAS and PC-DARTS, our approach costs \textbf{26\%} and \textbf{38\%} lower memory in the standard search space of DARTS respectively. The source code will be made publicly available.

\section{Related Work}
\textbf{Differentiable Neural Architecture Search.}
Similar to \cite{zoph2017learning,pham2018efficient} that uses a directed acyclic graph to represent a cell, DARTS~\cite{liu2018darts} constructs a cell-based supernet and further introduces architectural weights to represent the importance of each operation. 
DARTS proposes two types (first-order and second-order) of approximation that alternatively update operation parameters and architectural weights with stochastic gradient descent. However, since the supernet subsumes all connections and operations within the search space, DARTS risks exhausting GPU memory.
A possible attempt to resolve this issue is done by progressively pruning operations
\cite{chen2019progressive}, which is still an indirect approach and requires strong regularization tricks. Apart from that, recent works~\cite{zela2020understanding,chen2020stabilizing} also point out an instability phenomenon of DARTS. These issues significantly restrict its application.

\textbf{Memory-efficient NAS.} 
To reduce GPU memory cost, several prior works have revised the forward procedure of the supernet. PC-DARTS~\cite{Xu2020PC-DARTS} makes use of partial connections instead of the full-fledged supernet. Some works~\cite{wang2020mergenas,eautodet,easynas} proposes to merge all parametric operations into one convolution, a similar super-kernel strategy is also used in single-path NAS \cite{stamoulis2019single}. 
ProxylessNAS~\cite{cai2018proxylessnas} samples two operations on each edge during the search process, which enables proxyless searching on large datasets. Single-path supernets like SPOS~\cite{guo2019single} and FairNAS~\cite{chu2021fairnas} sample only a single path at each iteration, however, both require an additional searching stage to choose the final models. Single-path differentiable methods like GDAS \cite{dong2019searching} sample a subgraph of the DAG at each iteration, which is by far the most efficient. However, we observe that a severe instability issue occurs, which has been previously neglected.

\textbf{Performance collapse of DARTS.} The collapse issue is one of the most critical problems in differentiable architecture search \cite{liang2019darts+,Xu2020PC-DARTS,chu2019fair,chen2020stabilizing,zela2020understanding}. It has been shown that  DARTS~\cite{liu2018darts} prefers to choose parameterless operations, leading to its performance collapse \cite{liang2019darts+,chu2019fair}. Recent works \cite{zela2020understanding,chen2020stabilizing} utilize the eigenvalue of the Hessian matrix as an indicator of collapse and design various techniques to regularize high eigenvalues. Instead, other works~\cite{liang2019darts+,chen2019progressive} directly constrain the number of skip connections as 2 to avoid collapse. Nonetheless, the previous methods are specifically designed for the full-path training scheme. Whereas the collapse problem in single-path has been rarely studied.

\section{Methodology}

\subsection{Collapse in Single-Path Differentiable NAS} \label{subsec:preliminary}
DARTS~\cite{liu2018darts} optimizes a supernet stacked by normal cells and reduction cells. A cell owns $N$ nodes $\{x_i\}_{i=1}^{N}$ denoting latent representation. Edge $e_{i,j}$ from node $x_i$ to $x_j$ integrates all candidate operations $\mathcal{O}$ whose importance is represented by architecture parameter $\alpha^{o}_{i,j}$.
Since the weights of all operations are involved in the forward and backward process, DARTS is very memory-consuming.

\begin{figure}[tb!]
\centering
\includegraphics[width=0.45\columnwidth]{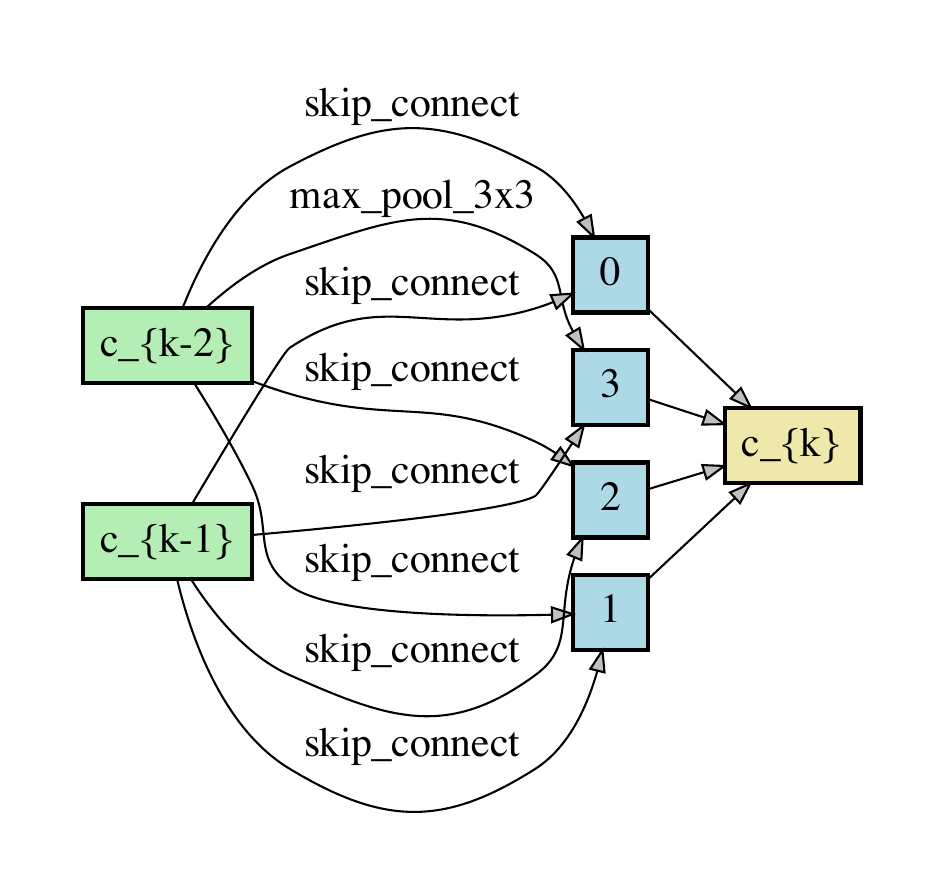}
\includegraphics[width=0.45\columnwidth]{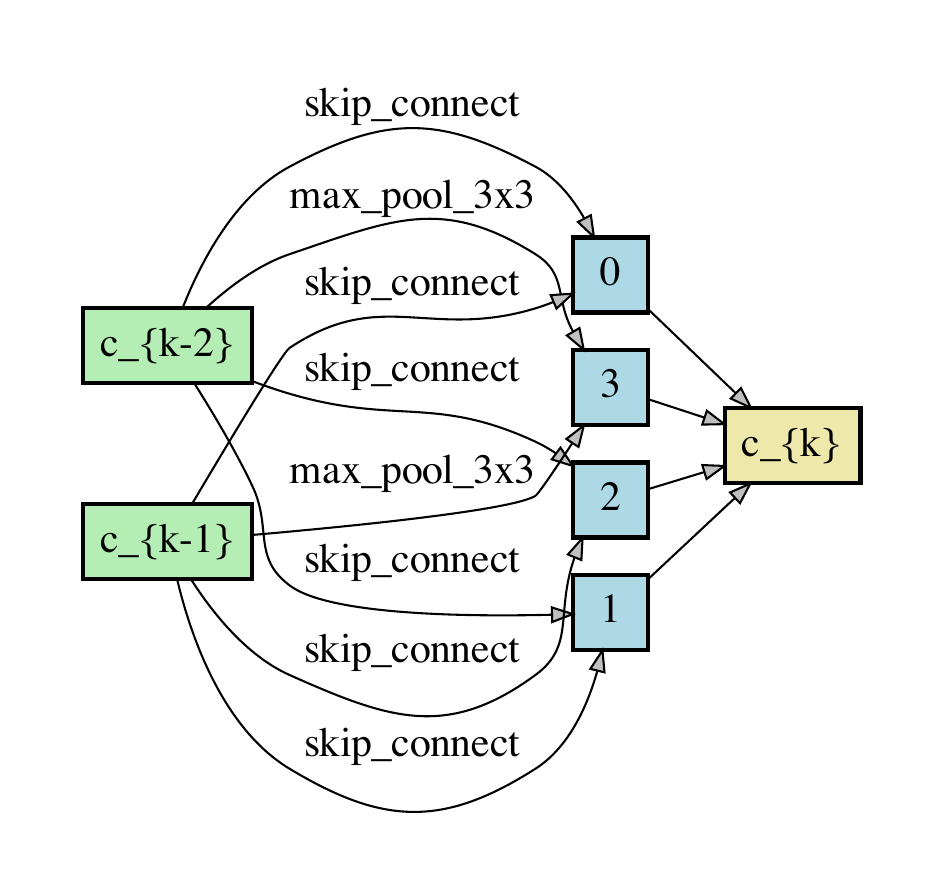}
\vspace{-10pt}
\caption{Two failure examples of GDAS \cite{dong2019searching} in the DARTS search space in our experiment by running the authors' code, where normal cells are full of parameterless operations. The average accuracy is 96.52\% on CIFAR-10, while models searched by our method can achieve  97.42\%.}
\label{fig:gdas-fail-geno}
\end{figure}

To reduce memory cost, GDAS~\cite{dong2019searching} proposes to sample a sub-set of operations at each iteration. For edge $e_{i,j}$, a one-hot random vector $z_{i,j} \in \{0,1\}^{|\mathcal{O}|}$ is sampled, indicating only one candidate operation is selected during the forward pass and back-propagation.

However, we observe that the normal cell learned by GDAS has 4 skip connections, and GDAS (FRC) even contains 5 skip connections, implying performance collapse issue also exists in single-path based methods.

\begin{figure}[tb!]
\centering
\includegraphics[width=0.98\columnwidth]{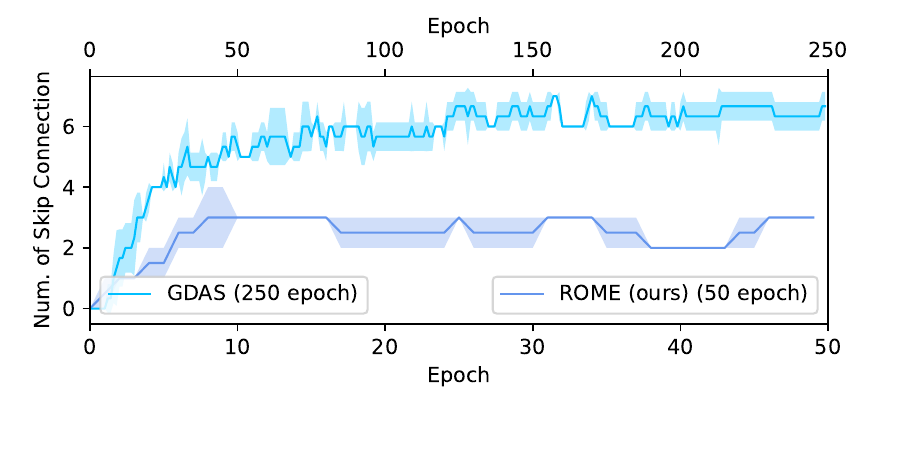}
\vspace{-15pt}
\caption{Evolution of the number of skip connections in a normal cell of GDAS vs. ROME. 
}
\label{fig:gdas-num-skip}
\end{figure}

We rerun the released code of GDAS~\cite{dong2019searching} for several times and observe that the normal cells are dominated by skip connections and max pooling, shown in Fig.~\ref{fig:gdas-fail-geno}. We also draw the evolution of skip connections of GDAS in Fig.~\ref{fig:gdas-num-skip}.

\subsection{Possible Reasons of Performance Collapse}\label{subsec:issue}
We conjecture that the following two factors contribute most to the collapse issue, which motivates us to provide a remedy in each regard.

\textbf{Inconsistency between the searching and evaluation stage.} 
Structural inconsistency between the supernet and the final network mainly appears at the operation level and the topology level. Operation-level inconsistency, \ie, searching with many operations but evaluating only with the most significant one, has been alleviated in recent single-path methods~\cite{dong2019searching,xie2018snas} by sampling one operation on each edge at each iteration in the search phase. However, topology-level inconsistency has long been neglected. Specifically, the nodes in the supernet connect with all its predecessors, while the nodes in the final network must only have in-degree 2. In this paper, we eliminate such inconsistency by disentangling topology and operation search. 

\textbf{Insufficient sampling for candidate operations.}
The instability issue for single-bath methods can be largely attributed to its stochastic nature that involves sub-sampling. Specifically, at each iteration, only one operation is sampled for each edge, resulting in an insufficient training of weights $\bm{\theta}$. It also causes high variance for the gradients of $\bm{\alpha}$, hence influencing the searching convergence. To this end, we propose \emph{multiple sampling} and \emph{gradient accumulation} to train the supernet and reduce the gradient variance of architectural weights.

Based on the above reasoning, we propose topology disentanglement (Sec~\ref{subsec:disentanglement}) to resolve inconsistency and gradient accumulation (Sec~\ref{subsec:gradient_accumulate}) to rectify the instability caused by insufficient sampling. Fig.~\ref{fig:gdas-num-skip} illustrate the number of skip connections in a normal cell searched by GDAS and our method (ROME), showing that our method can effectively alleviate the collapse issue.

\subsection{Topology Disentanglement} \label{subsec:disentanglement}

To disentangle the search for topology and operations on each edge, we use an indicator $\bm{B}_{i,j}\in\{0,1\}$ to denote whether edge $e_{i,j}$ is selected, and $\bm{A}_{i,j}^o\in\{0,1\}$ for whether operation $o$ on edge $e_{i,j}$ is selected.
Sampling architecture $z$ with $M$ connections can be decomposed into two parts: sample $M$ edges first, and their operations second.

\textbf{Sampling for edges.}
Topology inconsistency exists in single-path based methods~\cite{dong2019searching,xie2018snas}, as all 14 edges in a cell are selected in the search stage but the final architecture only has 8 edges. To address this issue, we propose to sample the same number of edges in search.

Each intermediate node should connect with exact two predecessors, satisfying DARTS's constraints. Formally, we use $\bm{B}_{i, j}$ to indicate whether the edge $e_{i, j}$ between node $x_i$ and $x_j$ is sampled, and we enforce,
\begin{gather}
    \label{eq:constraints_topology}
    \sum_{i < j} \bm{B}_{i, j} = 2, \quad \forall j.
\end{gather}

We give two techniques of edge sampling in Sec~\ref{subsec:gumbel-max-top2}.

\textbf{Sampling for operations.} 
We use $\bm{A}_{i, j}^o$ to denote whether the operator $o$ is sampled on the edge $e_{i, j}$, and we adopt Gumbel-Max technique to sample operations, where $\bm{g}_{i,j}^o$ is sampled from Gumbel$(0,1)$ distribution\footnote{$\bm{g}_{i,j}^o=-\log(-\log(\bm{\epsilon}_{i,j}^o))$, $\bm{\epsilon}_{i,j}^o$ obeys uniform distribution}, and $\tilde{\bm{\alpha}}_{i,j}^o =  \frac{\exp(\bm{\alpha}^{o}_{i,j})} {\sum_{o'\in\mathcal{O}} \exp(\bm{\alpha}^{o'}_{i,j})}$ is the normalized architectural weights:
\begin{align}
    \label{eq:prob_alpha}
    \bm{A}_{i, j} = \mathrm{one}\underline{\hbox to 0.2cm{}} \mathrm{hot} &\left[\arg\max_{o\in\mathcal{O}}(\log {\tilde{\bm{\alpha}}_{i,j}^{o}} + \bm{g}_{i,j}^{o})\right] \in \mathbb{R}^{|\mathcal{O}|},
\end{align}

To make the objective function differentiable to architectural weights $\bm{\alpha}$, we relax the discrete distribution to a continuous one by Gumbel-Softmax: 
\begin{gather}
    \tilde{\bm{A}}_{i,j}^{o} = \frac{\exp\left[(\log\tilde{\bm{\alpha}}_{i,j}^{o}+\bm{g}_{i,j}^{o})/\tau\right]}   {\sum_{o'=1}^{|\mathcal{O}|} \exp\left[(\log\tilde{\bm{\alpha}}_{i,j}^{o'}+\bm{g}_{i,j}^{o'})/\tau\right]}, \quad \nonumber \\
    \bm{A}_{i, j} = \mathrm{one}\underline{\hbox to 0.2cm{}} \mathrm{hot} \left[\arg\max_{o\in\mathcal{O}}{\tilde{\bm{A}}}_{i,j}^{o}\right], \label{eq:gumbelsoftmax}
\end{gather}
where the temperature $\tau$ gradually decreases in search.

\subsection{From Gumbel-Max to Gumbel-Top2 Reparameterization}\label{subsec:gumbel-max-top2}
We propose two variations of edge sampling techniques, \ie Gumbel-Max and Gumbel-Top2, based on which we derive two versions of ROME (v1 and v2).

\subsubsection{Gumbel-Max (ROME-v1).} 
Suppose node $x_j$ has $j$ possible predecessors, there are $\frac{j \times (j - 1)}{2}$ types of edge choices. 
We use $\bm{I}_j^{(i,k)} (i<k<j)$ to indicate whether node $x_j$ is connected both to $x_i$ and $x_k$, i.e. when $\bm{I}_j^{(i,k)}=1$, we have $\bm{B}_{i,j}=\bm{B}_{k,j}=1$ and $\bm{B}_{m,j}=0 (\forall m<j, m\neq i,j)$.

We then set a trainable variable $\bm{\beta}_j^{(i,k)}$ to denote the importance of each edge  choice for node $x_j$, such that
 \begin{gather}
     p\left(\bm{I}^{(i, k)}_j = 1\right) = \frac{\exp(\bm{\beta}^{(i, k)}_j)} {\sum_{i'< k'<j} \exp(\bm{\beta}^{(i', k')}_j)} \triangleq \tilde{\bm{\beta}}_j^{(i,k)}. 
 \end{gather}
 
We use Gumbel-Max technique where $\bm{g}^{(i,k)}_j$ obeys Gumbel (0,1) distribution,
\begin{gather} \label{eq:I}
    \bm{I}_j = \mathrm{one}\underline{\hbox to 0.2cm{}} \mathrm{hot} \left[\arg\max_{i<k<j}(\log {\tilde{\bm{\beta}}^{(i, k)}_j} + \bm{g}^{(i, k)}_j)\right] 
    \in \mathbb{R}^{\frac{j \times (j - 1)}{2}}.
\end{gather}
Take a cell as a whole, if edge $e_{i, j}$ is sampled, there must be another chosen $e_{k, j}$.
Thus $\textbf{B}_{i,j}$ can be formulated by $\textbf{I}_j$:
\begin{equation}
    \label{eq:v1_sample}
    \bm{B}_{i,j} = \sum_{k<i} \bm{I}_j^{(k,i)} + \sum_{k>i} \bm{I}_j^{(i,k)}.
\end{equation}

Gumbel-Softmax reparameterization is used to retain gradient information,
\begin{gather} \label{eq:I_gumbelsoftmax}
    \tilde{\bm{I}}_j^{(i,k)} = \frac{\exp\left\{\left[\log {\tilde{\bm{\beta}}^{(i, k)}_j} + \bm{g}^{(i, k)}_j\right]/\tau\right\}}{\sum_{s<t<j} \exp\left\{\left[\log {\tilde{\bm{\beta}}^{(s, t)}_j} + \bm{g}^{(s, t)}_j\right]/\tau\right\}}, \quad \nonumber \\
    \bm{I}_j = \mathrm{one}\underline{\hbox to 0.2cm{}} \mathrm{hot} \left[\arg\max_{i<k<j}\tilde{\bm{I}}^{(i, k)}_j\right]. \nonumber
\end{gather}

\subsubsection{Gumbel-Top2 (ROME-v2).} Enumerating all possible edge combinations as in ROME-v1 is straightforward but superfluous, hence in ROME-v2 we directly sample two edges per node. We define the probability of each edge $e_{i,j}$ as $p(e_{i,j})$. Given edge importance is denoted by $\bm{\beta}$, the sampling probability $p(e_{i,j}) = \frac{\exp(\bm{\beta}_{i,j})}{\sum_{k<j}\exp(\bm{\beta}_{k,j})} \triangleq \tilde{\bm{\beta}}_{i,j}$.

To satisfy the constraints on the cell topology (Eq.~\ref{eq:constraints_topology}), ROME-v2 extends Gumbel-Max to Gumbel-Top2:
\begin{align}\label{eq:pq}
    \tilde{\bm{B}}_{i,j} &= \frac{\exp\left((\log\tilde{\bm{\beta}}_{i,j}+\bm{g}_{i,j})/\tau\right)}   {\sum_{i'<j} \exp\left((\log\tilde{\bm{\beta}}_{i',j}+\bm{g}_{i',j})/\tau\right)}, \quad \\
    \bm{B}_{i,j}&=\left\{
\begin{aligned}
1, &\quad i \in \arg\mathop{\mathrm{top2}}\limits_{i'<j} (\tilde{\bm{B}}_{i',j}) \\
0, &\quad otherwise
\end{aligned}
\right.
\end{align}

We also demonstrate that the Gumbel-Top2 technique is equivalent to sampling two different edges \emph{without replacement} with probability simplex $p_i$, so that Gumbel-Top2 sampling can be made differentiable. Details can be found in Sec.~\ref{supp:sec:proof-gumbel} in the supplementary.

\subsection{Gradient Accumulation}  \label{subsec:gradient_accumulate}
Lastly, we tackle the issues caused by insufficient sampling. Suppose architectural weights $\bm{\alpha}$ and $\bm{\beta}$ be the parameters of a distribution for architectures. A candidate architecture $z$ is obtained by independently sampling edges and operations. 
Suppose $z$ owns $M$ edges $\{e_1,...,e_M\}$ and the corresponding operations $\{o_1,...,o_M\}$, then the probability of $z$ being selected is given by
\begin{equation}
	p(z;\bm{\alpha}, \bm{\beta}) = \prod_{i=1}^M p(e_i;\bm{\beta})\times p(o_i|e_i;\bm{\alpha}).
\end{equation}

The search process can be thus modeled as finding optimal $\bm{\alpha}$ and $\bm{\beta}$ to minimize the expectation of validation loss of the architectures as Eq.~\ref{eq:opt_operation}, where $\bm{\theta}_z^*$ denotes the optimal operation parameters for the sampled architecture $z$.
\begin{equation}
	\begin{split}
		\min_{\bm{\alpha}, \bm{\beta}} \quad \mathbb{E}_{z\sim p(z;\bm{\alpha}, \bm{\beta})}\left[L_{val}(\bm{\theta}_z^*, z) \right], \qquad \\
		\mathrm{s.t.} \quad  \bm{\theta}_z^* = \arg\min_{\bm{\theta}} L_{train}(\bm{\theta}, z) \label{eq:opt_operation}
	\end{split}
\end{equation}

However, architecture $z$ changes at each iteration while the corresponding operation weights $\bm{\omega}$ are  updated only once. Apparently, single-path approaches suffer from two problems: \emph{unfair} and \emph{biased training} for candidate operations, and creating \emph{large variance} for architectural weights.

We propose two effective techniques based on \emph{gradient accumulation}. 
\textbf{First}, to boost fair training for operations, we sample $K$ sub-models from the supernet and accumulate gradients for each sub-model within one iteration. Weights $\bm{\omega}$ are updated by the accumulation of gradients from $K$ sub-models.
\textbf{Second}, to reduce the variance for architectural weights,
we sample another $K$ sub-models and average the gradients of architectural weights.
Suppose that the gradient of $\alpha$ be a random variable whose variance is $\sigma^2$, then averaging among $K$ samples reduces the variance to $\sigma^2/K$. Specifically, we alternately update operation parameters $\bm{\theta}$ and architectural 
 parameters $\bm{\alpha}$ (similar for $\bm{\beta}$) as:
\begin{align}
\bm{\alpha} \leftarrow \bm{\alpha} - \frac{1}{K}\sum_{k=1}^K \nabla_{\bm{\alpha}} L_{val}(\bm{\theta}, z_k),\quad \label{eq:update_alpha} \\
\bm{\theta} \leftarrow \bm{\theta} - \sum_{k=1}^K \nabla_{\bm{\theta}} L_{train}(\bm{\theta}, z'_k) \label{eq:update_theta},
\end{align}
where $z_k,z'_k$ denote the sampled architectures. We can now summarize our ROME method wholely in Alg.~\ref{alg:romeNAS}.

\begin{algorithm}[tb!]
\caption{ROME (with two versions v1 and v2).}
\label{alg:romeNAS}
\textbf{Input}: iteration count $T$; number of sampling $K$; initialized operation parameters $\bm{\theta}$; and architectural weights $\bm{\alpha}$, $\bm{\beta}$;\\
\textbf{Output}: optimal architecture $z^*$;

\begin{algorithmic}[1] 
\FOR {$t = 1 \to T$}
    \STATE Sample two batches of data samples $D_s$ and $D_t$ from two disjoint datasets; \\
    \FOR {$k = 1 \to K$} 
        \STATE Topology sampling by Eq.~\ref{eq:v1_sample} (v1) or  Eq.~\ref{eq:pq} (v2); Operation sampling by Eq.~\ref{eq:gumbelsoftmax}; \\
        \STATE Get sampled architecture $z_k$;\\
    \ENDFOR
    \STATE Gradient accumulation and update $\bm{\alpha}, \bm{\beta}$ by Eq.~\ref{eq:update_alpha} on $D_s$, where $L_{val}$ is cross entropy; \\
    \FOR {$k = 1 \to K$} 
        \STATE Topology sampling by Eq.~\ref{eq:v1_sample} (v1) or  Eq.~\ref{eq:pq} (v2); Operation sampling by Eq.~\ref{eq:gumbelsoftmax};\\
        \STATE Get sampled architecture $z'_k$;\\
    \ENDFOR
    \STATE Gradient accumulation and update $\bm{\theta}$ by Eq.~\ref{eq:update_theta} on $D_t$, where $L_{train}$ is cross entropy;
\ENDFOR
\STATE \textbf{return}: $z^*=\arg\max\limits_z p(z;\bm{\alpha}, \bm{\beta})$
\end{algorithmic}
\end{algorithm}

\section{Experiments}\label{sec:experiments}
\subsection{Protocols} \label{subsec:protocol}
\textbf{Search spaces.}
In this paper, we denote DARTS's search space as S0, which comprises a stack of duplicate normal cells and reduction cells. Each cell is represented by a DAG with 4 intermediate nodes. Candidate operations between each two nodes are \{maxpool, avgpool, skip\_connect, sep\_conv 3$\times$3 and 5$\times$5, dil\_conv 3$\times$3 and 5$\times$5\}. We exclude \{none\} operation from the default DARTS search space to satisfy the topology constraint in Eq. \ref{eq:constraints_topology}.
Under S0 space, we search and evaluate on CIFAR-10~\cite{krizhevsky2009learning} and ImageNet~\cite{deng2009imagenet} respectively.

We also conduct experiments on four reduced search spaces, S1-S4, introduced by R-DARTS~\cite{zela2020understanding} to evaluate the stability of our method.
S1 is a pre-optimized search space, where each edge in the supernet has a predefined set of candidate operations. 
In the other 3 search spaces, candidate operations on each edge are the same (see the details in the supplementary).
Following~\cite{zela2020understanding}, we  search and evaluate in these 4 search spaces on CIFAR-10, CIFAR-100~\cite{krizhevsky2009learning}, and SVHN~\cite{netzer2011reading} (12 benchmarks in total).

\textbf{Search settings.}
Similar to DARTS, the supernet consists of 8 cells with 16 initial channels.
We search for $50$ epochs and set the sampling number $K=7$. Unless explicitly written, ROME-v2 is used by default throughout the paper since it is more efficient and robust. For operation parameters, we use the SGD optimizer with a momentum of 0.9 and an initial $lr$ of $0.05$; For architectural weights, we adopt the Adam optimizer with an initial $lr$ of $3\times 10^{-4}$.

\textbf{Evaluation settings.}
We use standard evaluation settings as DARTS \cite{liu2018darts} by training the inferred model for 600 epochs using SGD with a batch size of 96. 
For search space S0, inferred models are constructed by stacking 20 cells with 36 initial channels, and trained under the same settings following~\cite{chen2019progressive,liu2018darts}.
For S1-S4, we strictly follow the settings in R-DARTS \cite{zela2020understanding} for a fair comparison.

\begin{table*}[tb]
    \small
	\centering
	\setlength{\tabcolsep}{2.0pt}
	\begin{tabular}{cccccccHHcc}
		\toprule
		\multicolumn{2}{c}{\textbf{Benchmark}}  &  \textbf{DARTS}$^\dagger$ & 	\textbf{ES}$^\dagger$  & \textbf{ADA}$^\dagger$  & \multicolumn{2}{c}{\textbf{GDAS}} &  \textbf{PC-DARTS}  &  \textbf{ROME-v1} & \multicolumn{2}{c}{\textbf{ROME}} \\
		\cmidrule(lr){3-3} \cmidrule(lr){4-4} \cmidrule(lr){5-5} \cmidrule(lr){6-7} \cmidrule(lr){10-11}
		\multicolumn{2}{c}{} & Error (\%) & Error (\%) & Error (\%) & Error (\%) & Num & & & Error(\%) & Num \\
		\midrule
		\multirow{4}*{C10} & S1 & 4.66$\pm$0.71 & 3.05$\pm$0.07 & 3.03$\pm$0.08 & 2.89$\pm$0.09 & 3.8$\pm$0.4  &  &  2.93$\pm$0.09 & \textbf{2.66$\pm$0.06} & 1.3$\pm$0.4 \\
		~ &  S2 &   4.42$\pm$0.40 & 3.41$\pm$0.14 & 3.59$\pm$0.31 & 3.89$\pm$0.17 & 6.0$\pm$0.7 &  $\pm$ & \textbf{3.34$\pm$0.12}  & \textbf{3.14$\pm$0.14} & 2.0$\pm$0.0 \\
		~    &  S3 &   4.12$\pm$0.85 & 3.71$\pm$1.14 & 2.99$\pm$0.34 & 3.04$\pm$0.10 & 6.5$\pm$0.5 &  $\pm$ & \textbf{2.72$\pm$0.09} & \textbf{2.61$\pm$0.00} & 2.0$\pm$0.0 \\ 
		~ & S4 & 6.95$\pm$0.18 & 4.17$\pm$0.21 & 3.89$\pm$0.67 & 3.34$\pm$0.10 & 0.0$\pm$0.0 & $\pm$ & \textbf{\underline{3.21$\pm$0.00}} & \textbf{3.21$\pm$0.12} & 0.0$\pm$0.0 \\
		\midrule
		\multirow{4}*{C100} & S1 & 29.93$\pm$0.41 & 28.90$\pm$0.81 & 24.94$\pm$0.81 & 24.49$\pm$0.08 & 4.0$\pm$0.0 & $\pm$ & \textbf{\underline{22.65$\pm$0.45}} & \textbf{22.71$\pm$0.71} & 2.3$\pm$0.4 \\
		~ &  S2 &    28.75$\pm$0.92 & 24.68$\pm$1.43 & 26.88$\pm$1.11 & 24.57$\pm$0.47 & 6.3$\pm$0.4 &  $\pm$  & \textbf{23.14$\pm$0.98}   & \textbf{22.91$\pm$0.75} & 3.5$\pm$0.5 \\ 
		~    &  S3  &  29.01$\pm$0.24 & 26.99$\pm$1.79 & 24.55$\pm$0.63 & 22.86$\pm$0.17 & 3.0$\pm$0.7 & $\pm$ & 23.03$\pm$0.66 & \textbf{22.43$\pm$0.36} & 2.5$\pm$0.5 \\ 
		~ & S4 & 24.77$\pm$1.51 & 23.90$\pm$2.01 & 23.66$\pm$0.90 & 24.14$\pm$0.89 & 2.3$\pm$1.1 & $\pm$ & \textbf{21.33$\pm$0.00} & \textbf{20.95$\pm$0.45} & 0.0$\pm$0.0 \\
		\midrule
		\multirow{4}*{SVHN}&  S1 & 9.88$\pm$5.50 & 2.80$\pm$0.09 & 2.59$\pm$0.07 & 2.48$\pm$0.04 & 2.8$\pm$0.4 & $\pm$ & \textbf{2.37$\pm$0.04} & \textbf{2.34$\pm$0.06} & 0.8$\pm$0.4 \\
		~ & S2 & 3.69$\pm$0.12 & 2.68$\pm$0.18 & 2.79$\pm$0.22 & 3.05$\pm$0.02 & 7.8$\pm$0.4 & $\pm$& \textbf{2.49$\pm$0.14} & \textbf{2.41$\pm$0.07} & 1.0$\pm$0.0 \\
		~ & S3 & 4.00$\pm$1.01 & 2.78$\pm$0.29 & 2.58$\pm$0.07 & 3.62$\pm$0.36 & 7.5$\pm$0.5 & $\pm$ & \textbf{2.61$\pm$0.03} & \textbf{2.56$\pm$0.03} & 1.5$\pm$0.5 \\
		~ & S4 & 2.90$\pm$0.02 & 2.55$\pm$0.15 & 2.52$\pm$0.06 & 2.51$\pm$0.06 & 1.5$\pm$1.5 & $\pm$ & \textbf{2.43$\pm$0.00} &  \textbf{2.34$\pm$0.00} & 0.0$\pm$0.0 \\
		\bottomrule
	\end{tabular}
	\caption{Comparison in 4 search spaces and 3 datasets. For each algorithm, we independently search for 3 times under the settings in R-DARTS~\cite{zela2020understanding} and report the averaged performance. `EA' and `ADA' are two methods proposed by R-DARTS.
	`Num': To reveal the collapse issue, we also report the average number of parameterless operations in the discovered normal cell for GDAS and ROME
	$\dagger$: Results are obtained from R-DARTS.}
	\label{tab:comparison-rdarts-s1-4-avg}
\end{table*}

\subsection{Robustness Evaluation} \label{subsec:experiment_robust}
For comprehensive evaluation, we follow the recommended best practices for NAS by \cite{lindauer2020best,zela2020understanding,chen2020stabilizing,chu2021darts,chu2019fair} to report the mean and variance across several times of parallel \emph{searching with various random seeds}, through which the robustness of a method can be measured.  We appeal to the community for avoiding a common mistake that trains a single best-searched model several times, which only tests the convergence stability of a single model.

\begin{figure}[tb!]
	\centering
	 \begin{subfigure}{0.45\columnwidth}
    \includegraphics[width=0.98\columnwidth]{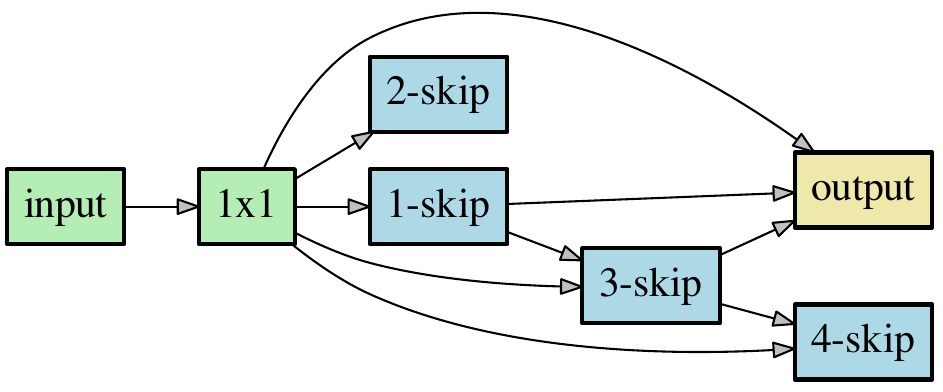}
    \caption{GDAS \cite{dong2019searching}}
    \label{fig:nas-bench-101-gdas}
  \end{subfigure}
    \hfill
    \begin{subfigure}{0.49\columnwidth}
		\includegraphics[width=0.98\columnwidth]{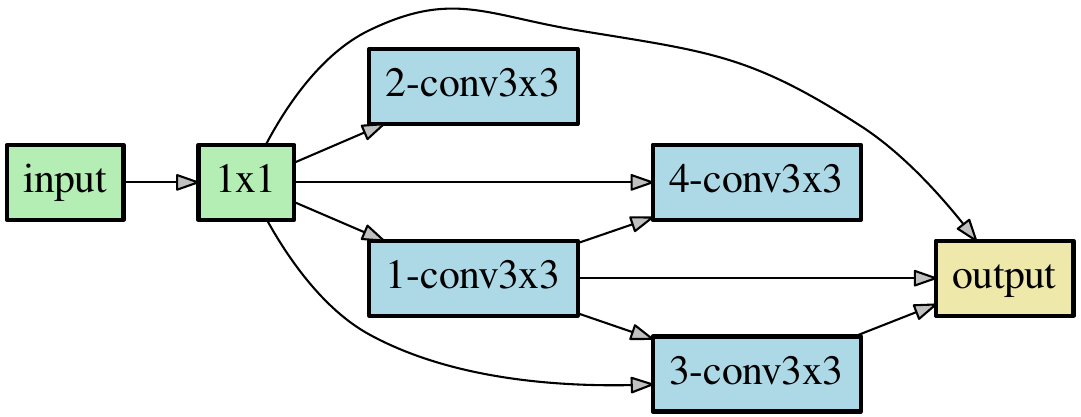}
    \caption{ROME}
    \label{fig:nas-bench-101-rome}
  \end{subfigure}
	\vspace{-10pt}
	\caption{GDAS fails on NAS-Bench-1Shot1~\cite{zela2020nas} on CIFAR-10 when adding skip connection to the second search space. Notice that nodes with no out-degrees have no contribution to the output.
	}
	\label{fig:1shot1-gdas-fail}
\end{figure}

\textbf{Discussion on collapse behavior across popular NAS benchmarks.} We argue that excluding an important operation for search space can cause illusive conclusions. Specifically, NAS-Bench-1Shot1 \cite{zela2020nas} suggests that Gumbel-based NAS is quite robust. However, this observation is laying on the basis that popular skip connections are not included in the search space \cite{pmlr-v97-ying19a}. After adding skip connection into the choices, we perform GDAS using their released code, whose best model found is full of skip connections, which again supports our discovery of collapse issue in single-path based NAS, see Fig.~\ref{fig:1shot1-gdas-fail}. In contrast, ROME does not suffer the same issue in these search spaces\footnote{More results see Fig.~\ref{fig:1shot1-gdas-fail-rest} and Fig.~\ref{fig:1shot1-rome-best} in the supplementary.}.

\textbf{Robustness evaluation on 12 hard benchmarks}.
We follow R-DARTS~\cite{zela2020understanding} to evaluate the performance and across 3 datasets in S1-S4 search spaces, see Table~\ref{tab:comparison-rdarts-s1-4-avg}. Our methods robustly outperform other methods with a clear margin across all these benchmarks.

Additionally, we observe that parameterless operations in GDAS dominate the normal cell in both S2 and S3, while our method effectively handles their numbers and thus stabilizes the searching  stage. The best single models' comparison is reported in Table~\ref{supp:tab:comparison-rdarts-s1-4-best} (supplementary). 

\begin{table}[ht]
    \small
	\centering
	\setlength{\tabcolsep}{1pt}
	\begin{tabular}{l*{4}{c}c} 	
		\toprule	
		\textbf{Models}   &  \textbf{Params}  &  \textbf{FLOPs}  &  \textbf{Error}  &  \textbf{Cost}& SP\\
		& {(M)}  & {(M)}  & {(\%)} & {GPU Days} & \\
		\midrule
		
		DARTS-V1 \cite{Yu2020Evaluating} &-&-&3.38$\pm$0.23&0.4&\xmark\\
		P-DARTS \cite{chen2019progressive}$^\ddagger$   &  3.3$\pm$0.2  &  540$\pm$34  & 2.81$\pm$0.14  &  0.3 &\xmark\\
		PC-DARTS \cite{Xu2020PC-DARTS}$^\ddagger$  &  3.6$\pm$0.5  &  592$\pm$90  &  2.89$\pm$0.22  &  0.1 &\xmark\\ 
		PR-DARTS \cite{zhou2020NAS}$^\ddagger$  & 3.4 & - & 2.68$\pm$0.10 & 0.2 &\xmark\\
		ISTA-NAS \cite{yang2020ista} $^\ddagger$  & 3.3 & - & 2.71$\pm$0.10 & 0.05 &\xmark\\
		R-DARTS \cite{zela2020understanding}  & - &  - &  2.95$\pm$0.21 & 1.6&\xmark\\
		SDARTS-ADV \cite{chen2020stabilizing}  & 3.3  & -  &  2.61$\pm$0.02  & 1.3&\xmark\\
            DARTS-PT~\cite{darts-PT} & 3.0 & - & 2.61$\pm$0.08 & 0.8 & \xmark\\
            NASI-FIX~\cite{nasi} & 3.9 & - & 2.79$\pm$0.07 & 0.24 & \xmark\\
            ZARTS~\cite{zarts} & 3.7 & - & 2.54$\pm$0.07 & 1.0 & \xmark\\
        Few-shot NAS~\cite{fewshotnas} & 3.8 & - & 2.31$\pm$0.08 & 1.35 & \xmark\\
		
		GDAS \cite{dong2019searching}  & 3.4  & -  &  2.93  & 0.2 &\checkmark\\
		SNAS \cite{xie2018snas} & 2.8  & -  &  2.85$\pm$0.02  & 1.5 & \checkmark\\
	    \midrule	
	    ROME-v1 (best) &4.5 & 683 & 2.53& 0.3&\checkmark\\
		ROME-v1 (avg.) & 4.0$\pm$0.6 & 670$\pm$21 & 2.63$\pm$0.09 & 0.3&\checkmark\\
		\textbf{ROME-v2} (best) & 3.6& 591 &2.48&0.3&\checkmark\\
		\textbf{ROME-v2} (avg.) & 3.7$\pm$0.2 & 595$\pm$28&2.58$\pm$0.07&0.3&\checkmark\\
		
		\bottomrule
	\end{tabular}
	\caption{Averaged performance among 4 independent runs of search on CIFAR-10. $^\ddagger$: reproduced result using their released code since they didn't report the average performance. $^\dagger$: FLOPs are calculated by their released architecture. SP: single-path based method
	}
	\label{tab:cifar10-s0}
\end{table}

\begin{table}[tb!]
	\centering
	\begin{tabular}{*{2}{l}H*{2}{l}} 	
		\toprule	
		\textbf{Models}   &  \textbf{Params}  &  \textbf{FLOPs} &  \textbf{Error } &  \textbf{Cost}  \\
		& \scriptsize{(M)}  & \scriptsize{(M)}  & \scriptsize{(\%)} & \scriptsize{GPU Days}   \\
		\midrule
		ResNet \cite{he2016deep}   &  1.7 &     &   22.10$^\diamond$  &  -  \\
		AmoebaNet \cite{real2019regularized} & 3.1 & &  18.93$^\diamond$ & 3150  \\
		PNAS \cite{liu2018progressive} & 3.2 & & 19.53$^\diamond$ & 150 \\
		ENAS \cite{pham2018efficient}  &  4.6  &    &  19.43$^\diamond$  & 0.45    \\
		DARTS \cite{liu2018darts}  & - & & 20.58$\pm$0.44$^\star$ &  0.4  \\ 
		GDAS \cite{dong2019searching}  &  3.4  &    &  18.38  & 0.2 \\
		P-DARTS \cite{chen2019progressive}  &  3.6  &   &  17.49$^\ddagger$  &  0.3 \\
		R-DARTS \cite{zela2020understanding}  & - &  - &  18.01$\pm$0.26 & 1.6 \\
            NASI-FIX~\cite{nasi} & 4.0 & - & 16.12$\pm$0.38 & 0.24 \\
            ZARTS~\cite{zarts} & 4.1 & - & 16.29$\pm$0.53 & 1.0 \\

		\midrule
		ROME-V1 (avg.) &  4.4$\pm$0.2 & 602$\pm$28 & 17.41$\pm$0.12 & 0.3\\
		ROME-V1 (best) &  4.4 & 603 & 17.33 & 0.3\\
		ROME-V2 (avg.) & 3.4$\pm$0.3 &  513$\pm$1.3  &  17.71$\pm$0.11 & 0.3 \\
		ROME-V2 (best) &  3.3 & 511   & 17.57  & 0.3\\
		\bottomrule
	\end{tabular}
 	\caption{Comparison of searched models on CIFAR-100. $^\diamond$: Reported by \cite{dong2019searching}, $^\star$: Reported by \cite{zela2020understanding}, $^\ddagger$:Rerun their code. }
	\label{tab:comparison-cifar100}
\end{table}

\subsection{Performance Comparison} \label{subsec:experiment_compare}

\textbf{Performance in S0 on CIFAR.}
We follow DARTS \cite{liu2018darts} and search on the CIFAR-10. Table~\ref{tab:cifar10-s0} shows that ROME achieves state-of-the-art performance with only 0.3 GPU-days\footnote{GDAS searches 250 epochs and costs 0.2 GPU-days. ROME searches 50 epochs with $K=7$, equivalent to searching 350 epochs by GDAS.}.  ROME has an average of 2.58$\pm$0.07\% error rate, which is slightly higher than up-to-date SOTAs such as SDARTS-ADV~\cite{chen2020stabilizing}. However, ROME is more than 4$\times$ faster. Compared with R-DARTS \cite{zela2020understanding}, ROME robustly outperforms it with 5$\times$ fewer search costs. Our best model achieves $97.52\%$ accuracy with $3.6$M parameters (see the architecture in Fig.~\ref{fig:rome-v2-cifar10-best-geno} supplementary).

We also search on CIFAR-100 and show the results in Table~\ref{tab:comparison-cifar100}. ROME surpasses all the methods and achieves state-of-the-art with only 0.3 GPU-days search cost.

\textbf{Performance in S0 on ImageNet.}
First, we transfer the architecture searched on CIFAR-10 to ImageNet following the common practice \cite{liu2018darts,chen2019progressive,li2019sgas,chu2019fair}. Same as \cite{li2019sgas,chu2019fair}, we train models for 250 epochs with a batch size of 1024 by SGD optimizer with a momentum of 0.9 and an initial $lr$ of 0.5  base learning rate. We also utilize an auxiliary classifier strategy. 
The results are shown in Table~\ref{tab:comparison-imagenet}, where ROME achieves 75.3$\% $ top-1 accuracy.

Second, as ROME features low memory cost and great robustness, we directly search on ImageNet as well.
We randomly sample 10\% images to train operation parameters and another 10\% to train architectural weights. A supernet is constructed by stacking 8 cells with 16 initial channels. We search for 30 epochs with $K=3$. The batch size is set as 256. Our search cost is reduced to 0.4 GPU days on a single Tesla V100. We fully train the discovered model for 250 epochs with the same evaluation settings as above.
Results are illustrated in Tabel~\ref{tab:comparison-imagenet}, showing that ROME achieves 75.5$\%$ top-1 accuracy. 
To make fair comparisons, we reproduce GDAS~\cite{dong2019searching} under the same settings (90 epochs). However, the network is dominated by skip connections and only achieves $72.5\%$ top-1 accuracy.
The structure of cells searched by GDAS and ROME are shown in the supplementary (Fig.~\ref{fig:compare-s0-imagenet}, Fig.~\ref{fig:gdas-s0-imagenet} and Fig.~\ref{fig:rome-s0-imagenet}).

\begin{table}[tb!]
    \centering
	\setlength{\tabcolsep}{2pt}
	\begin{tabular}{lccccH} 			
	\toprule
	\textbf{Models} & \textbf{FLOPs}  & \textbf{Params} & \textbf{Top-1} & \textbf{Cost} & Way\\
	& (M) & {(M)} & {(\%)}  & {(GPU days)} & \\
	\midrule

	AmoebaNet-A \cite{real2019regularized} & 555  & 5.1 & 74.5& 3150&TF CIFAR10 \\


	NASNet-A \cite{zoph2017learning}  & 564 & 5.3 &74.0 & 2000&TF CIFAR10\\
	PNAS \cite{liu2018progressive} & 588 & 5.1 & 74.2  &225&TF CIFAR10\\ 


	DARTS \cite{liu2018darts} & 574 & 4.7 & 73.3 & 0.4&TF CIFAR10\\
	P-DARTS \cite{chen2019progressive}& 577 & 5.1 & 75.3 & 0.3& TF CIFAR100\\
	FairDARTS-B \cite{chu2019fair}& 541 & 4.8 &75.1 & 0.4&TF CIFAR10\\
	SNAS \cite{xie2018snas}  &522&4.3&72.7 & 1.5&TF CIFAR10\\


	PC-DARTS  \cite{Xu2020PC-DARTS} & 586 & 5.3 & 74.9 & 0.1&TF CIFAR10\\ 
	GDAS \cite{dong2019searching}  &581&5.3&74.0  & 0.2&TF CIFAR10 \\
	\textbf{ROME (ours)} &576&5.2&75.3&0.3&TF CIFAR10\\
	\midrule
    DARTS, P-DARTS &\multicolumn{4}{c}{OOM when batch-size $\geq$ 32}& DS ImageNet  \\
    FairNAS-C \cite{chu2021fairnas} &321 & 4.4 & 74.7 &12&DS ImageNet\\
    ProxylessNAS \cite{cai2018proxylessnas}  & 465 & 7.1  & 75.1 & 8.3&DS ImageNet\\
    FBNet-C \cite{wu2018fbnet}   & 375 & 5.5 &  74.9 & 9&DS ImageNet\\ 
    PC-DARTS \cite{Xu2020PC-DARTS}$^\ddagger$  & 597 & 5.3 & 75.4 & 3.8&DS ImageNet\\
	GDAS \cite{dong2019searching}$^\ddagger$&405&3.6&72.5&0.8&DS ImageNet\\
	\textbf{ROME (ours)} & 556 & 5.1 & 75.5 & 0.5&DS ImageNet\\
	\bottomrule
\end{tabular}
\caption{Performance on ImageNet. The first block indicates the models \emph{transferred} from CIFAR-10; The second block indicates that the models are  \emph{directly searched} on ImageNet.
$^\ddagger$: reproduced using their released code.}
\label{tab:comparison-imagenet}
\vspace{-5pt}
\end{table}

\section{Ablation Study}

\textbf{Sensitivity to the Sampling Number $K$.} 
$K$ is a hyper-parameter in our gradient accumulation strategy, which is designed to reduce the variance of noise on $\nabla_{\bm{\alpha}}L_{val}$ and stabilizes the search as analyzed in Sec.~\ref{subsec:gradient_accumulate}. 

\begin{table}[tb]
	\begin{center}
		\begin{tabular}{lccc}
			\toprule
			$\bm{K}$ & \textbf{Acc} (\%) & \textbf{\# Params} &  \textbf{\# Epochs} \\
			\midrule
			1 & 97.12$\pm$0.06 & 3.34M & 350 \\
			4 & 97.28$\pm$0.07 & 3.57M &  87 \\
			7 & 97.42$\pm$0.07 & 3.73M& 50 \\
			10 & 97.46$\pm$0.12 & 4.06M & 35 \\
			\bottomrule
		\end{tabular}
	\end{center}
 	\vspace{-10pt}
	\caption{Sensitivity study of sampling number $K$. For each setting, we adjust the number of search epochs according to $K$ for fair comparison. We do three parallel tests on each setting and report the mean and standard deviation. }
	\label{tab:sensitivity_K}
\end{table}

Table~\ref{tab:sensitivity_K} compares the performance by setting $K$ as $1, 4, 7, 10$ in ROME.  We search and evaluate on CIFAR-10.  Three parallel tests on each setting are conducted. Note we adjust the number of search epochs to have same number of iterations per test. We observe that the performance increases monotonically with $K$ which verifies our analysis that biased training shall be alleviated.  The result demonstrates the effectiveness of our gradient accumulation strategy. Also, as the accuracy saturates at $K$=7, we set $K$=7.

\begin{table}[tb]
	\begin{center}
		\setlength{\tabcolsep}{5pt}
		\small
		\begin{tabular}{cccc | cccc}
			\toprule
			\multirow{2}{*}{TD} & \multicolumn{2}{c}{GA} & \multirow{2}{*}{Acc (\%)} & \multirow{2}{*}{TD} & \multicolumn{2}{c}{GA} & \multirow{2}{*}{Acc (\%)} \\
			\cmidrule(lr){2-3}
			 & $\bm{\theta}$ & $\bm{\alpha}$ & & & $\bm{\theta}$ & $\bm{\alpha}$ & \\
			\midrule
			$\times$ & $\times$ & $\times$ & 96.52$\pm$0.07 &  \checkmark & $\times$ & $\times$ & 97.12$\pm$0.06\\
			$\times$ & 	$\times$ & \checkmark & 96.85$\pm$0.31 & \checkmark & $\times$ & \checkmark & 97.22$\pm$0.07 \\
			$\times$ & \checkmark &	$\times$ & 96.98$\pm$0.05 & \checkmark & \checkmark & $\times$ & 97.34$\pm$0.07\\
			$\times$ & \checkmark & \checkmark & 97.14$\pm$0.05 & \checkmark & \checkmark & \checkmark & \textbf{97.42$\pm$0.07} \\
			\bottomrule
		\end{tabular}
	\end{center}
 	\vspace{-10pt}
	\caption{Component study of ROME on CIFAR-10.}
	\label{tab:componet_analysis}
\end{table}

\textbf{Component analysis for instability issue.}  There are two major components that contribute to the cure for instability in ROME: topology disentanglement (TD) and gradient accumulation (GA) for $\bm{\theta}$ and $\bm{\alpha}$. To show their efficacy, we conduct ablation studies in S0 space on CIFAR-10.  

Results are shown in Table~\ref{tab:componet_analysis}, where `GA for $\bm{\theta}$' indicates that gradient accumulation is applied over $K=7$ sampled architectures to train operation parameters; `GA for $\bm{\alpha}$' indicates that gradients for architectural parameters over $K=7$ sampled architectures is accumulated and averaged. The setting without TD and GA (first line in Table~\ref{tab:componet_analysis}) degenerates to GDAS~\cite{dong2019searching}. 
Our ROME adopts both TD and GA, which is the last line in Table~\ref{tab:componet_analysis}.

We observe that TD alone can significantly improve the searching performance, showing that the \emph{inconsistency issue is the principal reason for performance collapse}. This is as expected. On the one hand, inconsistent topology between search and evaluation results in an inconsistent searching objective; On the other hand, training the weights of 14 operations (w/o TD) is much more difficult than training 8 operations (w/ TD), which degrades the convergence. 

Moreover, we observe that gradient accumulation on $\bm{\theta}$ and $\bm{\alpha}$ can further improve the search performance, which confirms our analysis that performance collapse issue also comes from insufficient sampling.

\begin{table}[tb]
	\begin{center}	
		\small
		\begin{tabular}{lccccccc}
			\toprule
			\multirow{2}{*}{\textbf{Method}} & \multirow{2}{*}{DARTS} & \multirow{2}{*}{GDAS} &  \multicolumn{2}{c}{PC-DARTS} & \multirow{2}{*}{ROME}\\
			\cmidrule(lr){4-5}
			&&&M=4&M=2&\\
			\midrule
			\textbf{Memory} (G) & 9.4 & 3.1 & 3.7 &5.7& \textbf{2.3} \\
			\bottomrule
		\end{tabular}
	\end{center}
     \vspace{-10pt}
	\caption{GPU Memory cost comparison. We measured the cost based on a batch size of 64, where the supernet has 16 initial channels, and 8 layers.}
	\label{tab:mem_com}
	\vspace{-5pt}
\end{table}

\textbf{Memory Analysis.} \label{subsec:memory_compare}
Table~\ref{tab:mem_com} compares GPU memory cost in S0 search space on CIFAR-10. ROME has the lowest memory cost thanks to our disentanglement of the search for topology. Unlike GDAS that preserves multiple edges for each node, we strictly sample 2 edges for each node leading to 26\% memory reduction compared to GDAS.

PC-DARTS~\cite{Xu2020PC-DARTS} uses partial channels during the search stage to reduce GPU memory cost, in which the partial ratio is controlled by a hyperparameter $M$. But $M$ requires careful calibration for different tasks.
In contrast, ROME doesn't require  calibrating such an extra hyperparameter and is more memory-efficient.

\section{Comparison with Prior Works}
Here we highlight the difference of ROME from the existing works.  \textbf{1) ROME vs. DOTS.} DOTS~\cite{gu2021dots} explores an edge importance representation for one-shot NAS in a multi-stage fashion but the operations are divided into two groups (parameter-free and parameter-bearing, following DropNAS~\cite{hong2022dropnas}) beforehand, which is a very strong prior. The length of each stage has to be tuned carefully from dataset to dataset. In contrast, no prior or extra hyper-parameters are used in ROME; 
\textbf{2) ROME vs. DDW.} Unlike DDW~\cite{yuan2021differentiable} that belongs to dynamic networks with a changeable topology dependent on inputs, ROME is a NAS method designed to search for a static architecture. 
\textbf{3) ROME vs. SNAS.} SNAS~\cite{cai2018proxylessnas} adopts Gumbel-softmax via masking, whose supernet still resides in the memory and thus not memory-efficient. In contrast, ROME is a truly single-path NAS method and inherits the property of low memory cost. SNAS didn't deal with the collapse issue either. See more details in Sec.~\ref{supp:sec:difference} (supplementary).

\section{Conclusion}In this paper, we highlight the performance collapse issue of single-path differentiable NAS, and attribute the cause to topology inconsistency between searching and evaluation, as well as the stochastic nature of sampling for candidate operations. 
To address the above issues, we propose ROME that features topology disentanglement and gradient accumulation strategy to stabilize the searching process. ROME achieves state-of-the-art results across 15 recent popular benchmarks, which manifests its strong performance, low memory cost and robustness. 

\section*{Acknowledgments}
This work was partly  supported by National Key R\&D Program of China (No. 2022ZD0118700),  NSFC (62222607), Shanghai Municipal Science and Technology Major Project (2021SHZDZX0102).

{\small
\bibliographystyle{ieee_fullname}
\bibliography{mybib}
}

\appendix
\section{Proof of Gumbel-Top2 Process}\label{supp:sec:proof-gumbel}
This section will prove that the sampling scheme proposed in ROME-v2, Gumbel-Top2 technique,is equivalent to sampling two different edges without replacement, with the probability simplex $p_i = \frac{\exp{\beta_i}}{\sum_{i'} \exp(\beta_{i'})}$. 

To complete this, we need to prove that each edge has the same probability of being selected in these two schemes. Let $p_i$ be the probability of choosing $i$-th edge among $n$ edges at one time. Without loss of generality, we suppose that $e_1$ is chosen.

\paragraph{i).} We first discuss sampling two edges in order without replacement. The cases that $e_1$ is chosen can be divided into two disjoint parts:
\begin{enumerate}[A)]
	\item  It is selected by the first choice, whose probability is $p_1$.
	\item It is selected by the second choice, and the probability is $\sum_{i=2}^n p_i \frac{p_1}{1 - p_i}$, where $\frac{p_1}{1 - p_i}$ is the scaled probability when taking $i$-th edge away without putting it back.
\end{enumerate}

In total, the probability of $e_1$ being chosen is 
\begin{gather}
	\label{eq1}
	p_1 + \sum_{i=2}^n p_i \frac{p_1}{1 - p_i}.
\end{gather}

\paragraph{ii).} Further, we discuss the Gumbel Top-2 scheme, in which we sample $n$ real numbers $\epsilon_k$ from $U[0, 1]$ at first, and the probability of choosing each edge $e_k$ is $\tilde{q}_k$:
\begin{align}
	&q_k = \log p_k - \log (-\log \epsilon_k), \quad
	&\tilde{q}_k = \frac{\exp (q_k)}{\sum_{k'=1}^n\exp(q_{k'})}.
\end{align}
There are also two cases where $e_1$ will be chosen:
\begin{enumerate}[A)]
	\item $\tilde{q}_1$ is the largest one among all edges, that is:
	\begin{align}
		q_1 > q_j, \forall j\notin \{1\}.
	\end{align}
	By reformatting these inequalities, we have
	\begin{align}
		\epsilon_j < \epsilon_1^{p_j/p_1}, \forall j\notin \{1\}
	\end{align}
	Since each $\epsilon_i$ is sampled from $U[0,1]$ independently, we can obtain the joint probability of all these events.
	\begin{align}
		P &=  \prod_{j=2}^n P(\epsilon_j < \epsilon_1^{p_j/p_1}) =\prod_{j=2}^n \left[\int_{0}^1\int_0^{\epsilon_1^{p_j/p_1}}1 \ \text{d}\epsilon_j\text{d}\epsilon_1 \right] \nonumber\\
		&=\int_{0}^1 \prod_{j=2}^n \epsilon_1^{p_j/p_1} \text{d}\epsilon_1 = \int_{0}^1 \epsilon_1^{\frac{1}{p_1}-1}\text{d}\epsilon_1
		=  p_1
	\end{align}
	So the probability of case A) is $p_1$.
	
	\item $\tilde{q}_1$ is the second largest one only next to $\tilde{q}_i$. That is
	\begin{align}
		q_1 < q_i; \quad
		q_1 > q_j, \forall j\notin \{1, i\}.
	\end{align}
	By reformatting these inequalities, we have
	\begin{align}
		\epsilon_i >  \epsilon_1^{p_i/p_1}; \quad 
		\epsilon_j < \epsilon_1^{p_j/p_1}, \forall j\notin \{1, i\}.
	\end{align}
\end{enumerate}

Similar to case A), since each $\epsilon_i$ is sampled from $U[0,1]$ independently, we can get the joint probability of all these events.
\begin{align}
	P &=  \int_0^1 (1-\epsilon_1^{p_i/p_1}) \prod_{j\notin\{1,i\}}^n\epsilon_1^{p_j/p_1} \text{d}\epsilon_1 =  \int_0^1 (1 - \epsilon_1^{\frac{p_i}{p_1}})\epsilon_1^{\frac{1 - p_i}{p_1} - 1} \text{d}\epsilon_1 \nonumber \\
	&= \int_{0}^1 \epsilon_1^{\frac{1 - p_i}{p_1} - 1} - \epsilon_1^{\frac{1}{p_1} - 1} d\epsilon_1 = \frac{p_1}{1 - p_i} - p_1 
	= p_i \frac{p_1}{1 - p_i}.
\end{align}
Enumerating $i$ from $2$ to $n$, the probability of case B) is $\sum_{i=2}^n p_i \frac{p_1}{1 - p_i}$.

In all, the probability of $e_1$ being chosen is $p_1 + \sum_{i=2}^n p_i \frac{p_1}{1 - p_i}$ as well, which meets Eq. \ref{eq1}. Therefore, these two schemes i) and ii) are equivalent.

\begin{table*}[h]
	\caption{Comparison in RobustDARTS \cite{zela2020understanding} reduced search spaces and 3 datasets. We report the \textbf{lowest error rate} of 3 found architectures. $^\dagger$:  using the settings of \cite{zela2020understanding} where CIFAR-100 and SVHN models have 8 layers and 16 initial channels, CIFAR-10 models have 20 layers and 36 initial channels except that S2 and S4 have 16 initial channels. 
		$^\star$: using the settings in S-DARTS~\cite{chen2020stabilizing}, where all models have 20 layers and 36 initial channels. Others utilize the settings in RobustDARTS. The best is underlined and in bold, the second-best is in bold.}
	\label{supp:tab:comparison-rdarts-s1-4-best}
	\center
	\resizebox{.99\textwidth}{!}{
		\setlength{\tabcolsep}{3pt}
		\begin{tabular}{c*{7}{c}H*{2}{c}|*{2}{c}H*{2}{c}}
			\toprule
			\multicolumn{2}{c}{\multirow{2}*{\textbf{Benchmark}}} &   \multirow{2}*{\textbf{DARTS}$^\dagger$}  &  \multicolumn{2}{c}{\textbf{R-DARTS}$^\dagger$} & \multicolumn{2}{c}{\textbf{DARTS}$^\dagger$} &  \multirow{2}*{\textbf{SDARTS-RS}$^\dagger$} & &  \multicolumn{2}{c|}{\textbf{ROME (ours)}$^{\dagger}$} & \multirow{2}*{\textbf{PC-DARTS}$^\star$} & \multirow{2}*{\textbf{SDARTS-RS}$^\star$} & & \multicolumn{2}{c}{\textbf{ROME (ours)}$^\star$} \\
			\cmidrule(lr){4-5} \cmidrule(lr){6-7} \cmidrule(lr){9-11} \cmidrule(lr){15-16}
			& & & DP & L2 & ES & ADA &   & ADV & v1 & v2 & &  & ADV & v1 & v2 \\
			\midrule
			\multirow{4}*{C10}& S1  & 3.84 & 3.11 & 2.78 & 3.01 & 3.10 & 2.78 & 2.73 &  \textbf{2.68}  & \textbf{\underline{2.62}} & 3.11 & 2.78 & 2.73 & \textbf{2.68} & \textbf{\underline{2.62}} \\
			~  & S2 &  4.85 &  3.48 &  3.31 &  3.26 &  3.35 & 3.33 & 3.41 & \textbf{3.24} & \textbf{\underline{2.95}} & 3.02 &  \textbf{2.75} &  \textbf{2.65} & 2.79 & \textbf{\underline{2.62}}  \\
			~     &  S3 &  3.34 &  2.93 &  \textbf{\underline{2.51}} &  2.74 &  2.59  &  \textbf{2.53} & \textbf{\underline{2.49}} & 2.65   & 2.58 & \textbf{\underline{2.51}} & \textbf{2.53} &  \textbf{\underline{2.49}} & 2.65 & 2.58 \\
			~ & S4 & 7.20 & 3.58 & 3.56 & 3.71 & 4.84 & 4.84 & 4.28  & \textbf{\underline{3.21}} & \textbf{3.31} & 3.02 & \textbf{2.93} & 2.87 & 3.61  & \textbf{\underline{2.68}} \\
			\midrule
			\multirow{4}*{C100} & S1 & 29.46 & 25.93 & 24.25  &  28.37 & 24.03 & 23.51 & \textbf{22.33} & 22.34 & \textbf{\underline{22.04}} & 18.87 & \textbf{\underline{17.02}} & \textbf{\underline{16.88}} & 17.27  & \textbf{17.24} \\ 
			~ &  S2 &  26.05 &  22.30 &  22.24 &  23.25 &  23.52 & 22.28 & \textbf{\underline{20.56}}  & \textbf{\underline{21.95}} & \textbf{22.12} & 18.23 & 17.56 &  17.24  &  \textbf{17.09} & \textbf{\underline{17.06}}   \\
			~     &  S3 &  28.90 &  22.36 &  23.99 &  23.73 &  23.37  & \textbf{\underline{21.09}} & \textbf{\underline{21.08}} & 22.56 & \textbf{22.11}  & 18.05 & 17.73 &  17.12   & \textbf{16.95}  & \textbf{\underline{16.94}} \\ 
			~ & S4 & 22.85 & 22.18 & 21.94 & \textbf{21.26} & 23.20 & 21.46 &  21.25 &   21.33 & \textbf{\underline{20.44}} & 17.16 & 17.17 & \textbf{\underline{15.46}} & \textbf{\underline{15.99}} & \textbf{16.18} \\
			\midrule
			\multirow{4}*{SVHN} & S1 & 4.58 & 2.55 & 4.79 & 2.72 & 2.53  & 2.35 & \textbf{2.29} & \textbf{2.33}  & \textbf{\underline{2.27}} &  2.28 & 2.26 & 2.16 & \textbf{\underline{2.07}} & \textbf{2.14} \\ 
			&S2 & 3.53 & 2.52 & 2.51 & 2.60 & 2.54 & 2.39 & \textbf{2.35} & \textbf{2.39} & \textbf{\underline{2.30}} & 2.39 & 2.37  & 2.07 & \textbf{2.14} & \textbf{\underline{2.07}} \\
			&S3 & 3.51 & 2.49 & \textbf{2.48} & 2.50 & 2.50 & \textbf{\underline{2.36}} & 2.40 & 2.58  & 2.51 & 2.27 & 2.21 & 2.05 & \textbf{2.14} & \textbf{\underline{2.07}}\\
			&S4 & 3.05 & 2.61 & 2.50 & 2.51 & 2.46 & 2.46 & \textbf{2.42} & \textbf{2.43} & \textbf{\underline{2.34}} & 2.37 & 2.35 & 1.98 & \textbf{2.00} & \textbf{\underline{1.99}} \\
			\bottomrule
		\end{tabular}
	}
\end{table*}

\section{Detailed Discussion with Prior Works}
\label{supp:sec:difference}

\subsection{Topology Disentanglement}
DOTS~\cite{gu2021dots} is related to our work that proposes to decouple the operation search and topology search into two separate stages. However, It is methodologically different from ROME. We highlight some key features of ROME. \textbf{ 1) No-Prior:} To alleviate the collapse, DOTS uses a strong human grouping prior as StacNAS, which classifies the operations into two groups: parametric and non-parametric. ROME uses no prior at all. \textbf{ 2) Single-phase searching with no extra hyperparameters:} DOTS contains two phases: search operations first, and then search topologies with the fixed operations. It uses three carefully designed and {tuned} hyperparameters ($T_0, T_{\beta}, T_{\alpha_{On}}$) to control the percentage of two phases for different datasets (through our communication with the authors of DOTS). In contrast, ROME is single-phase as DARTS and it requires no  specific hyperparameters tailored for different datasets. \textbf{ 3) Memory efficiency:} 
ROME (2.3G) costs 1/4 of DOTS's memory (9.5G), since DOTS trains the whole supernet during the operation search.

\subsection{Gumbel Reparameterization in NAS}
GDAS and SNAS are contemporary works based on Gumbel-softmax reparameterization technique. Nevertheless, GDAS is memory-efficient since it sample and activate a sub-set of candidate operations, while SNAS still belongs to one-shot NAS since all operations participate the forward and backward at each iteration in the search stage. 
This work research on GDAS and point out that the performance collapse issue also exists. We attribute it to two aspects, that differs from the reason in DARTS: 1) the topology inconsistency between searching and evaluation and 2) the stochastic nature of sampling for candidate operations. Topology disentanglement and gradient accumulation techniques are proposed to stabilize the search process for GDAS. 
Our method, ROME inherits the property of GDAS, i.e., low GPU memory requirement and high speed for searching.
In comparison, SNAS has little relation to ROME. It requires vast GPU memory like DARTS and still suffers performance collapse issue. 

\subsection{Dynamic Network}
DDW~\cite{yuan2021differentiable} is a kind of dynamic network whose topology dynamically changes based on the input. In contrast, ROME is a NAS method, whose topology is fixed after searching. Unlike DDW that limited to some handcrafted architectures e.g. ResNet, MobileNetV2, ROME supports more complex topologies as DARTS's search space. Moreover, DDW is not memory efficient as it keeps the whole supernet in the memory, while ROME requires much less GPU memory.

\section{Further Experiments}

\subsection{Robustness evaluation on 12 hard benchmarks}
We follow RobustDARTS~\cite{zela2020understanding} and evaluate the performance and generalization of our method across three datasets on S1-S4 search spaces, where DARTS severely suffers from performance collapse.
We independently search four times under different random seeds for each benchmark and train the discovered models to report their mean and standard variance performance. This process is recommended by \cite{Yang2020NAS,chen2020stabilizing,zela2020understanding,chu2021darts} to fairly compare different NAS methods. 
Table~\ref{supp:tab:comparison-rdarts-s1-4-best} reports the best performance, showing that our methods robustly outperform RobustDARTS with a clear margin across the 12 benchmarks. The best cells found by ROME are shown in the next section.

\subsection{Discussion on collapse behavior across popular NAS benchmarks.} We argue that excluding an important operation for search space can cause illusive conclusions. Specifically, NAS-Bench-1Shot1 \cite{zela2020nas} suggests that Gumbel-based NAS is quite robust. However, this observation is laying on the basis that popular skip connections are not included in the search space \cite{pmlr-v97-ying19a}. After adding skip connection into the choices, we perform the GDAS search using their released code \footnote{https://github.com/automl/nasbench-1shot1}. The best model found is full of skip connections, which again supports our discovery of collapse issue in single-path based NAS, see Fig.~\ref{fig:1shot1-gdas-fail} and more 
in Fig.~\ref{fig:1shot1-gdas-fail-rest} in the supplemental material.
Instead, we do not suffer the same issue while performing ROME in these search spaces (see 
Fig.~\ref{fig:1shot1-rome-best}).

\begin{figure}[tb!]
	\centering
	\begin{subfigure}{0.45\columnwidth}
		\includegraphics[width=0.98\columnwidth]{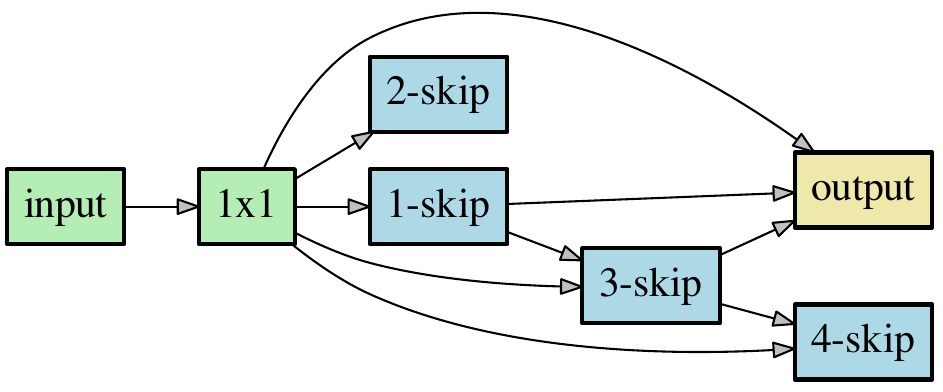}
		\caption{GDAS \cite{dong2019searching}}
		\label{fig:1shot1-gdas-fail-gdas}
	\end{subfigure}
	\hfill
	\begin{subfigure}{0.49\columnwidth}
		\includegraphics[width=0.98\columnwidth]{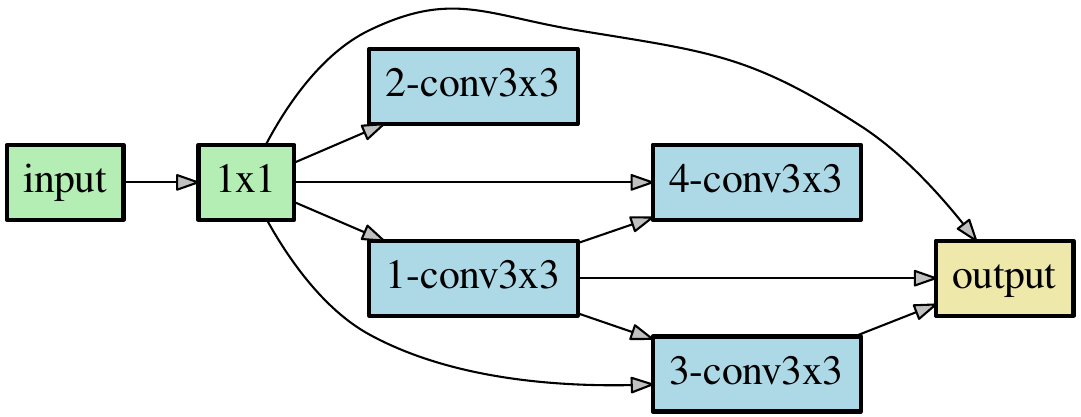}
		\caption{ROME}
		\label{fig:1shot1-gdas-fail-rome}
	\end{subfigure}
	\caption{GDAS fails on NAS-Bench-1Shot1~\cite{zela2020nas} on CIFAR-10 when adding skip connection to the second search space. Notice that nodes with no out-degrees have no contribution to the output.
	}
	\label{fig:1shot1-gdas-fail}
\end{figure}

\section{Figures of Genotypes}\label{supp:fig-geno}

\begin{figure}[tb!]
	\centering
	\includegraphics[width=0.45\columnwidth]{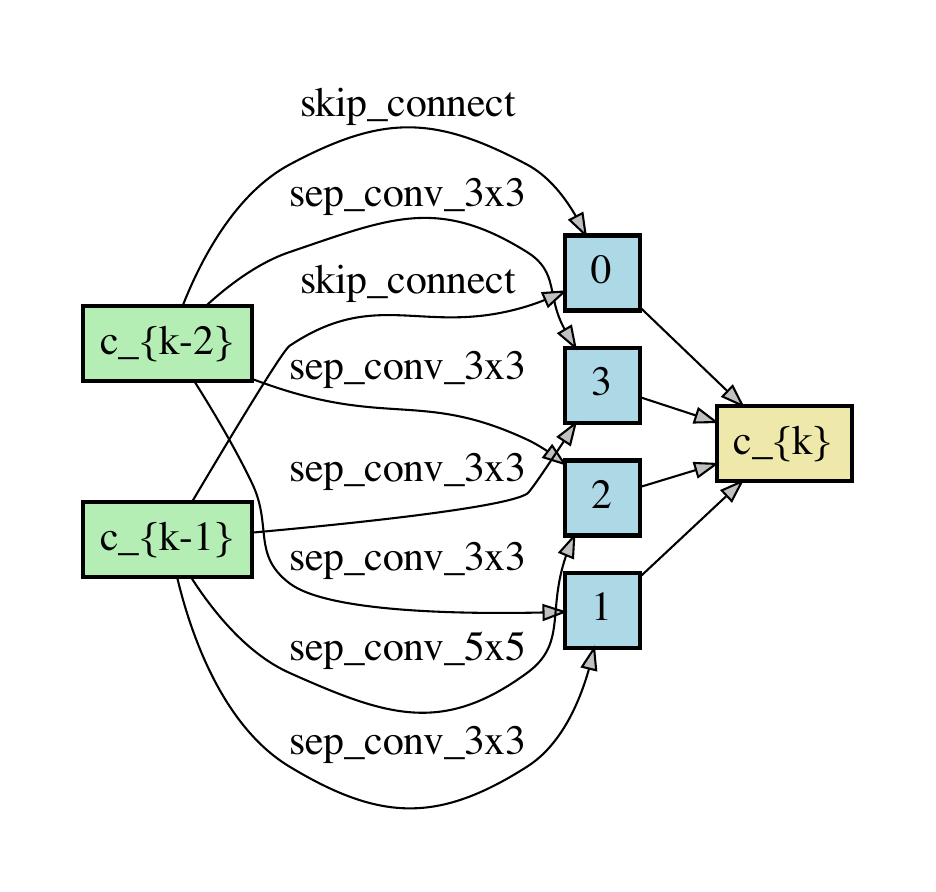}
	\includegraphics[width=0.45\columnwidth]{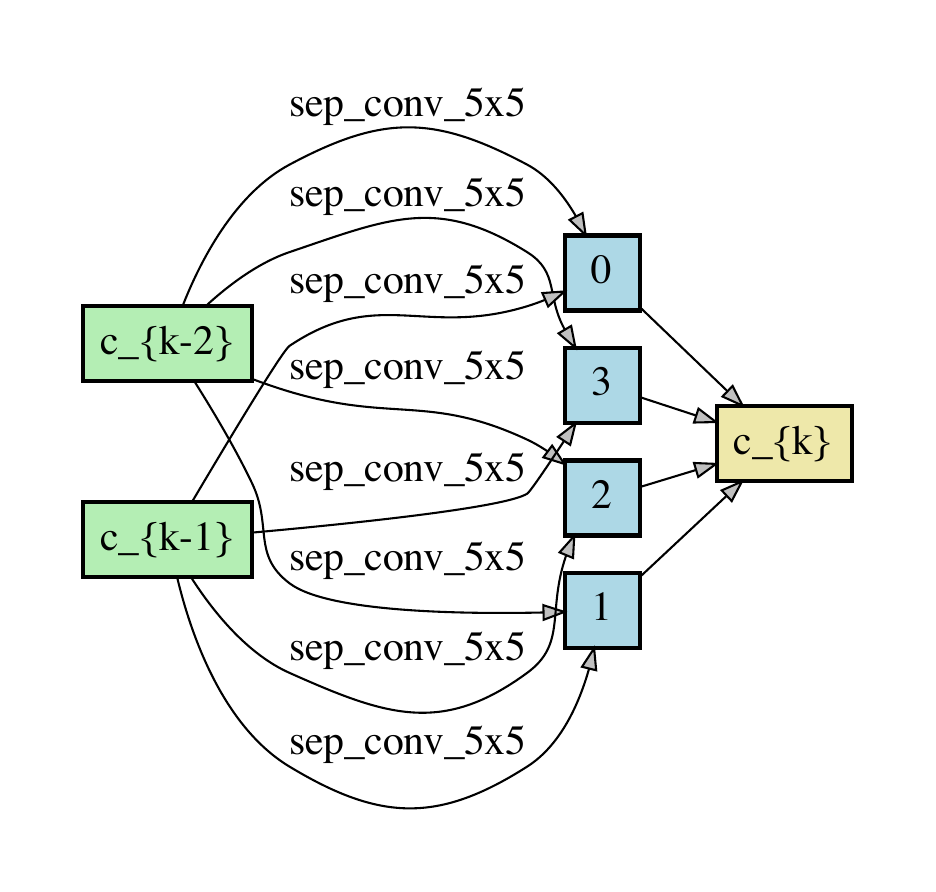}
	\caption{Best normal and reduction cells discovered by ROME-v1 on CIFAR-10.}
	\label{fig:rome-v1-cifar10-best-geno}
\end{figure}

\begin{figure}[tb!]
	\centering
	\includegraphics[width=0.43\columnwidth]{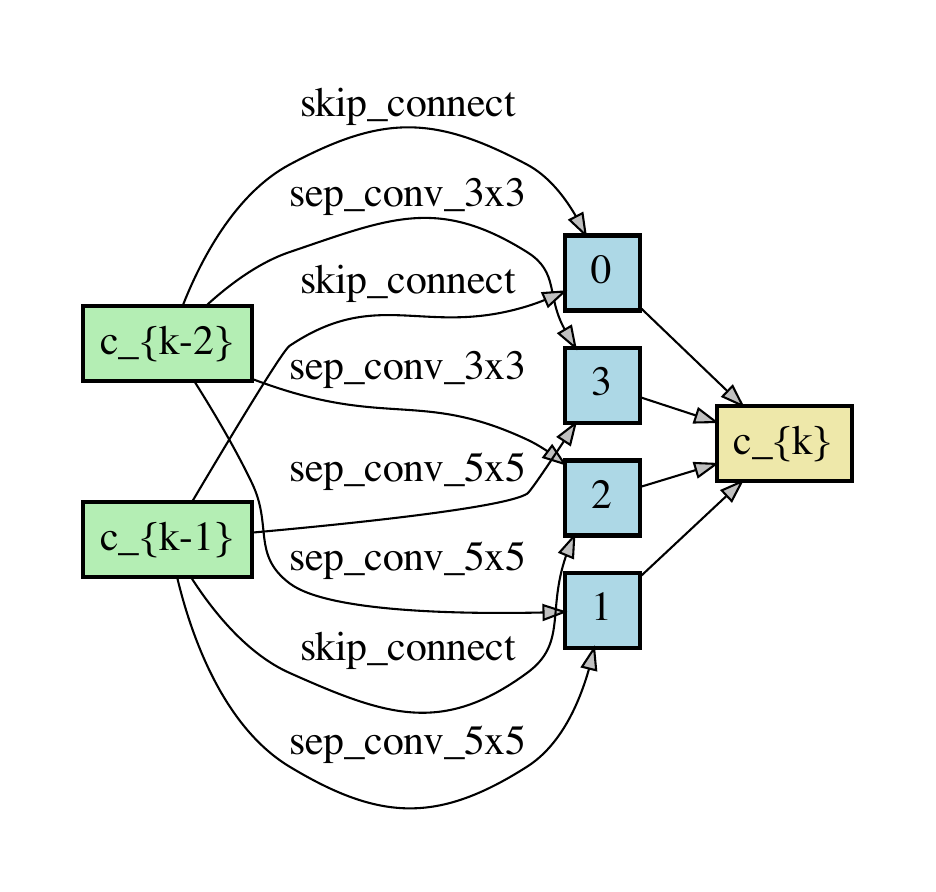}
	\includegraphics[width=0.48\columnwidth]{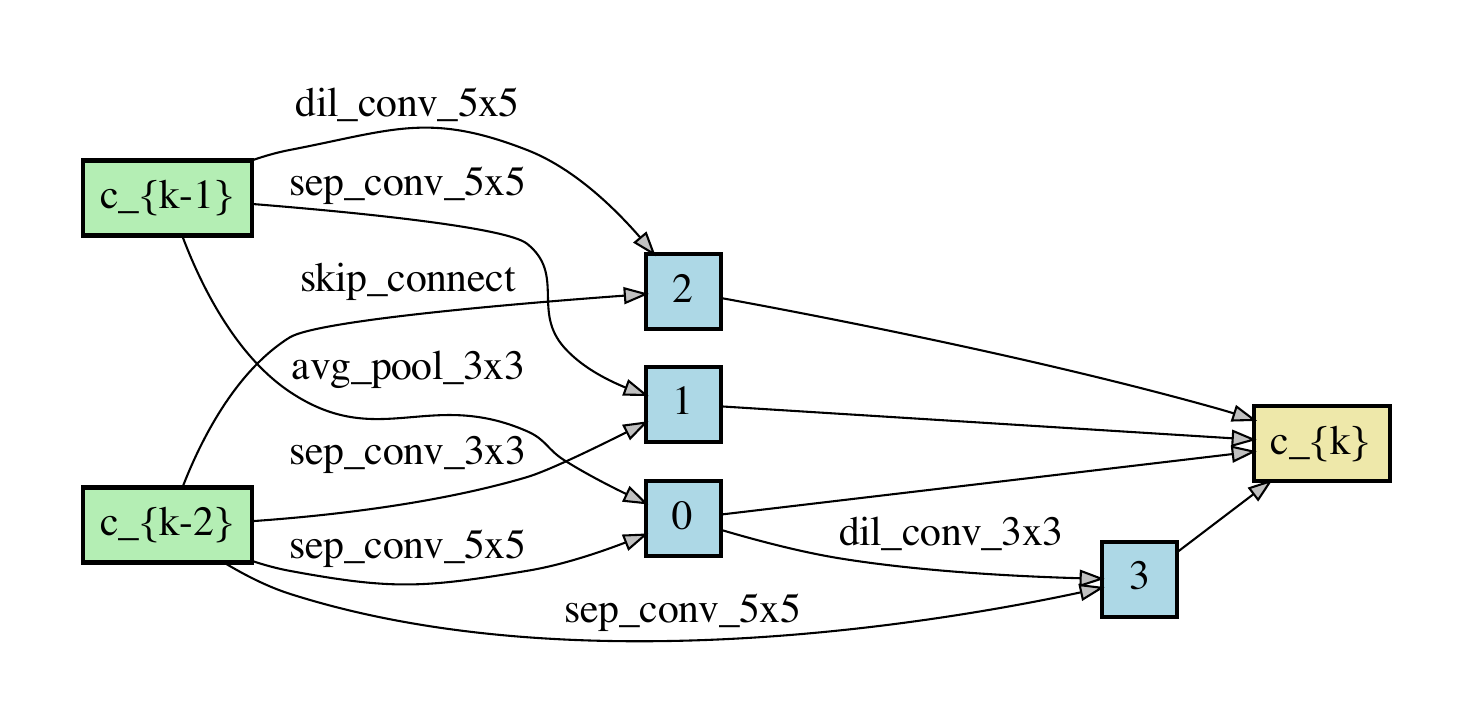}
	\caption{Best normal and reduction cells discovered by ROME-v2 on CIFAR-10.}
	\label{fig:rome-v2-cifar10-best-geno}
\end{figure}

\begin{figure}
	\begin{subfigure}{0.58\columnwidth}
		\includegraphics[width=0.98\columnwidth]{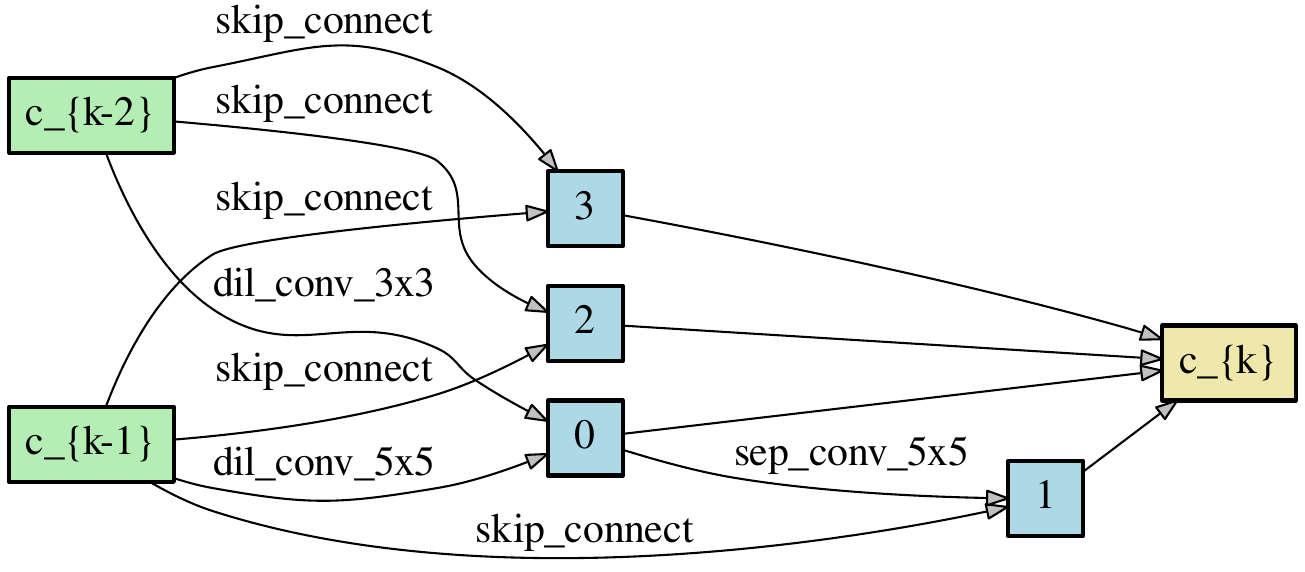}
		\caption{GDAS \cite{dong2019searching}}
		\label{fig:compare-s0-imagenet-gdas}
	\end{subfigure}
	\hfill
	\begin{subfigure}{0.36\columnwidth}
		\includegraphics[width=0.98\columnwidth]{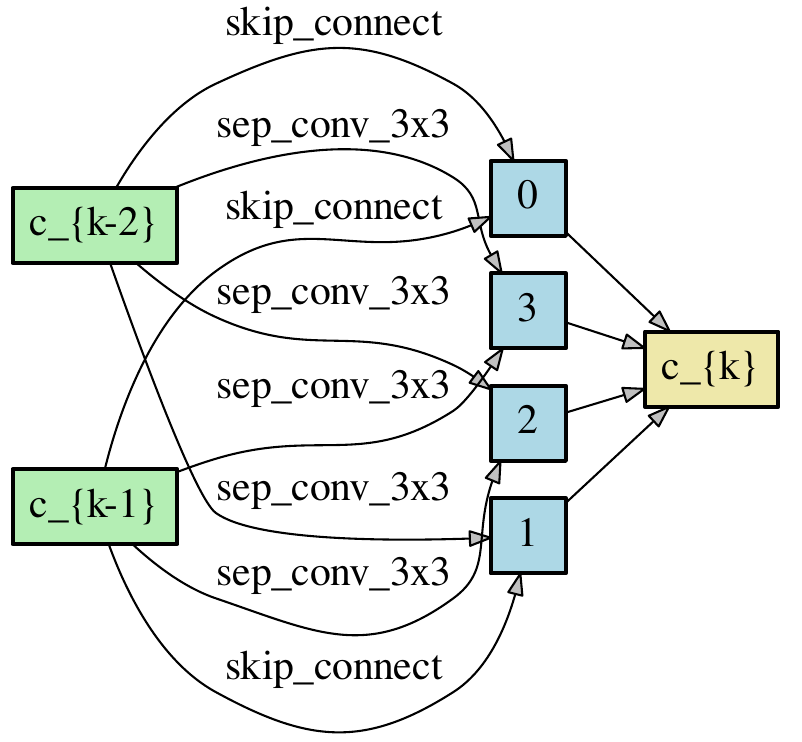}
		\caption{GDAS \cite{dong2019searching}}
		\label{fig:compare-s0-imagenet-rome}
	\end{subfigure}
	\caption{The architecture of normal cells searched by GDAS and ROME on ImageNet under S0 search space. Network searched by GDAS is dominated by skip connection and only obains $72.5\%$ accuracy on ImageNet, while our method is much more stable and achieves $75.5\%$ accuracy.}
	\label{fig:compare-s0-imagenet}
\end{figure}

\begin{figure}
	\centering
	\includegraphics[width=0.45\columnwidth]{./img/GDAS_IMAGENET_S0_best-normal}
	\includegraphics[width=0.50\columnwidth]{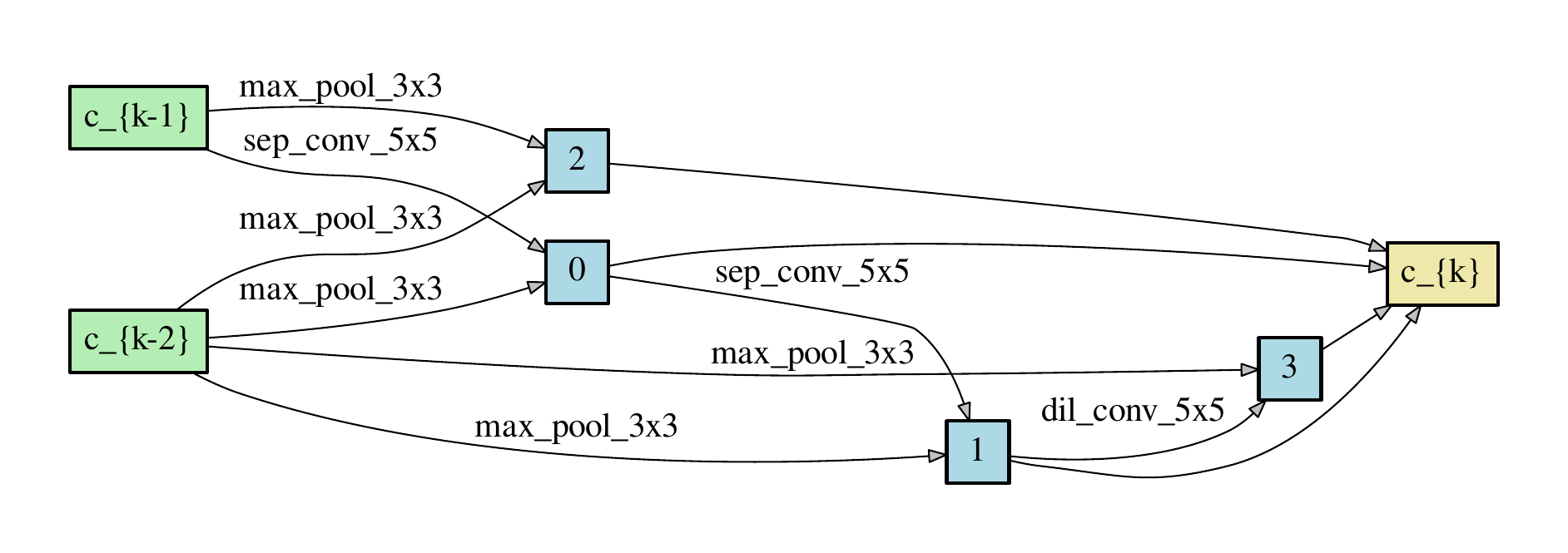}
	\caption{The best architecture found by GDAS on ImageNet in S0. Skip connection dominate the searched architecture. Top-1 accuracy on the validation set is 72.5\%.}
	\label{fig:gdas-s0-imagenet}
\end{figure}

\begin{figure}
	\centering
	\includegraphics[width=0.45\columnwidth]{./img/ROME_IMAGENET_S0_best-normal}
	\includegraphics[width=0.45\columnwidth]{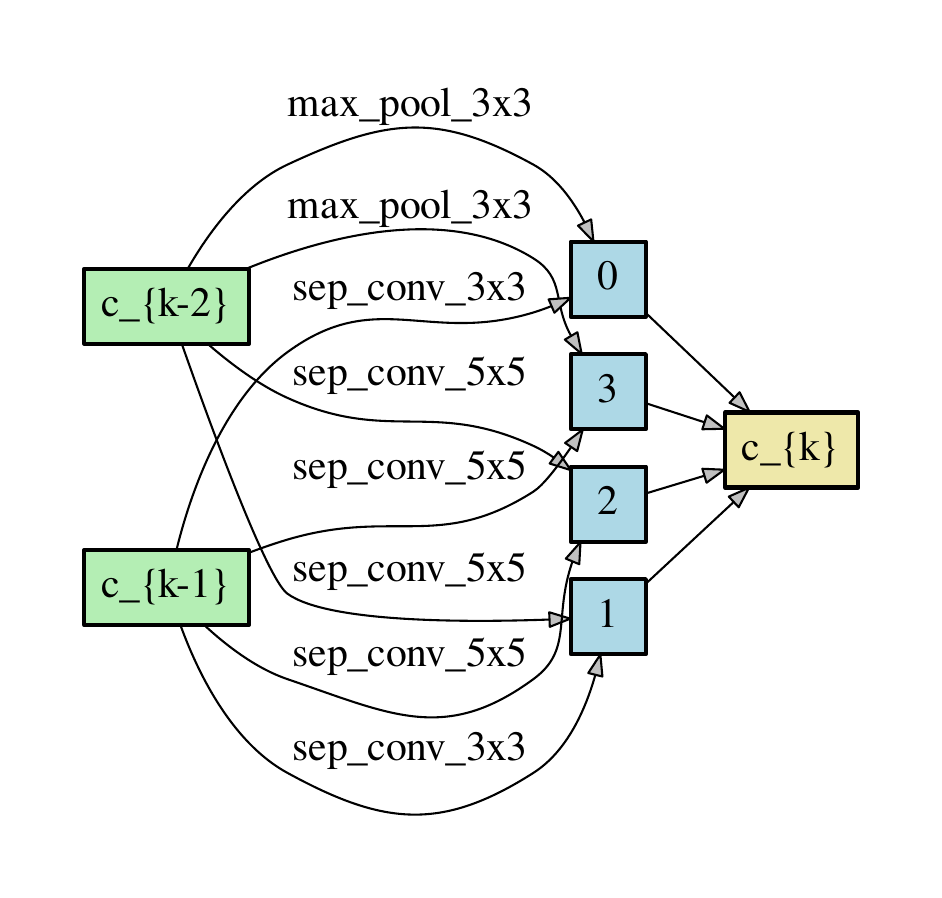}
	\caption{The best architecture found by ROME on ImageNet in S0. No performance collapse occurs. Top-1 accuracy on the validation set is 75.5\%}
	\label{fig:rome-s0-imagenet}
\end{figure}

\begin{figure}[ht]
	\centering
	\begin{subfigure}{0.98\columnwidth}
		\includegraphics[width=0.45\columnwidth]{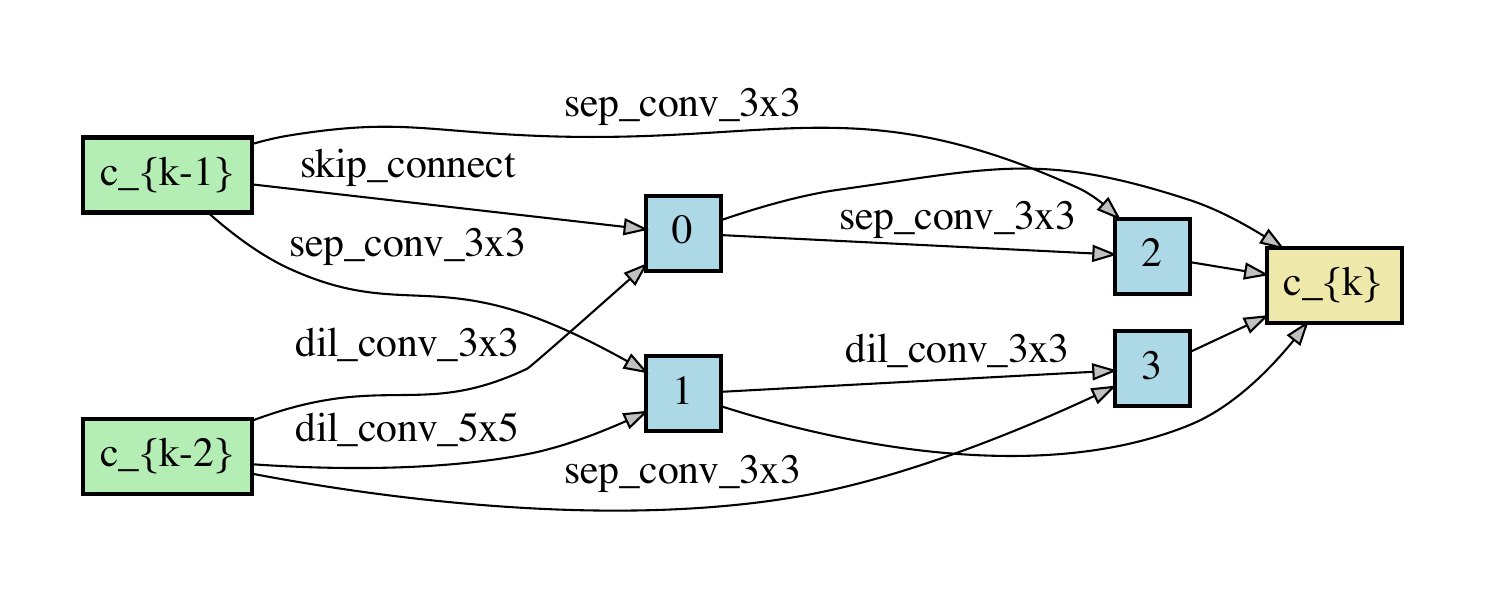}
		\includegraphics[width=0.45\columnwidth]{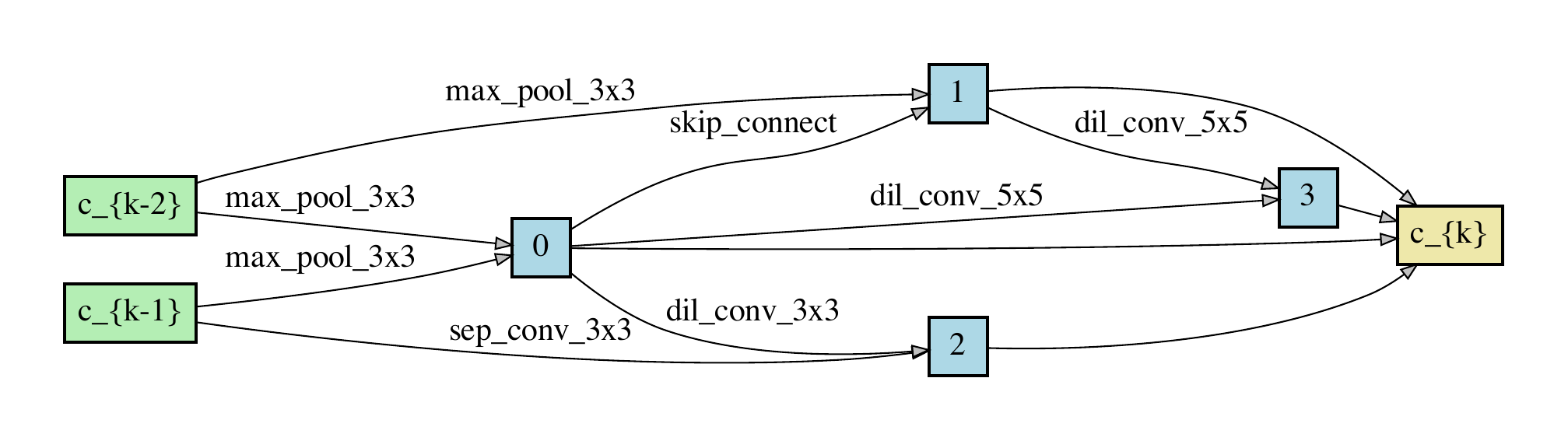}    
		\caption{S1}
		\label{fig:rome-v1-cifar10-rdarts-ss-s1}
	\end{subfigure}
	
	\begin{subfigure}{0.98\columnwidth}
		\includegraphics[width=0.45\columnwidth]{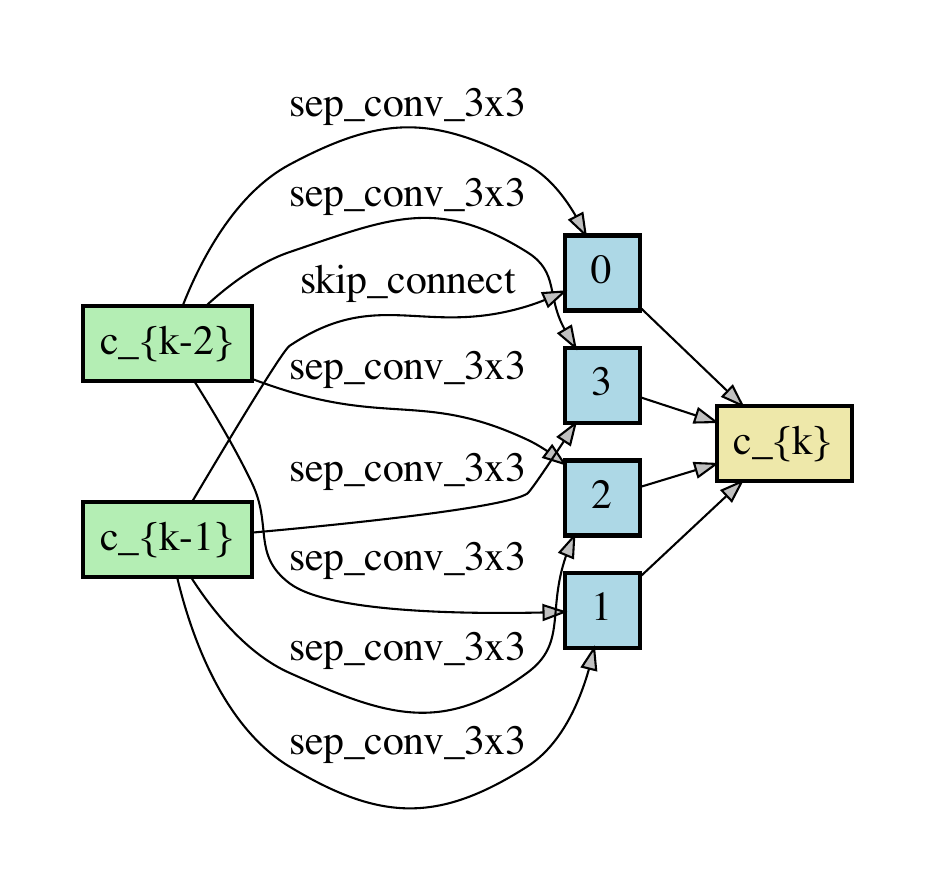}
		\includegraphics[width=0.45\columnwidth]{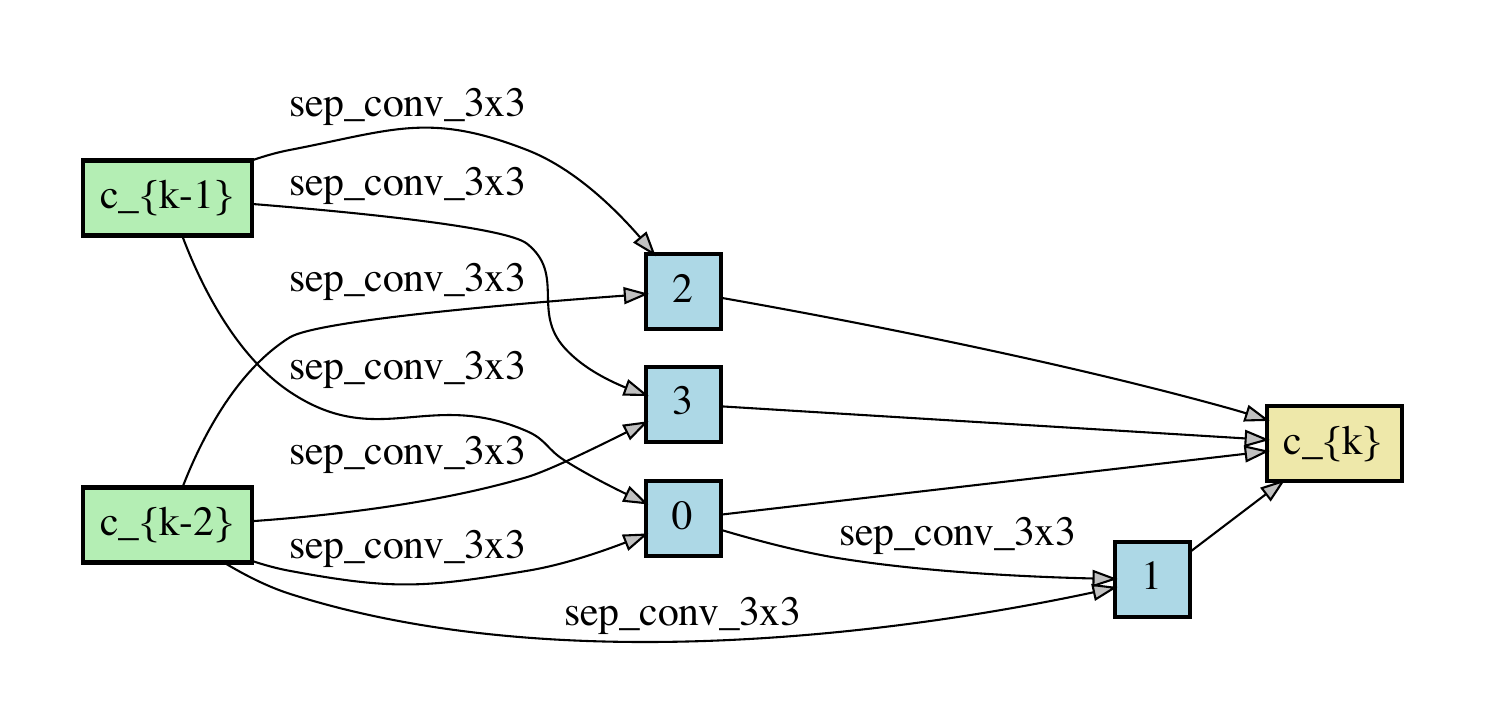} 
		\caption{S2}
		\label{fig:rome-v1-cifar10-rdarts-ss-s2}
	\end{subfigure}
	
	\begin{subfigure}{0.98\columnwidth}
		\includegraphics[width=0.45\columnwidth]{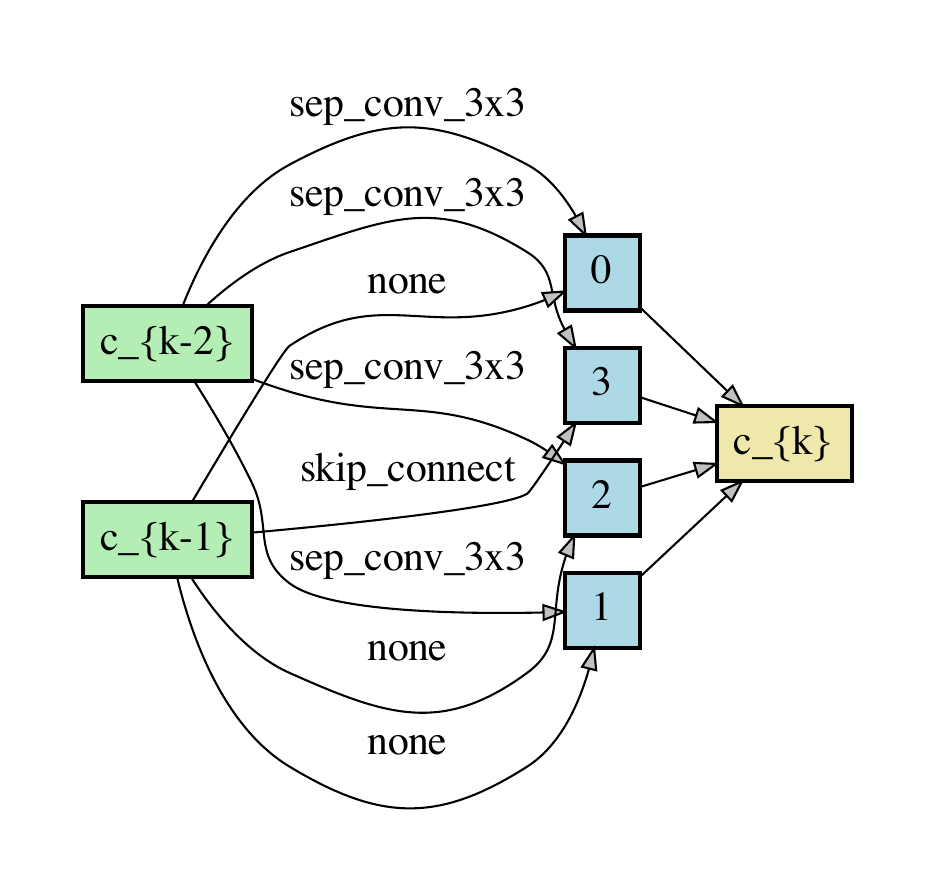}
		\includegraphics[width=0.45\columnwidth]{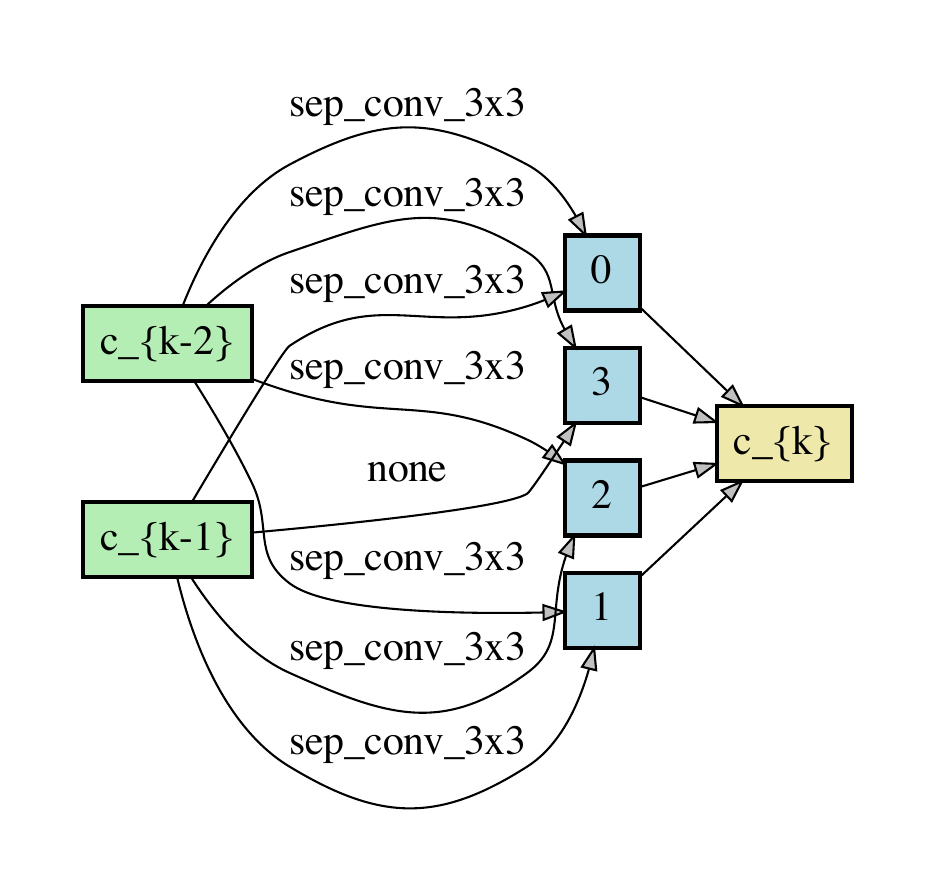}
		\caption{S3}
		\label{fig:rome-v1-cifar10-rdarts-ss-s3}
	\end{subfigure}
	
	\begin{subfigure}{0.98\columnwidth}
		\includegraphics[width=0.45\columnwidth]{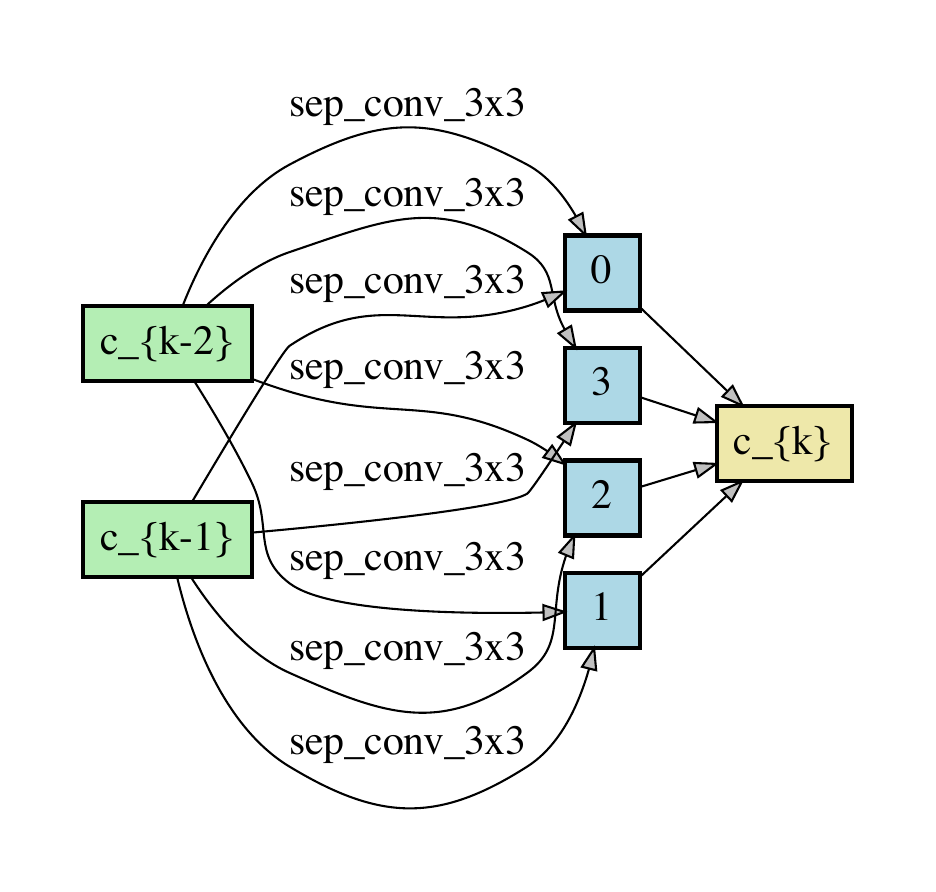}
		\includegraphics[width=0.45\columnwidth]{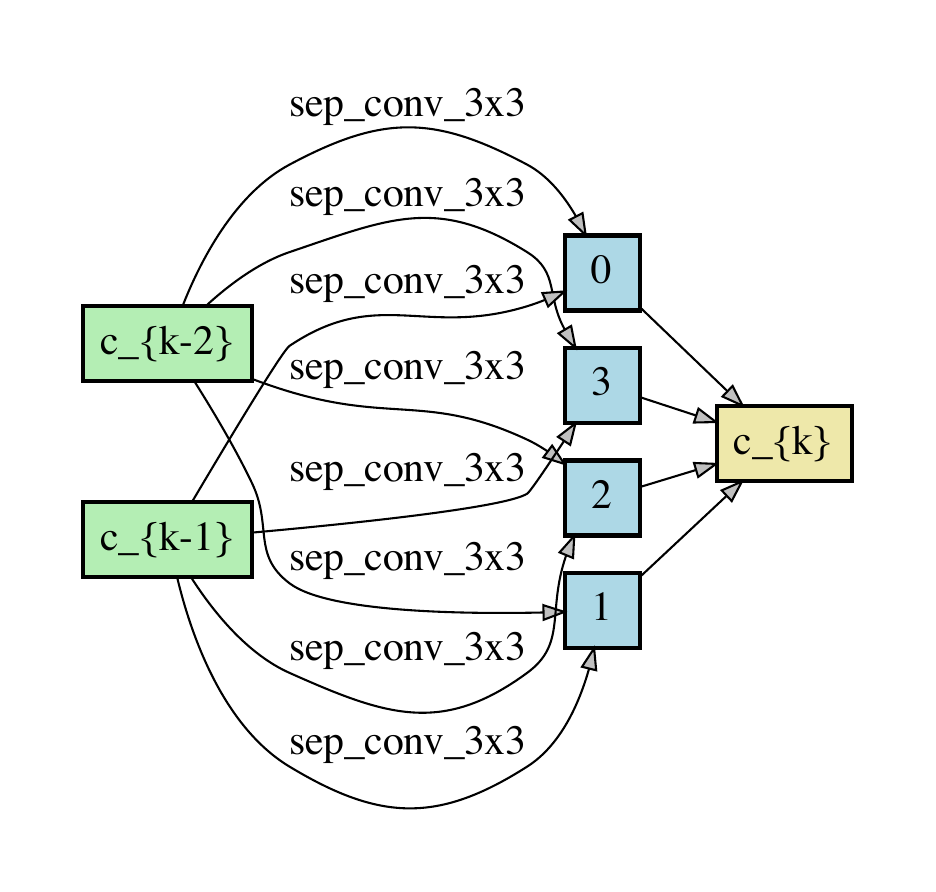}
		\caption{S4}
		\label{fig:rome-v1-cifar10-rdarts-ss-s4}
	\end{subfigure}
	\caption{ROME-V1 best cells (paired in normal and reduction) on CIFAR10 in reduced search spaces of RobustDARTS.}
	\label{fig:rome-v1-cifar10-rdarts-ss}
\end{figure}

\begin{figure}[ht]
	\centering
	\begin{subfigure}{0.98\columnwidth}
		\includegraphics[width=0.45\columnwidth]{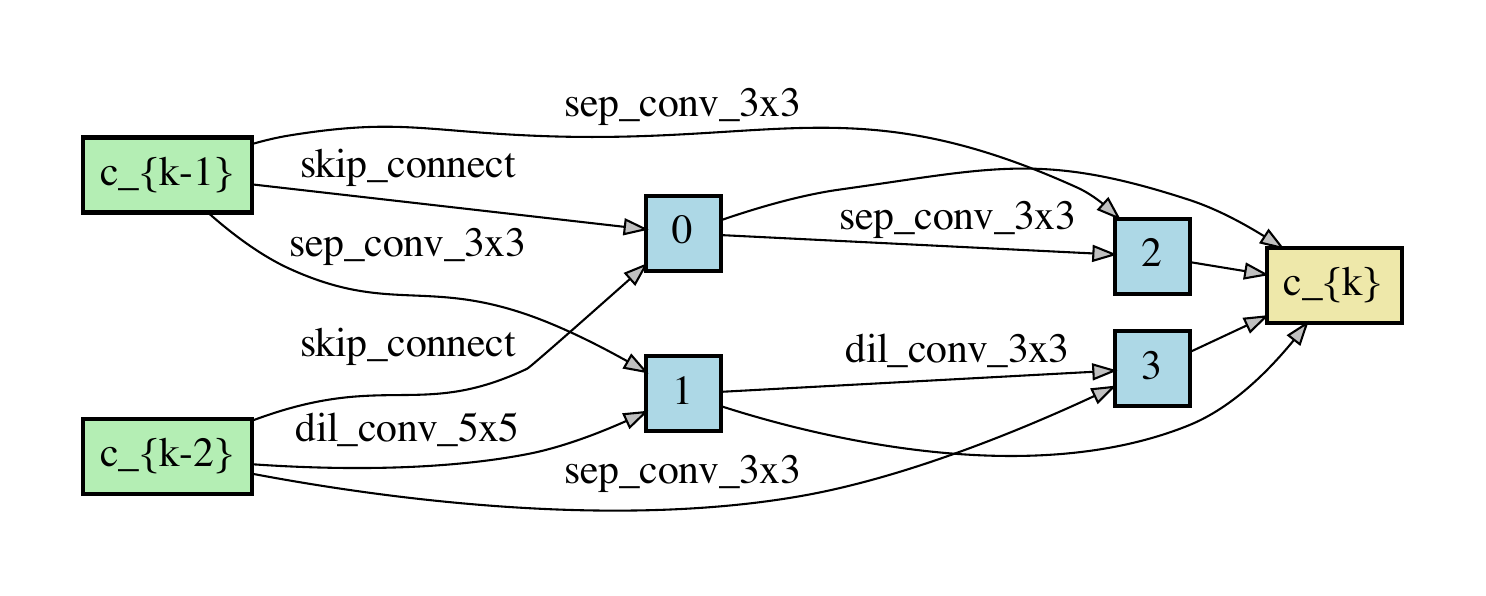}
		\includegraphics[width=0.45\columnwidth]{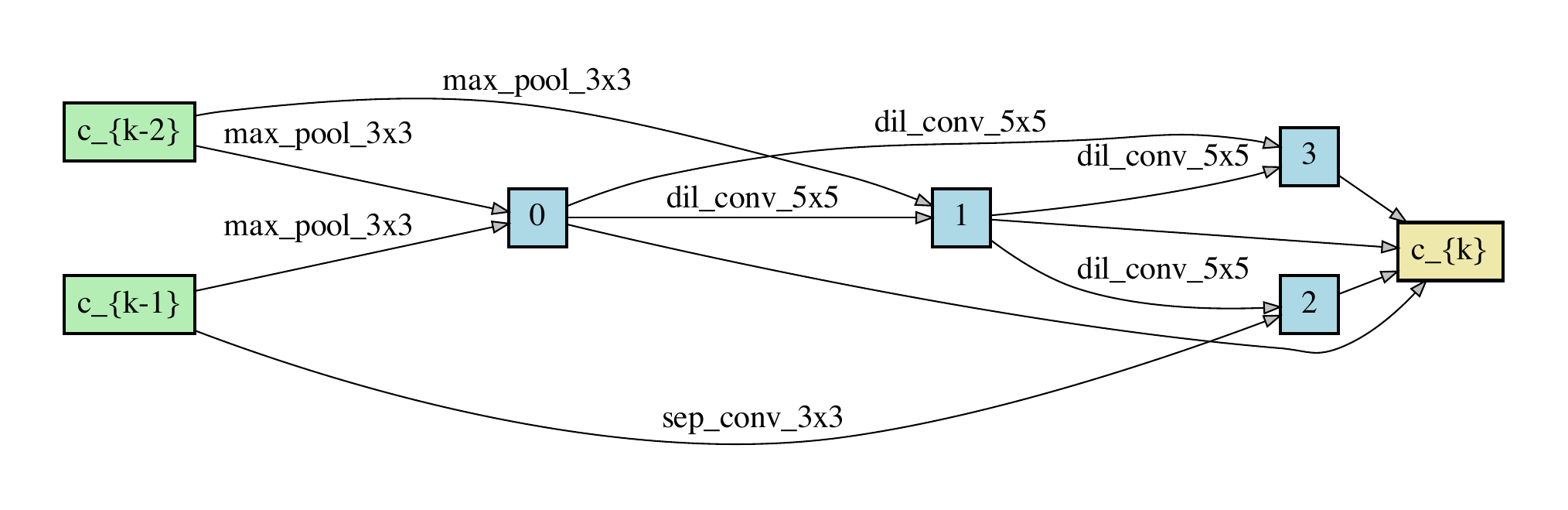}
		\caption{S1}
		\label{fig:rome-v1-cifar100-rdarts-ss-s1}
	\end{subfigure}
	
	\begin{subfigure}{0.98\columnwidth}
		\includegraphics[width=0.45\columnwidth]{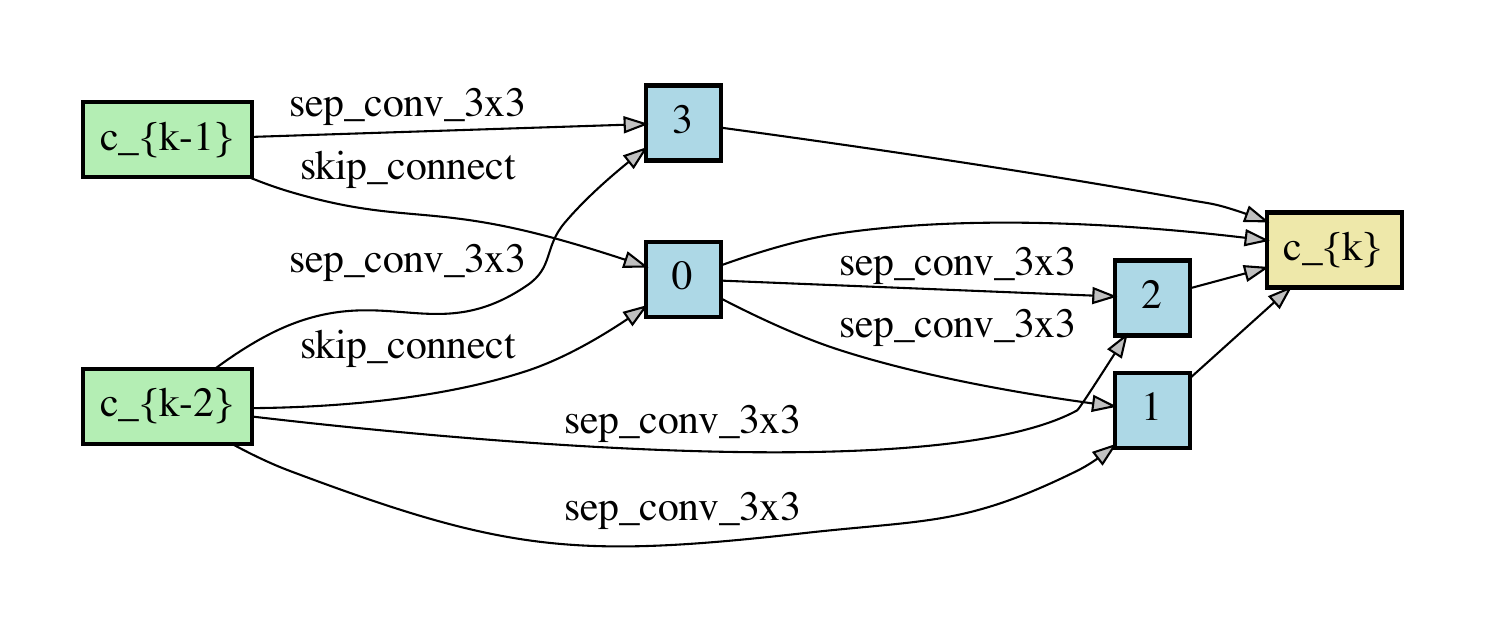}
		\includegraphics[width=0.45\columnwidth]{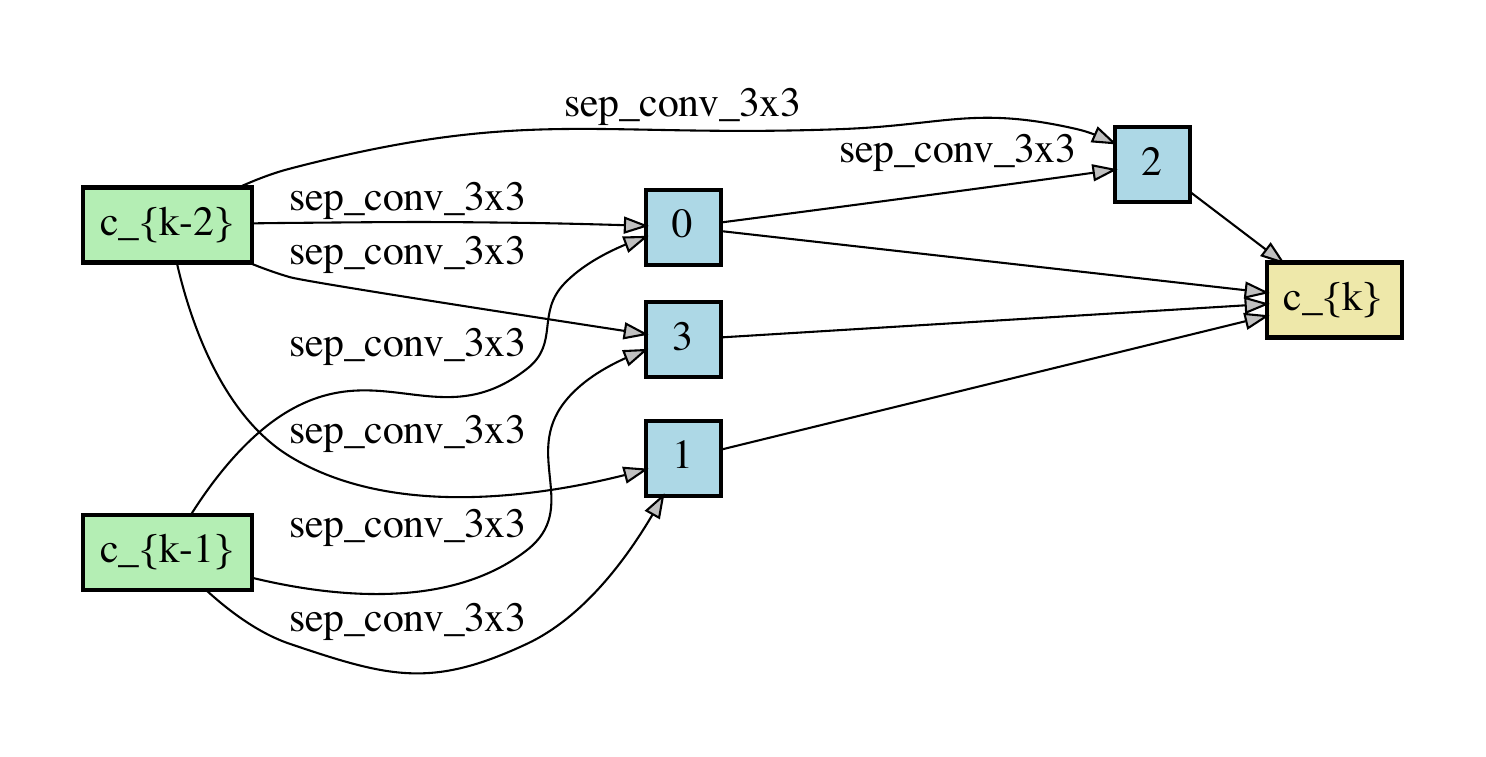}
		\caption{S2}
		\label{fig:rome-v1-cifar100-rdarts-ss-s2}
	\end{subfigure}
	
	\begin{subfigure}{0.98\columnwidth}
		\includegraphics[width=0.45\columnwidth]{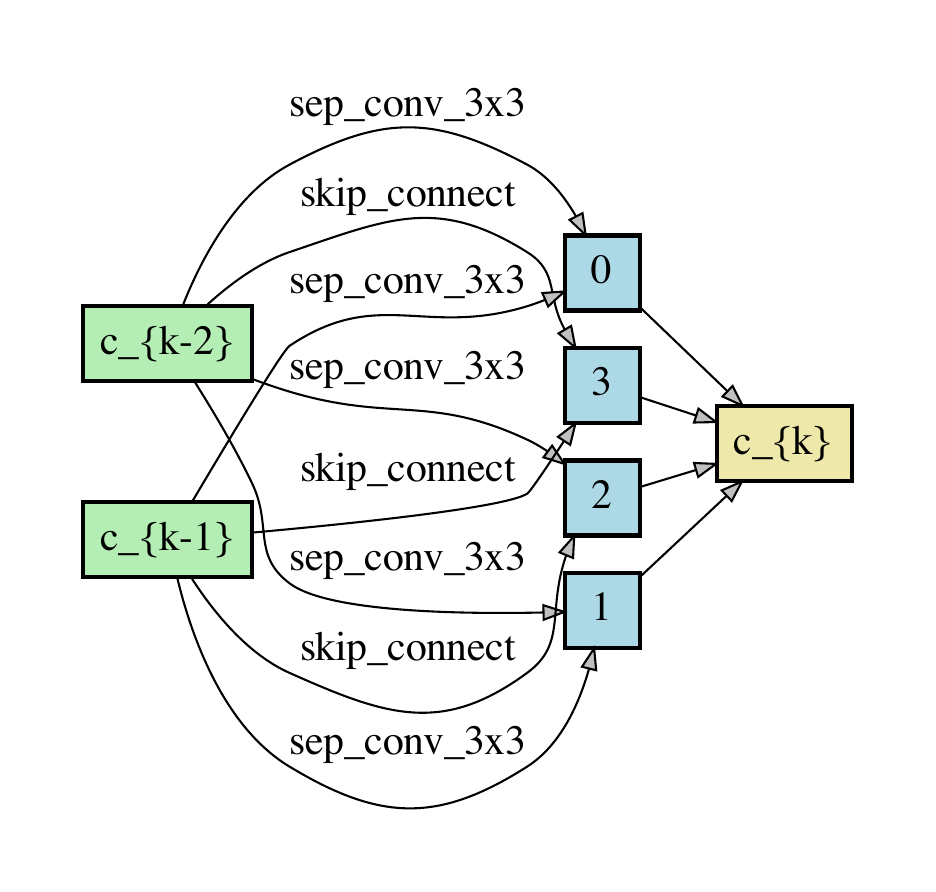}
		\includegraphics[width=0.45\columnwidth]{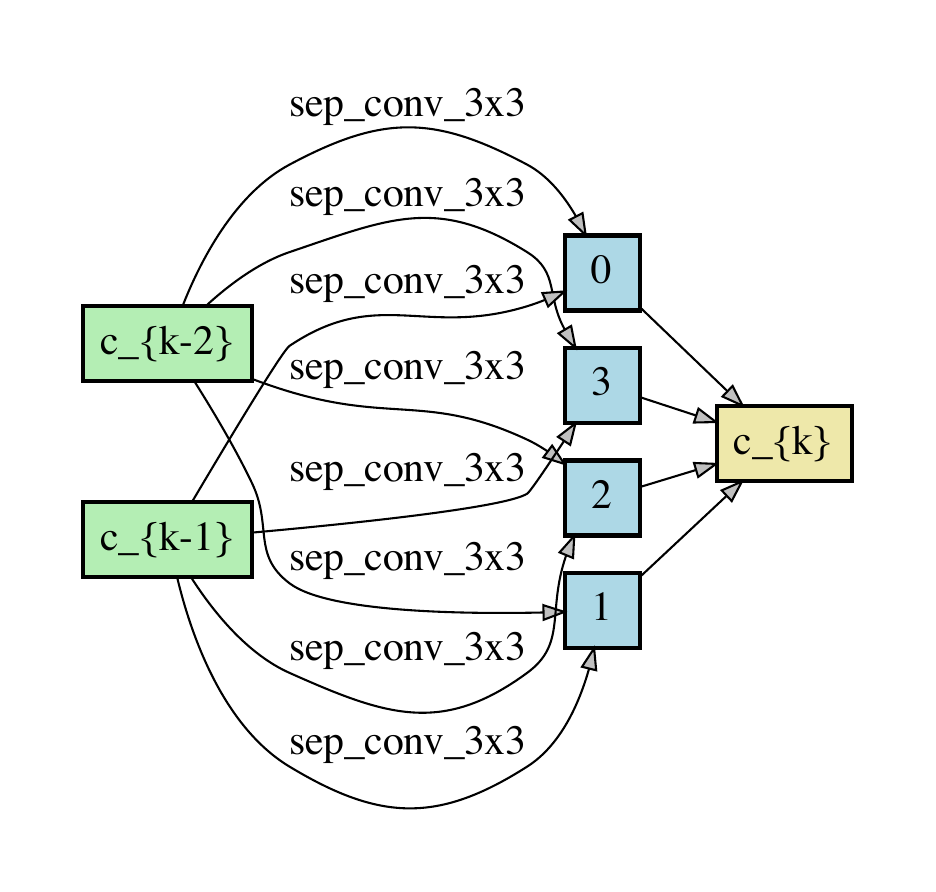}
		\caption{S3}
		\label{fig:rome-v1-cifar100-rdarts-ss-s3}
	\end{subfigure}
	
	\begin{subfigure}{0.98\columnwidth}
		\includegraphics[width=0.45\columnwidth]{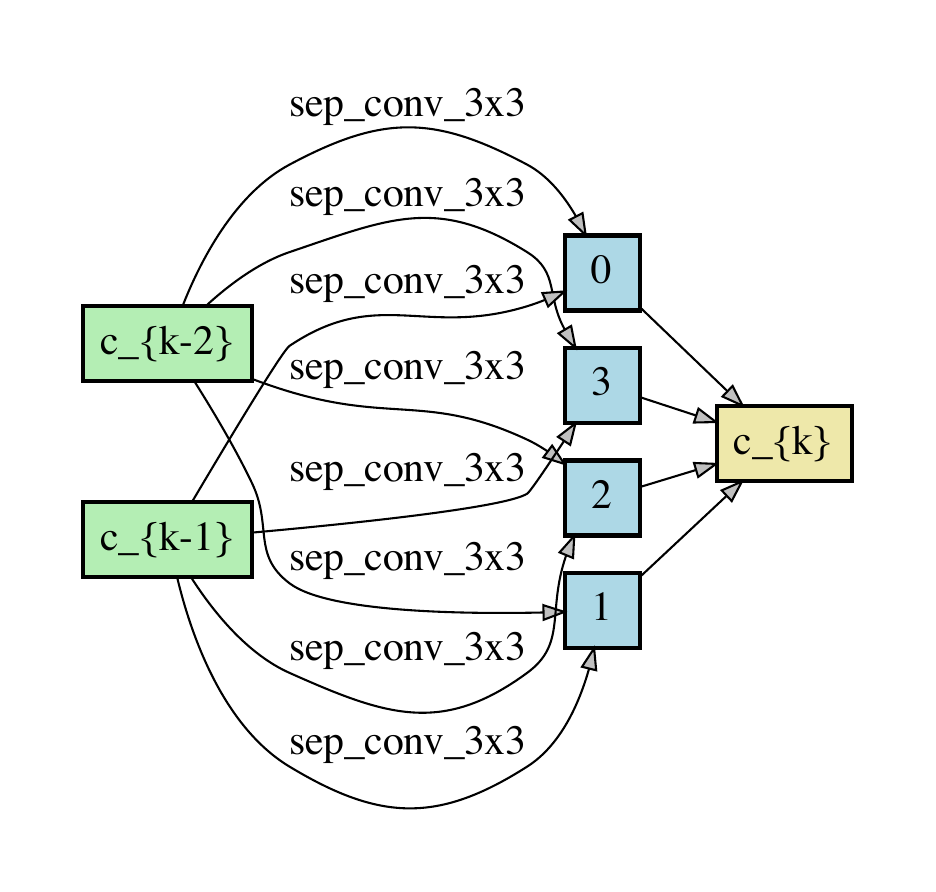}
		\includegraphics[width=0.45\columnwidth]{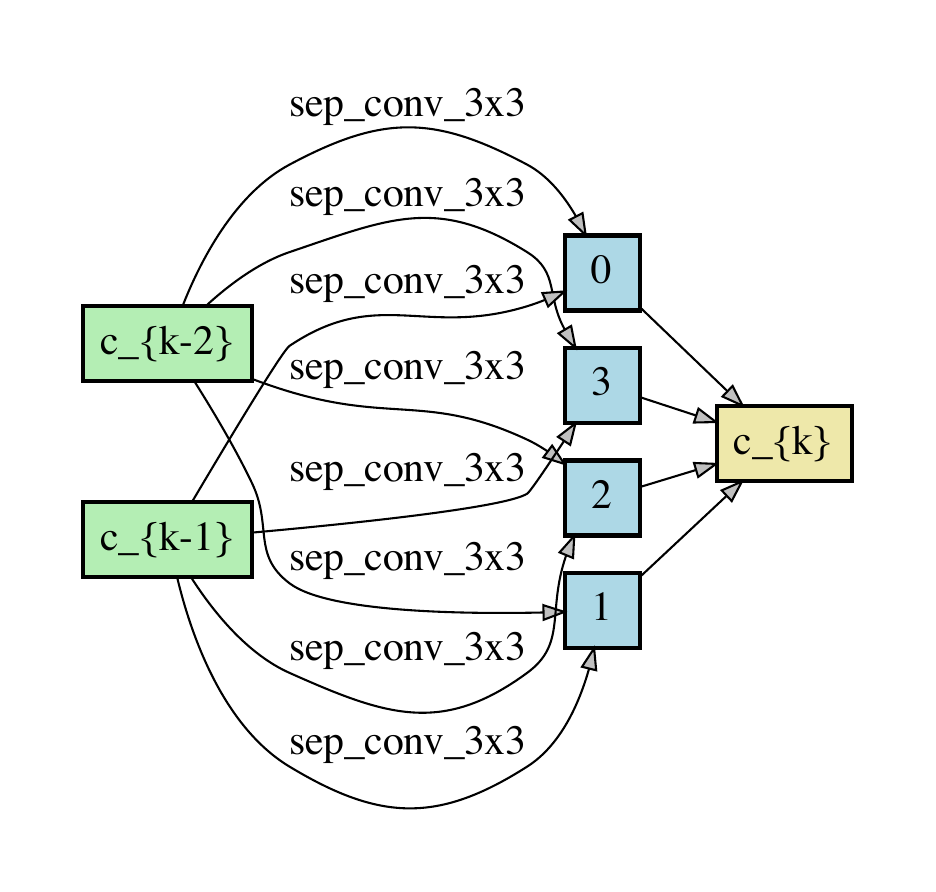}
		\caption{S4}
		\label{fig:rome-v1-cifar100-rdarts-ss-s4}
	\end{subfigure}
	
	\caption{ROME-V1 best cells (paired in normal and reduction) on CIFAR100 in reduced search spaces of RobustDARTS.}
	\label{fig:rome-v1-cifar100-rdarts-ss}
\end{figure}

\begin{figure}[ht]
	\centering
	\begin{subfigure}{0.98\columnwidth}
		\includegraphics[width=0.45\columnwidth]{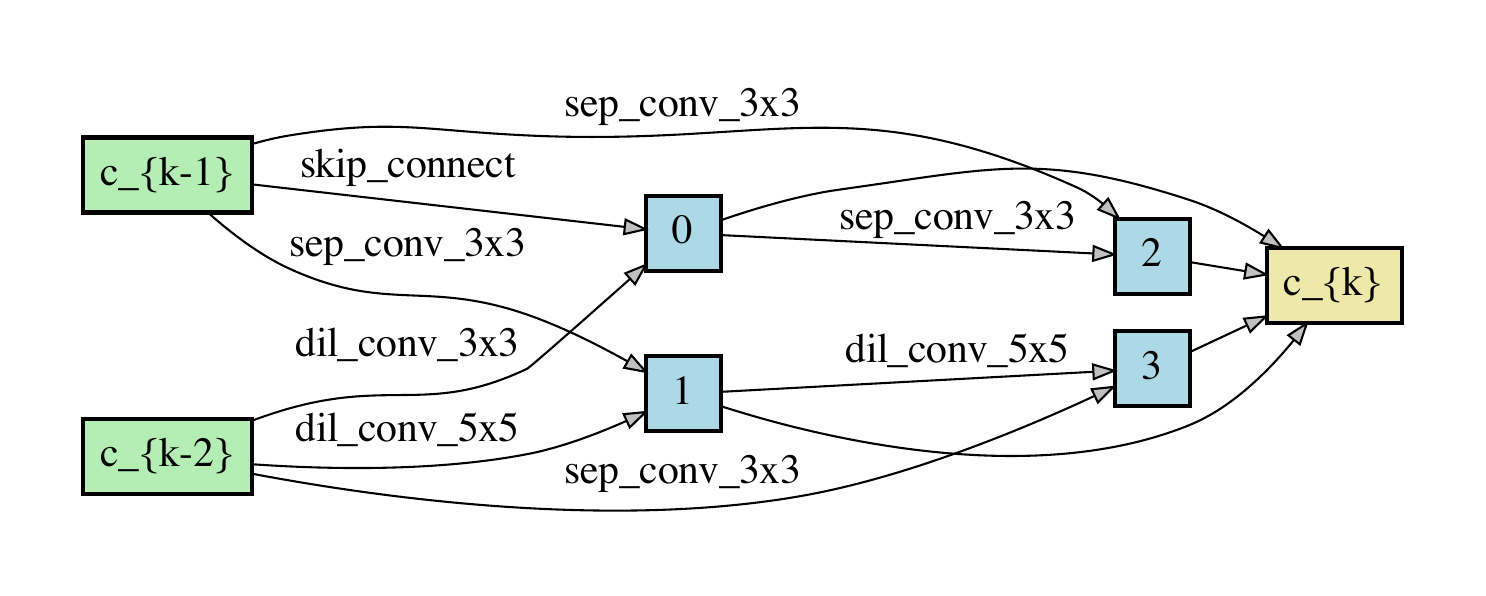}
		\includegraphics[width=0.45\columnwidth]{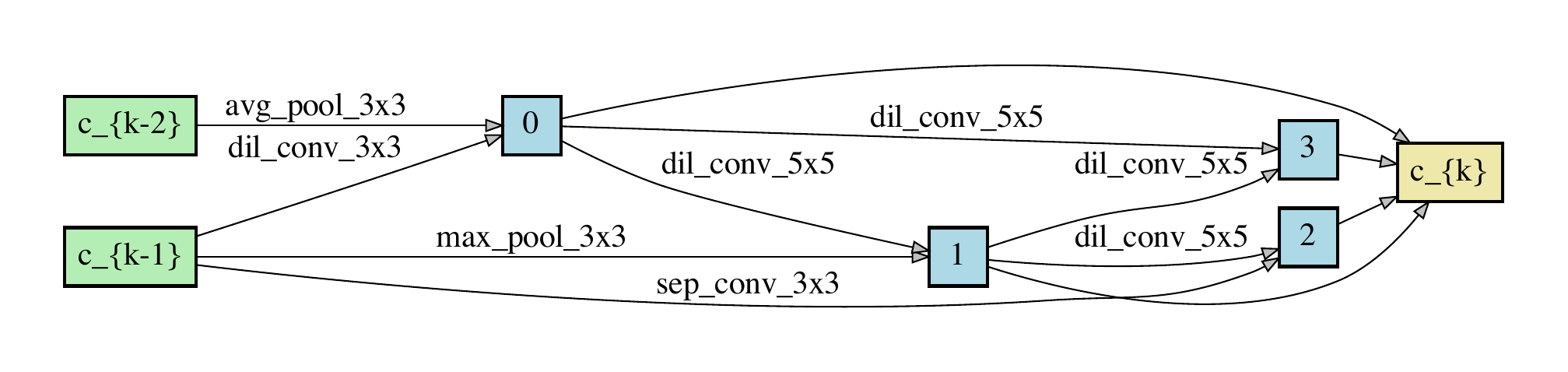}
		\caption{S1}
		\label{fig:rome-v1-svhn-rdarts-ss-s1}
	\end{subfigure}
	
	\begin{subfigure}{0.98\columnwidth}
		\includegraphics[width=0.45\columnwidth]{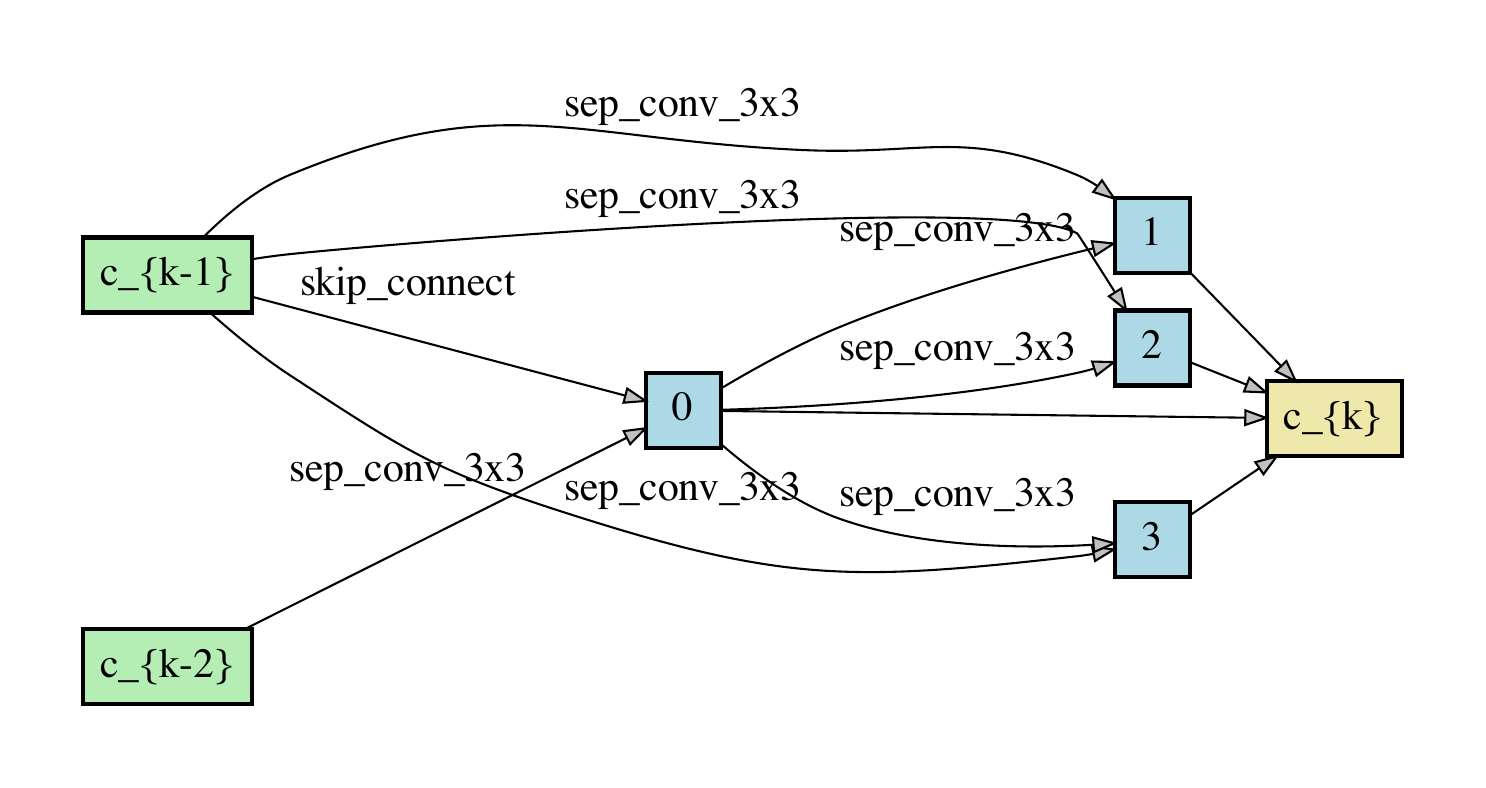}
		\includegraphics[width=0.45\columnwidth]{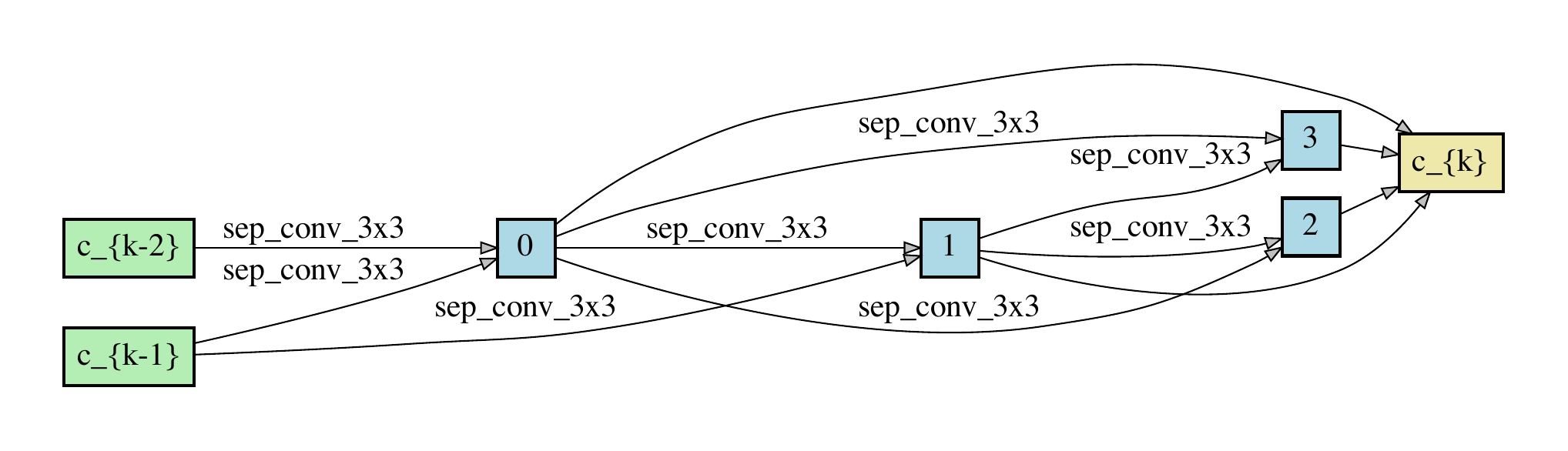}
		\caption{S2}
		\label{fig:rome-v1-svhn-rdarts-ss-s2}
	\end{subfigure}
	
	\begin{subfigure}{0.98\columnwidth}
		\includegraphics[width=0.45\columnwidth]{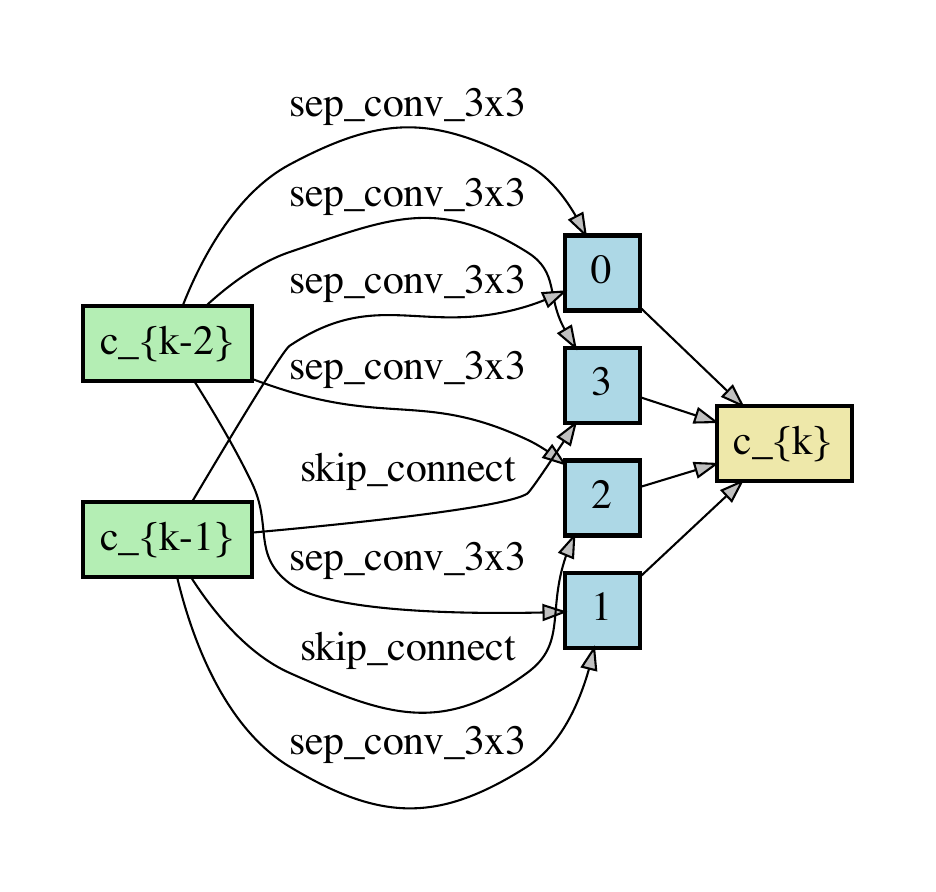}
		\includegraphics[width=0.45\columnwidth]{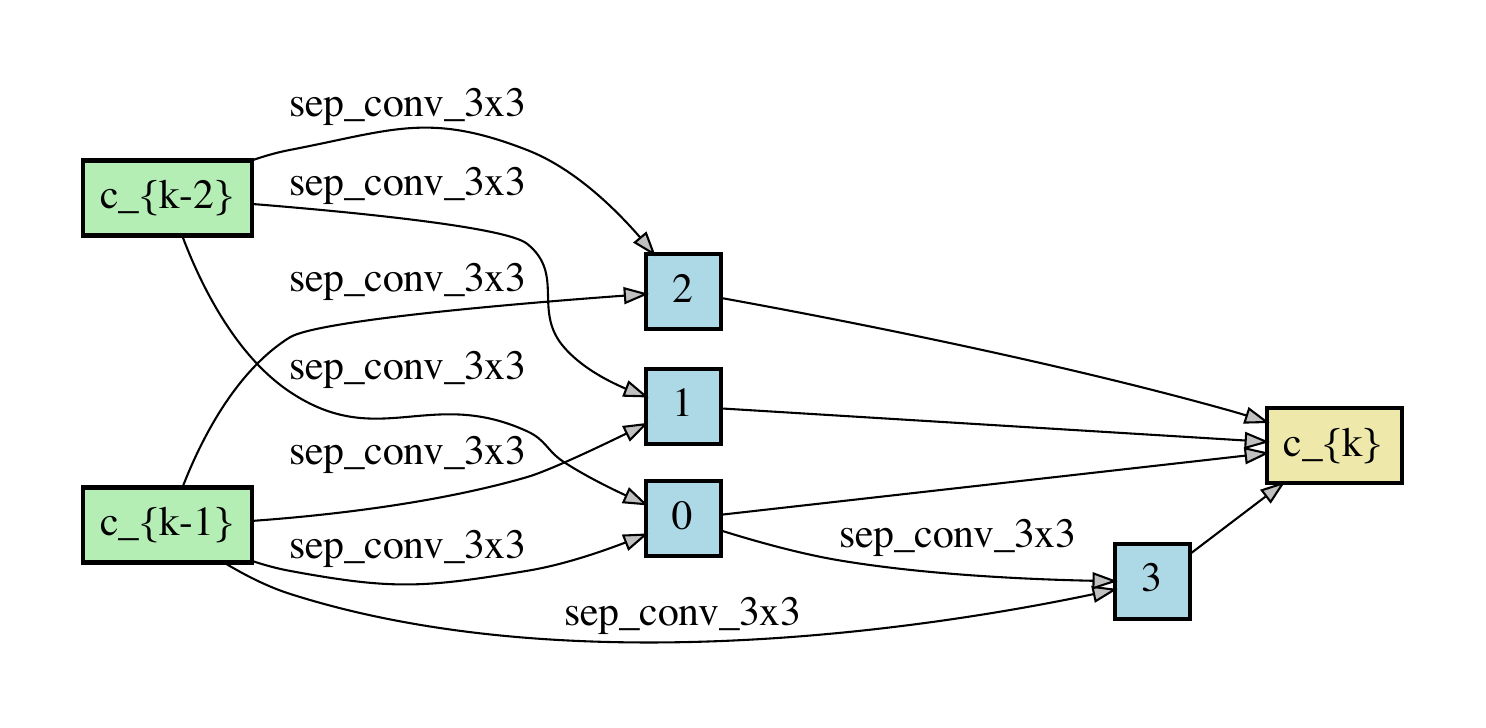}
		\caption{S3}
		\label{fig:rome-v1-svhn-rdarts-ss-s3}
	\end{subfigure}
	
	\begin{subfigure}{0.98\columnwidth}
		\includegraphics[width=0.45\columnwidth]{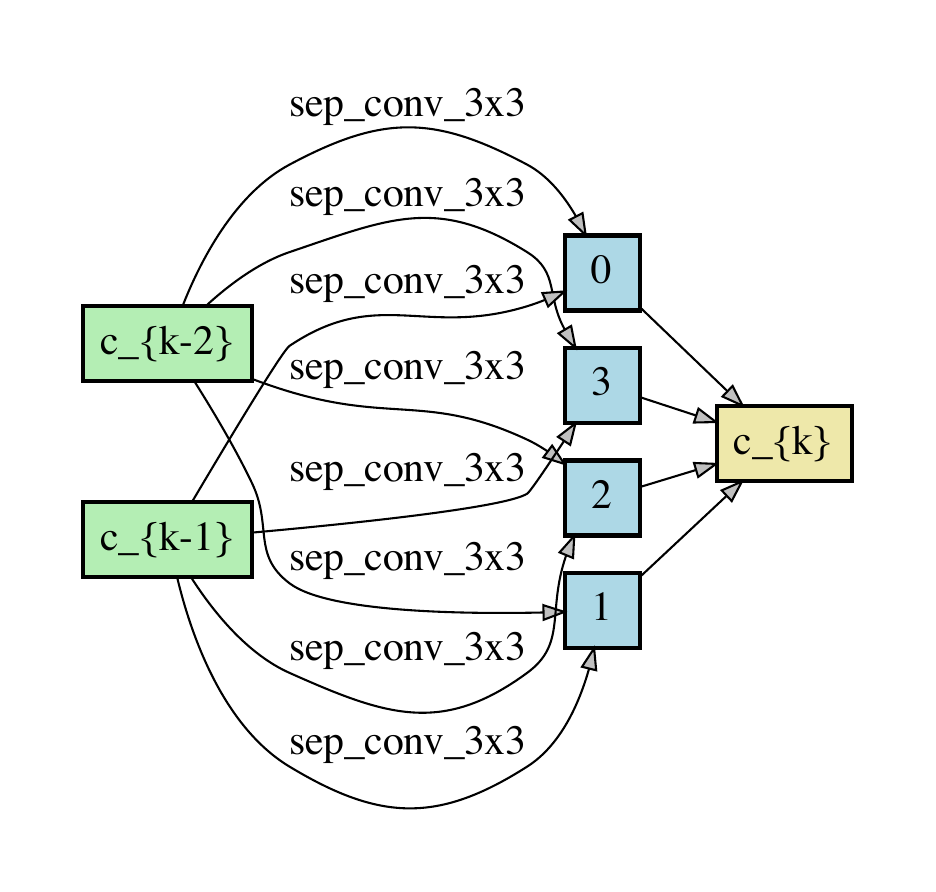}
		\includegraphics[width=0.45\columnwidth]{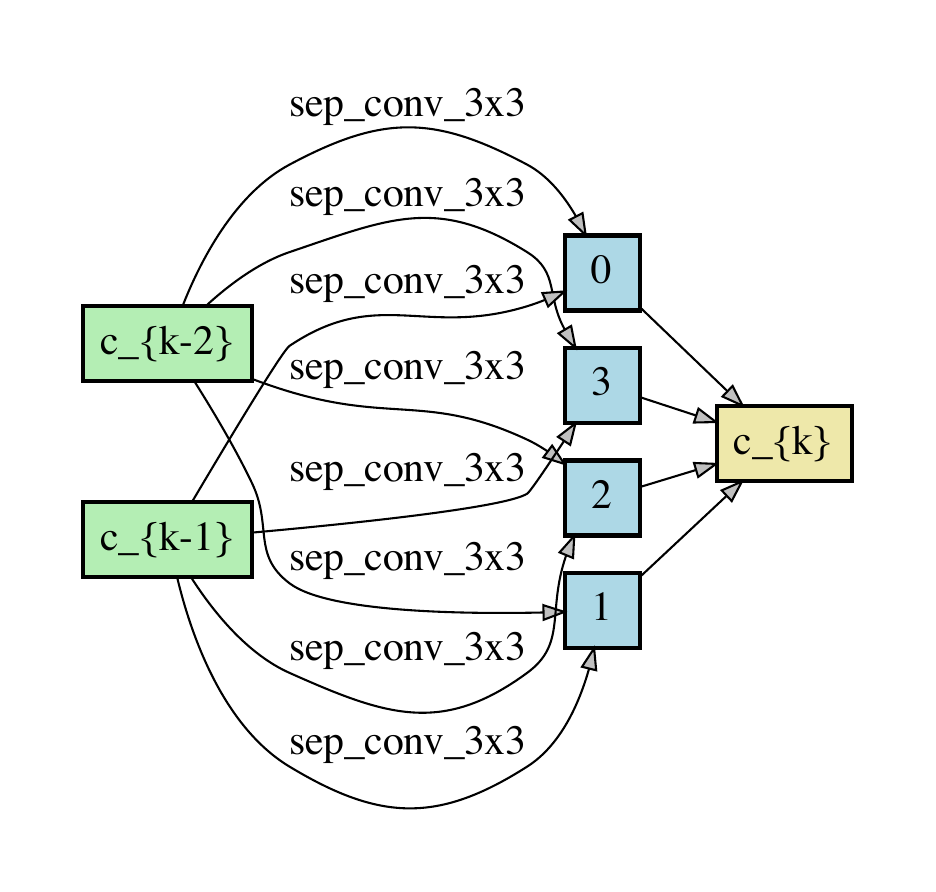}
		\caption{S4}
		\label{fig:rome-v1-svhn-rdarts-ss-s4}
	\end{subfigure}
	
	\caption{ROME-V1 best cells (paired in normal and reduction) on SVHN in reduced search spaces of RobustDARTS.}
	\label{fig:rome-v1-svhn-rdarts-ss}
\end{figure}

\begin{figure}[ht]
	\centering
	\begin{subfigure}{0.98\columnwidth}
		\includegraphics[width=0.45\columnwidth]{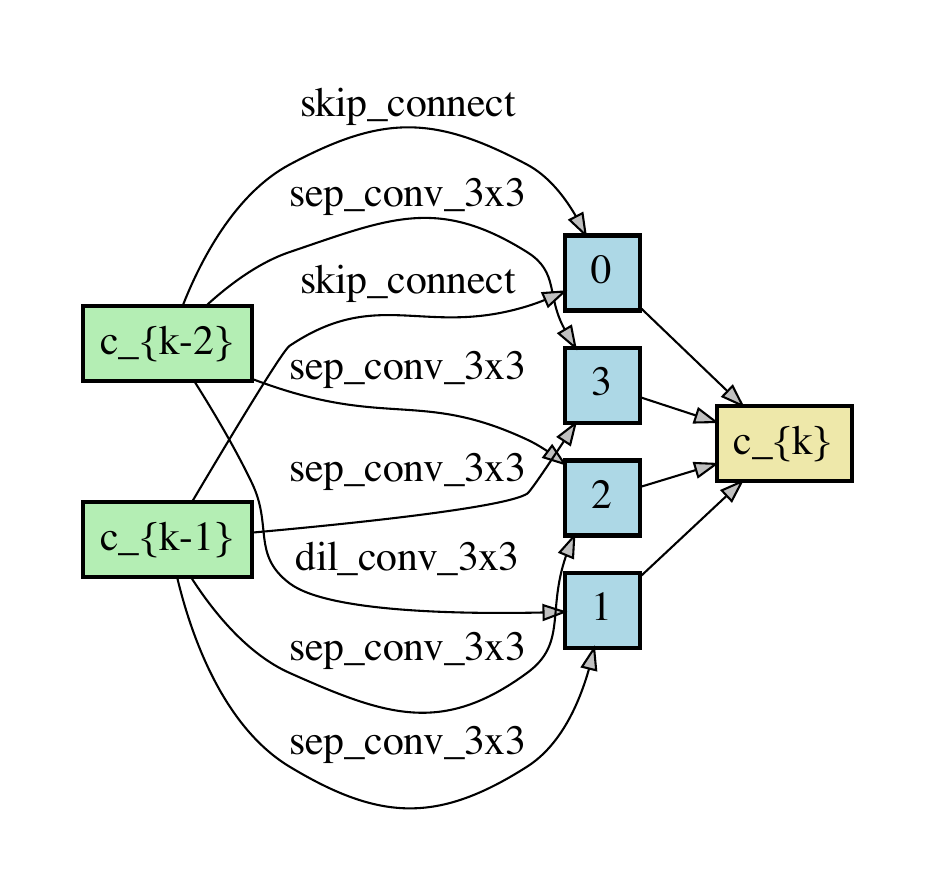}
		\includegraphics[width=0.45\columnwidth]{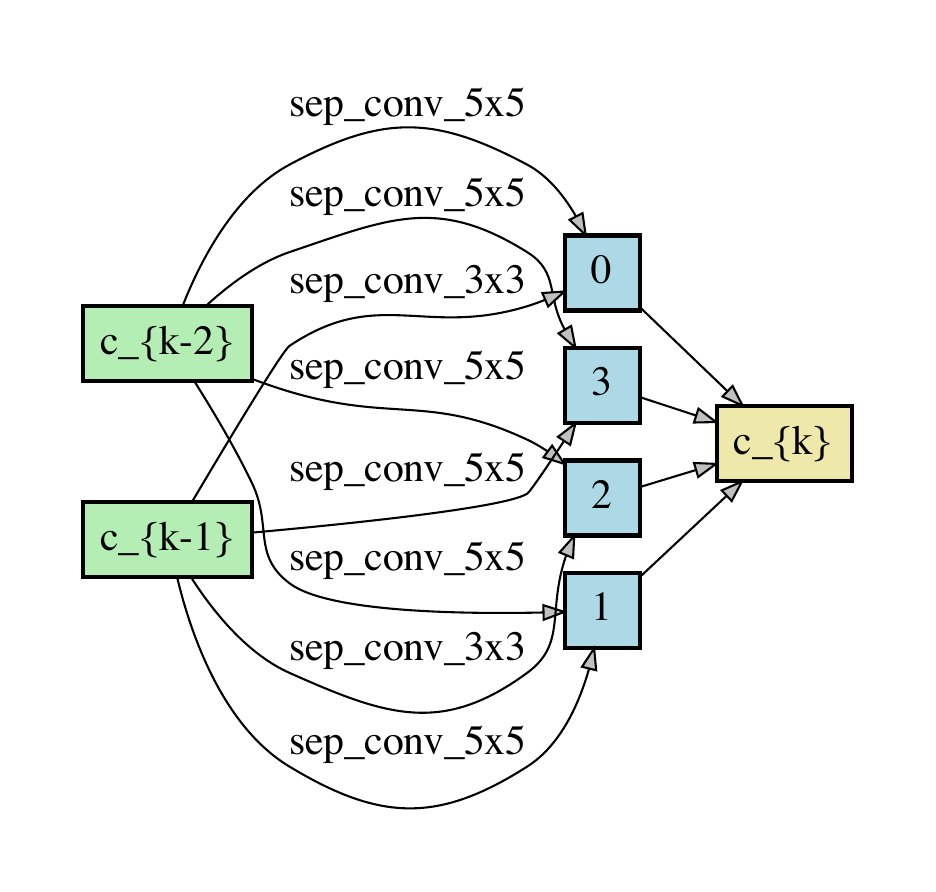}
		\caption{Architecture 1}
		\label{fig:rome-v1-cifar100-s0-1}
	\end{subfigure}
	
	\begin{subfigure}{0.98\columnwidth}
		\includegraphics[width=0.45\columnwidth]{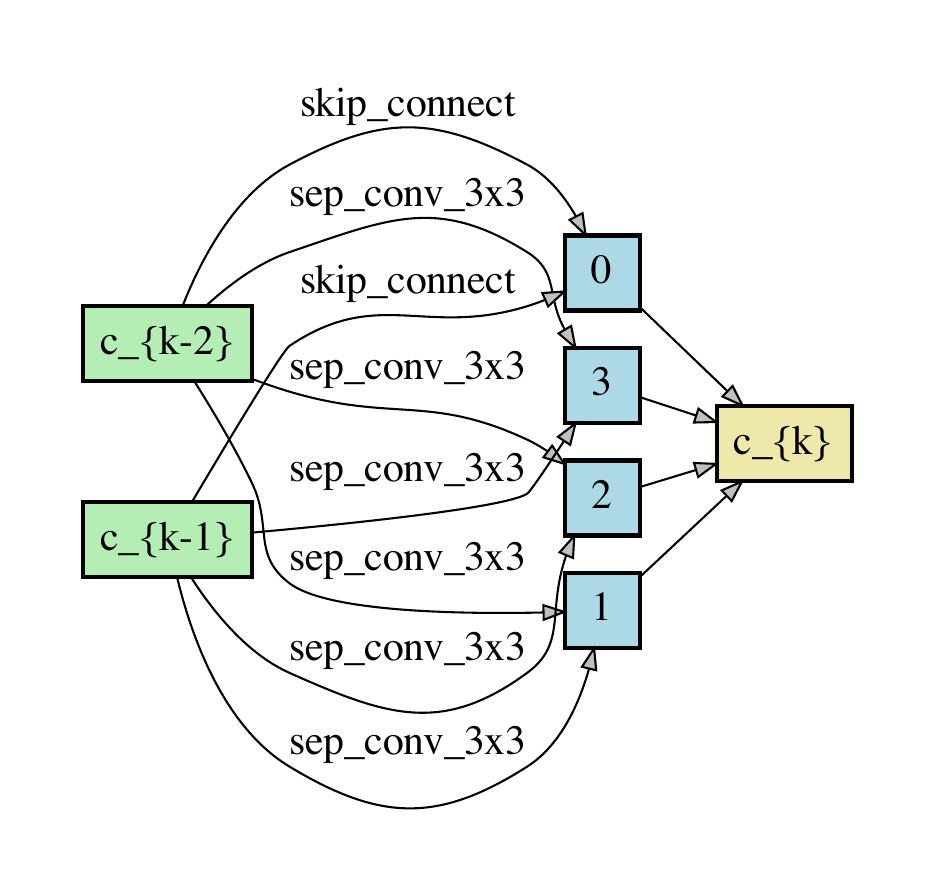}
		\includegraphics[width=0.45\columnwidth]{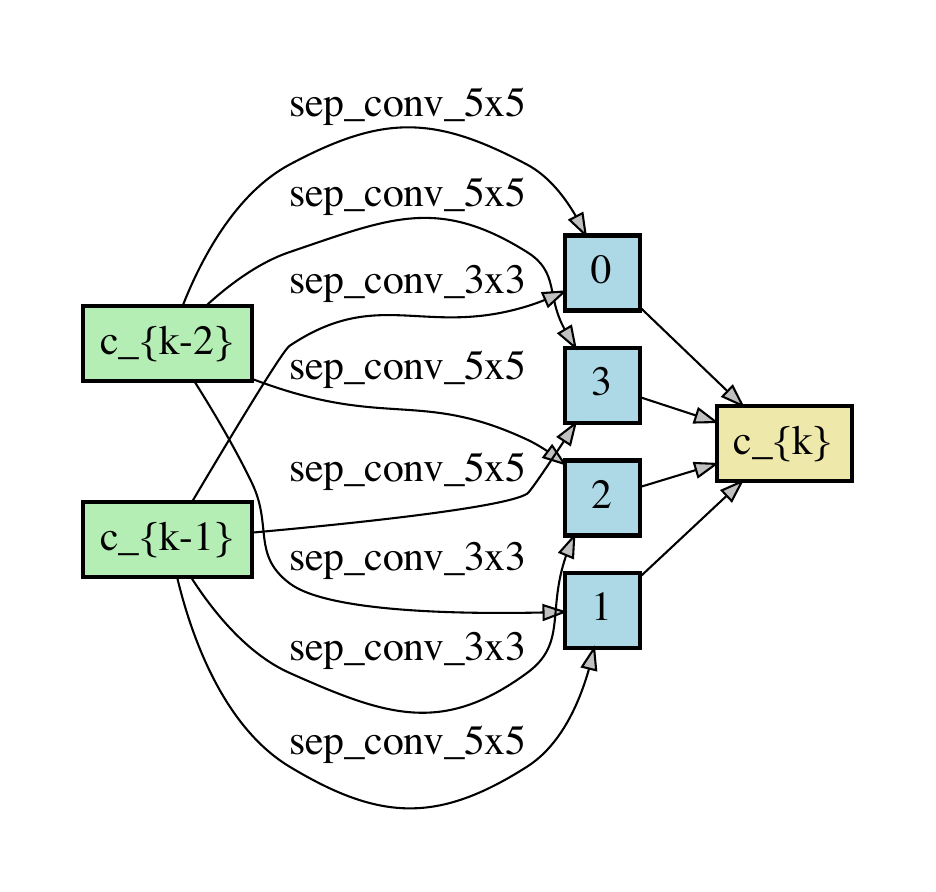}
		\caption{Architecture 2}
		\label{fig:rome-v1-cifar100-s0-2}
	\end{subfigure}
	
	\begin{subfigure}{0.98\columnwidth}
		\includegraphics[width=0.45\columnwidth]{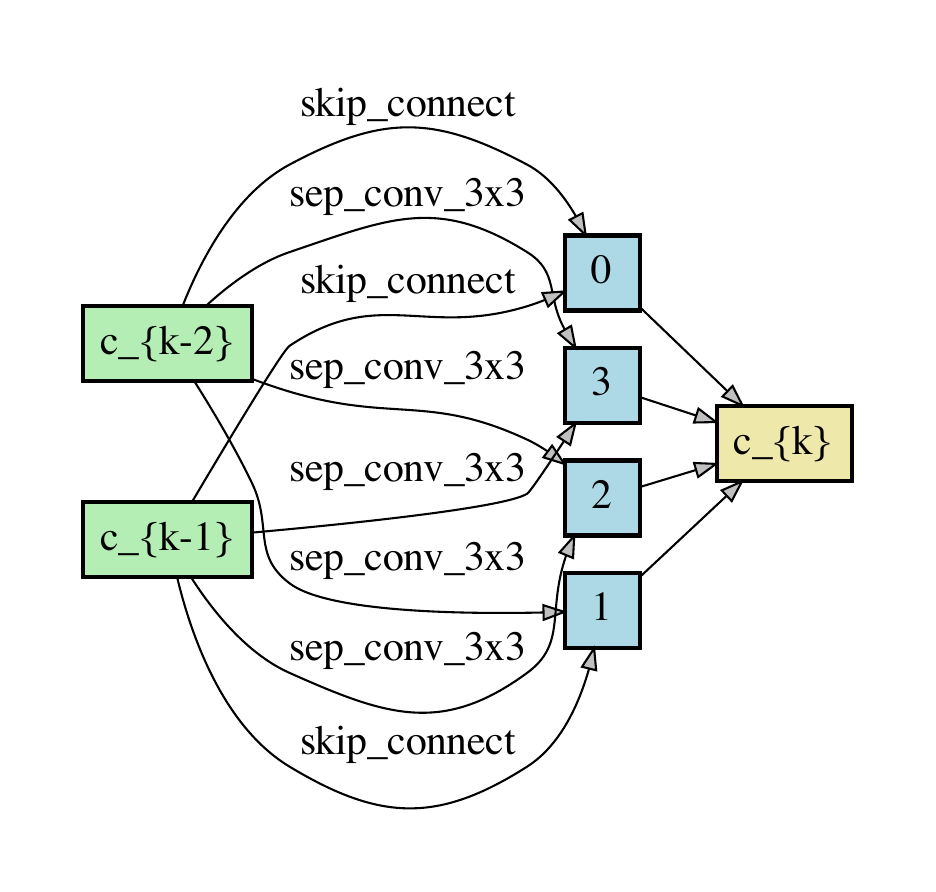}
		\includegraphics[width=0.45\columnwidth]{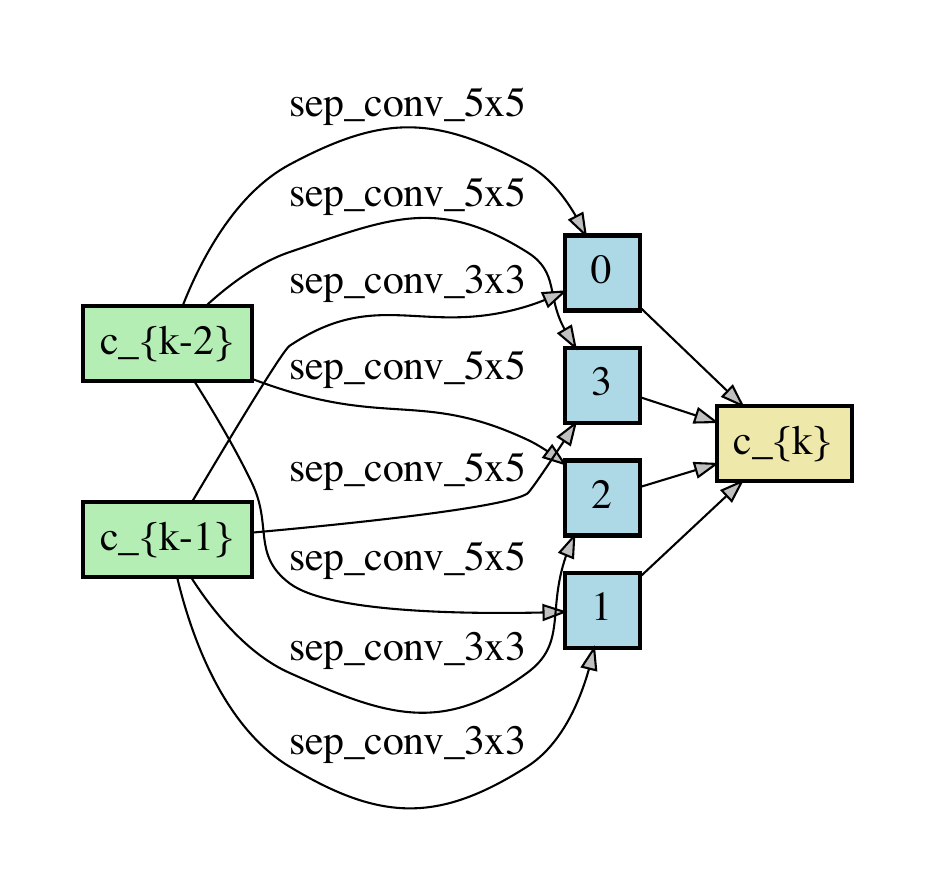}
		\caption{Architecture 3}
		\label{fig:rome-v1-cifar100-s0-3}
	\end{subfigure}
	
	\caption{ROME-V1 cells (paired in normal and reduction) on CIFAR-100 in DARTS's search space.}
	\label{fig:rome-v1-cifar100-s0}
\end{figure}

\begin{figure}[ht]
	\centering
	\begin{subfigure}{0.98\columnwidth}
		\includegraphics[width=0.45\columnwidth]{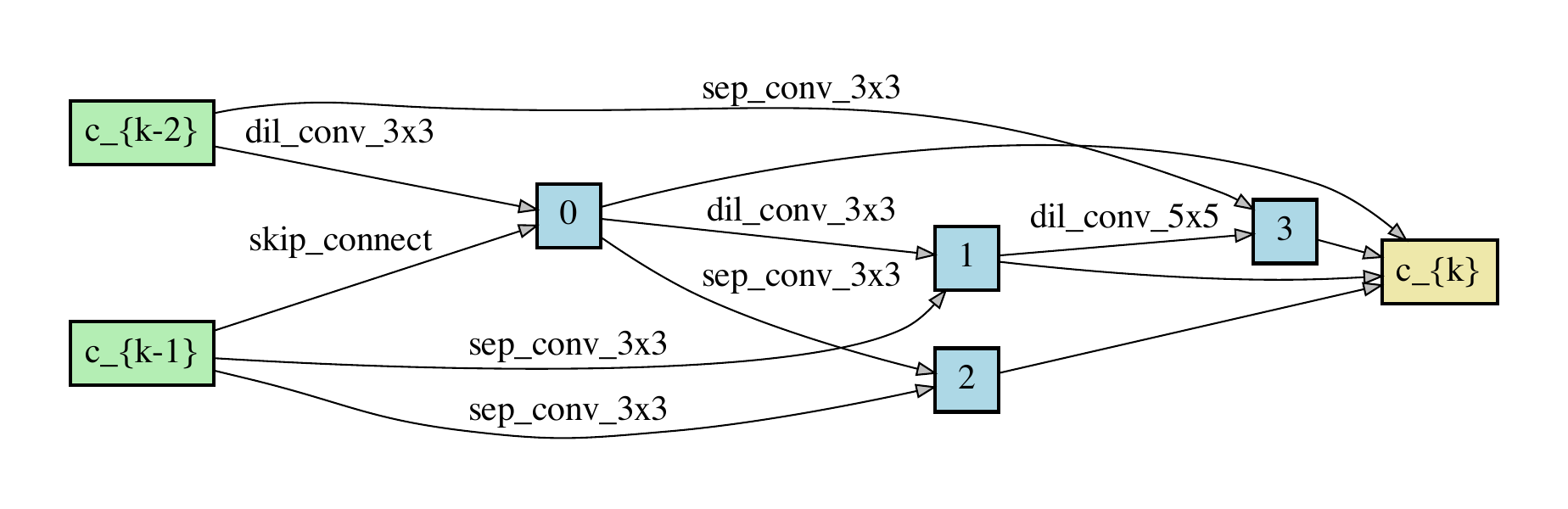}
		\includegraphics[width=0.45\columnwidth]{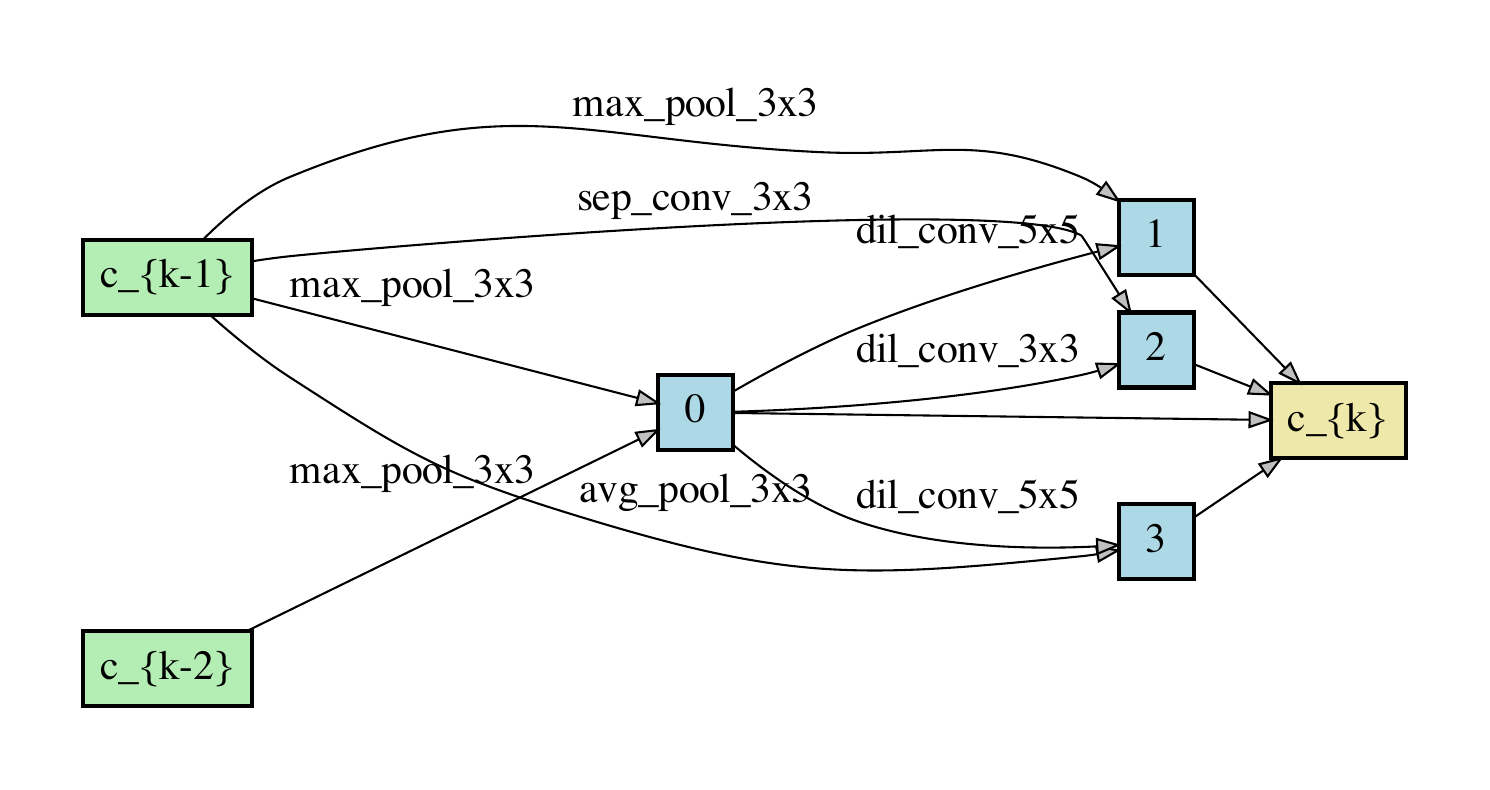}
		\caption{S1}
		\label{fig:rome-v2-cifar10-rdarts-ss-s1}
	\end{subfigure}
	
	\begin{subfigure}{0.98\columnwidth}
		\includegraphics[width=0.45\columnwidth]{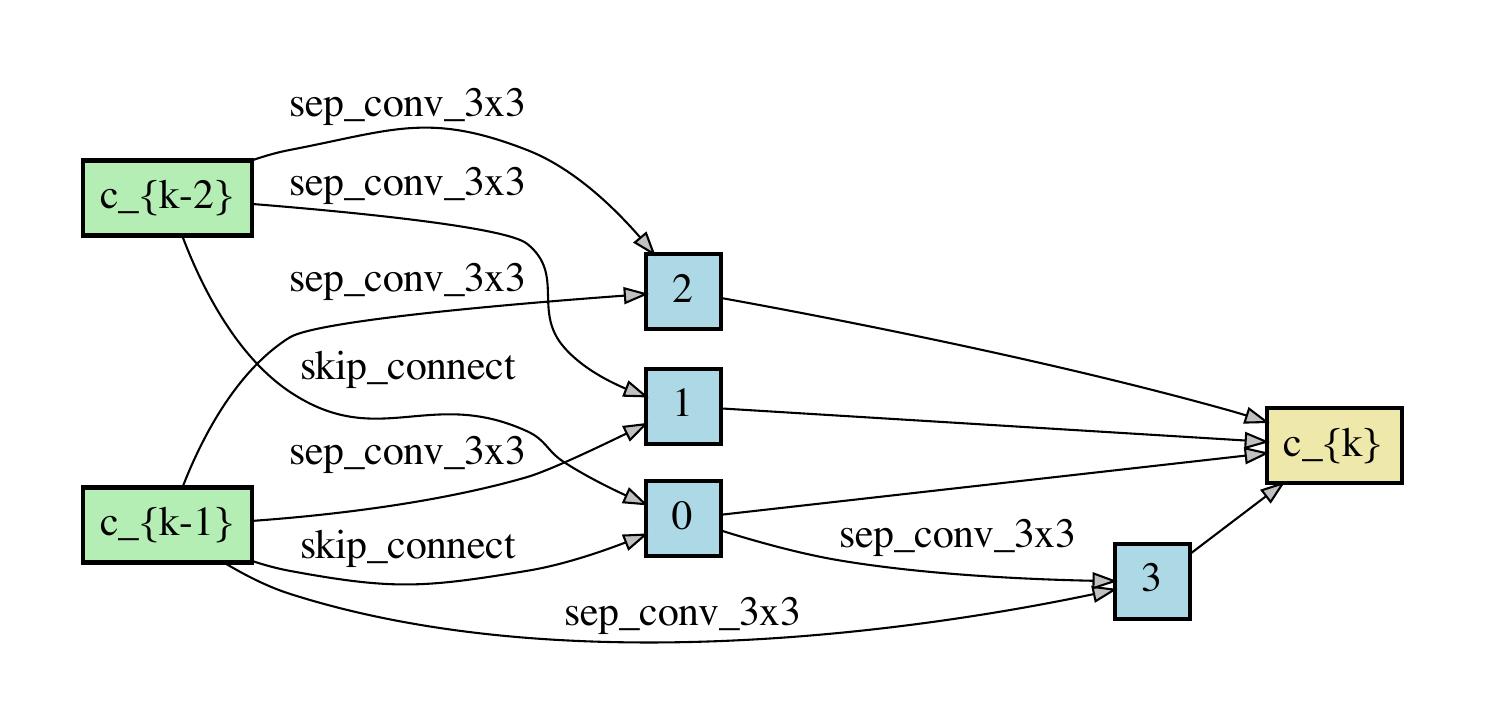}
		\includegraphics[width=0.45\columnwidth]{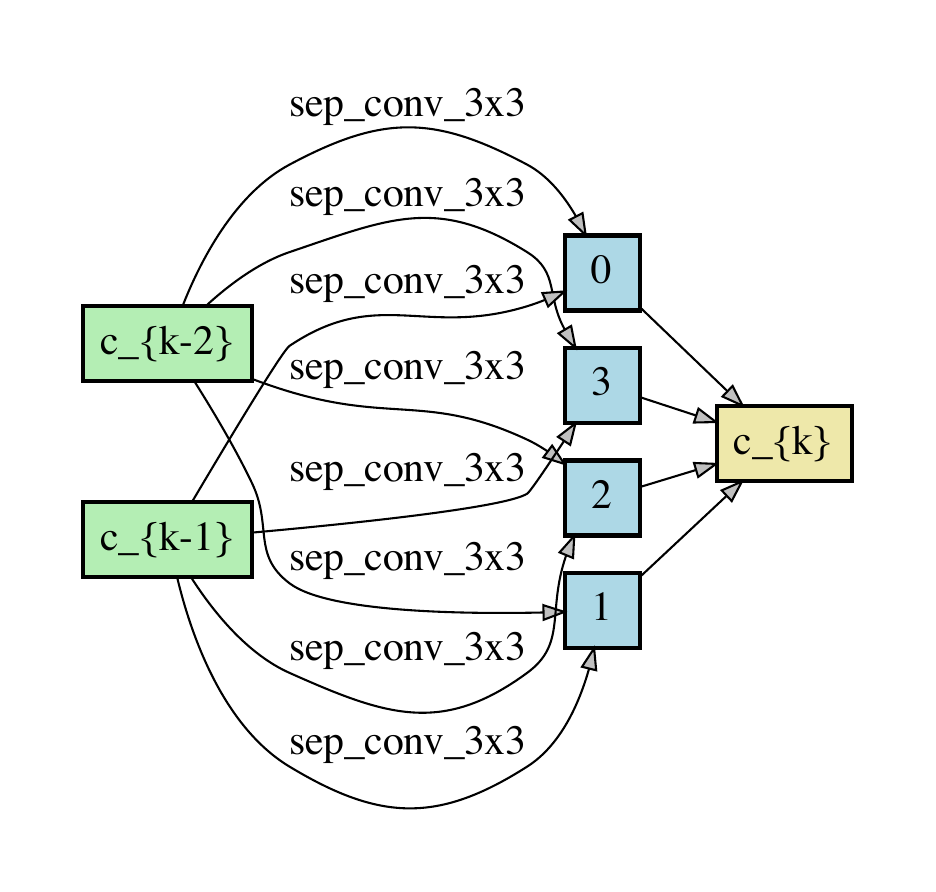}
		\caption{S2}
		\label{fig:rome-v2-cifar10-rdarts-ss-s2}
	\end{subfigure}
	
	\begin{subfigure}{0.98\columnwidth}
		\includegraphics[width=0.45\columnwidth]{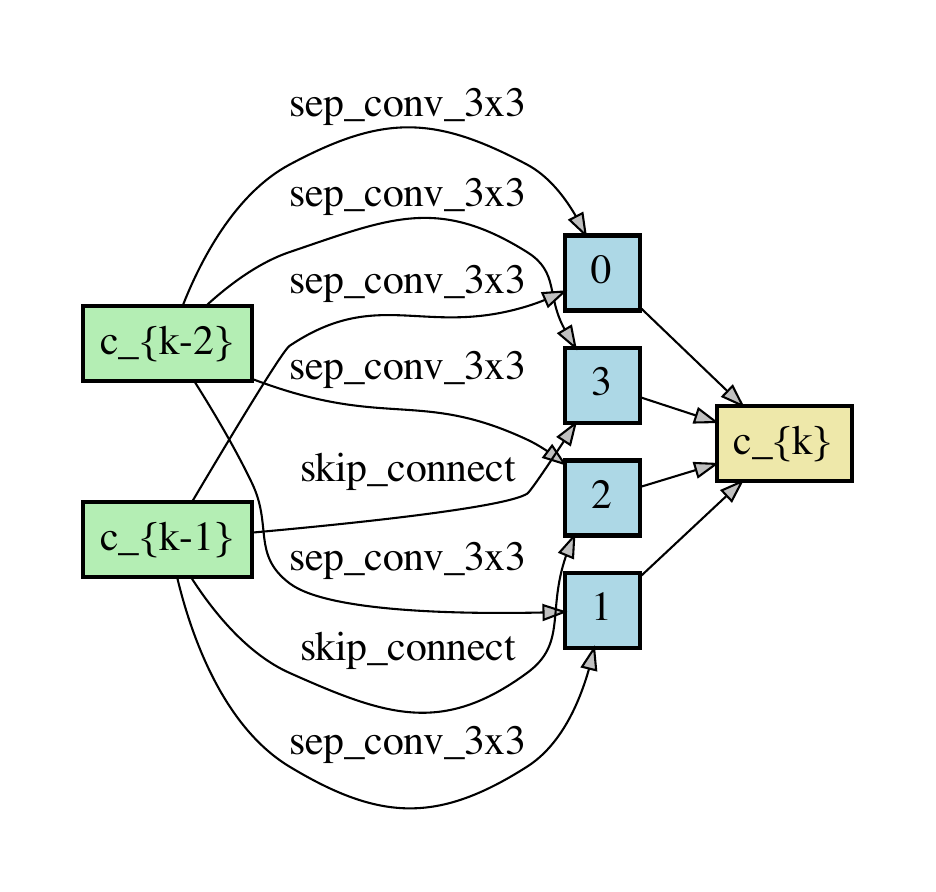}
		\includegraphics[width=0.45\columnwidth]{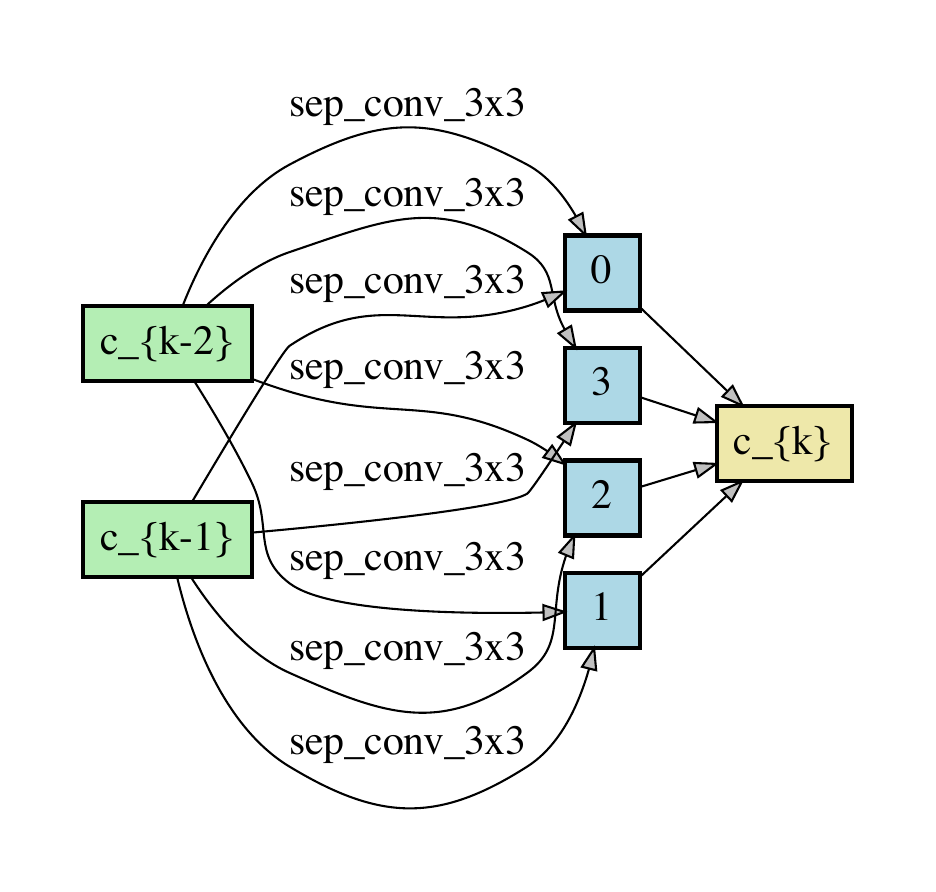}
		\caption{S3}
		\label{fig:rome-v2-cifar10-rdarts-ss-s3}
	\end{subfigure}
	
	\begin{subfigure}{0.98\columnwidth}
		\includegraphics[width=0.45\columnwidth]{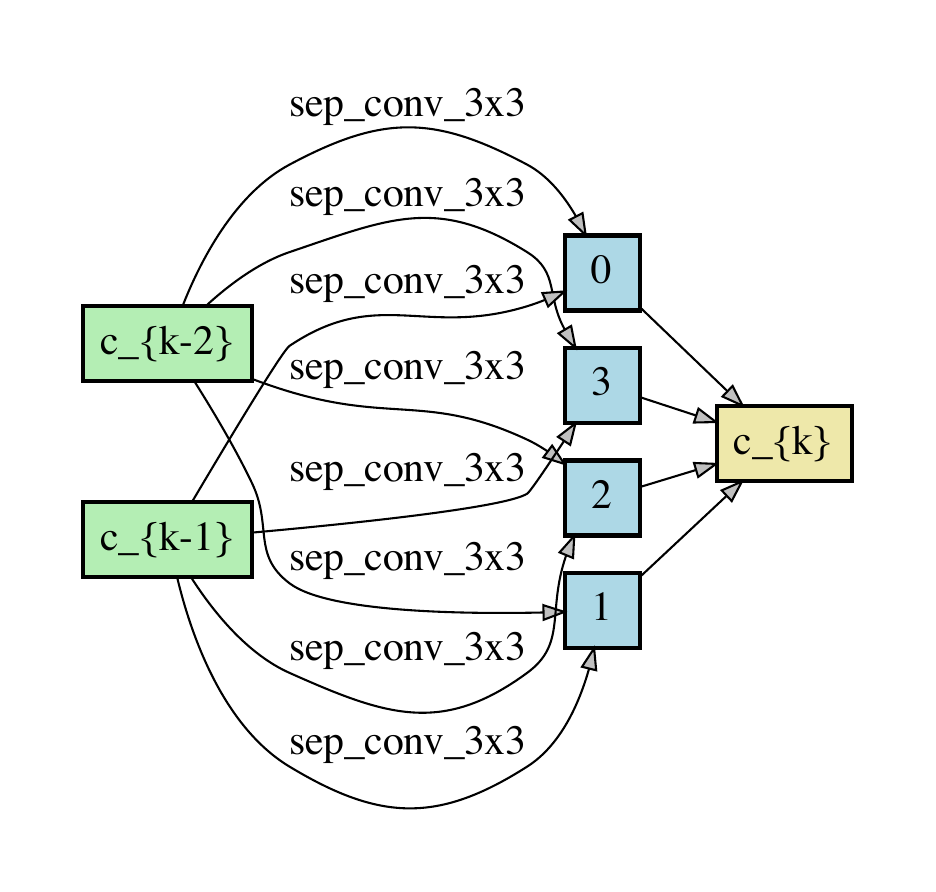}
		\includegraphics[width=0.45\columnwidth]{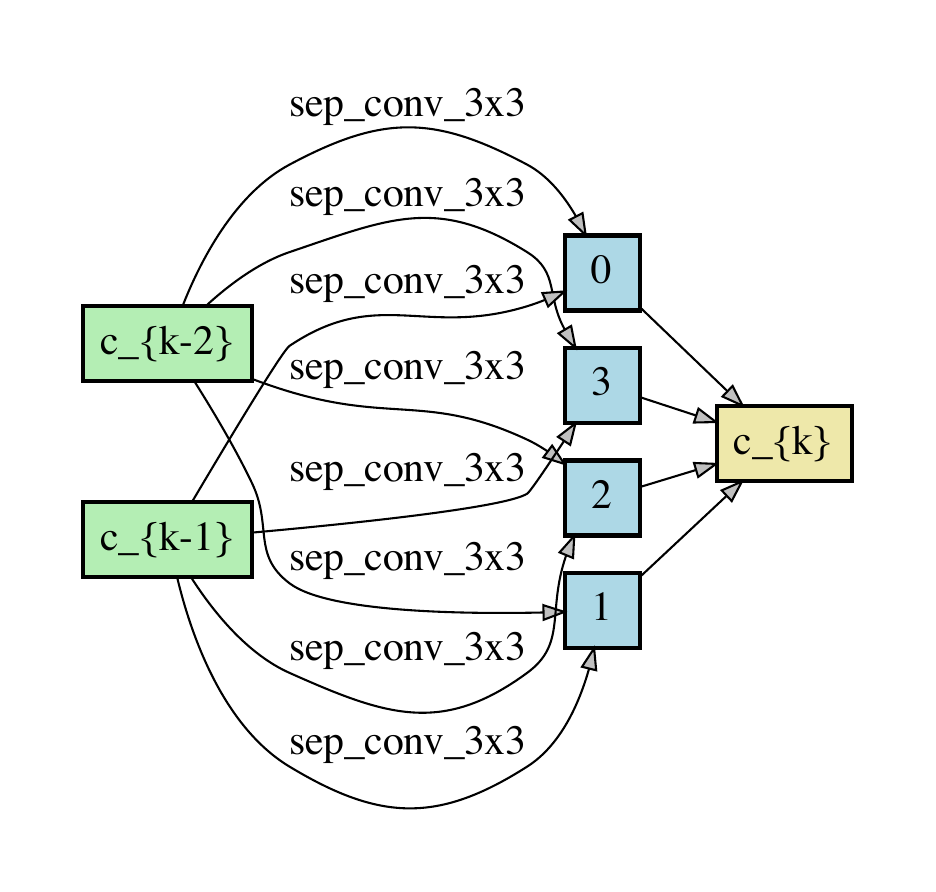}
		\caption{S4}
		\label{fig:rome-v2-cifar10-rdarts-ss-s4}
	\end{subfigure}
	
	\caption{ROME-V2 best cells (paired in normal and reduction) on CIFAR10 in reduced search spaces of RobustDARTS.}
	\label{fig:rome-v2-cifar10-rdarts-ss}
\end{figure}

\begin{figure}[ht]
	\centering
	\begin{subfigure}{0.98\columnwidth}
		\includegraphics[width=0.45\columnwidth]{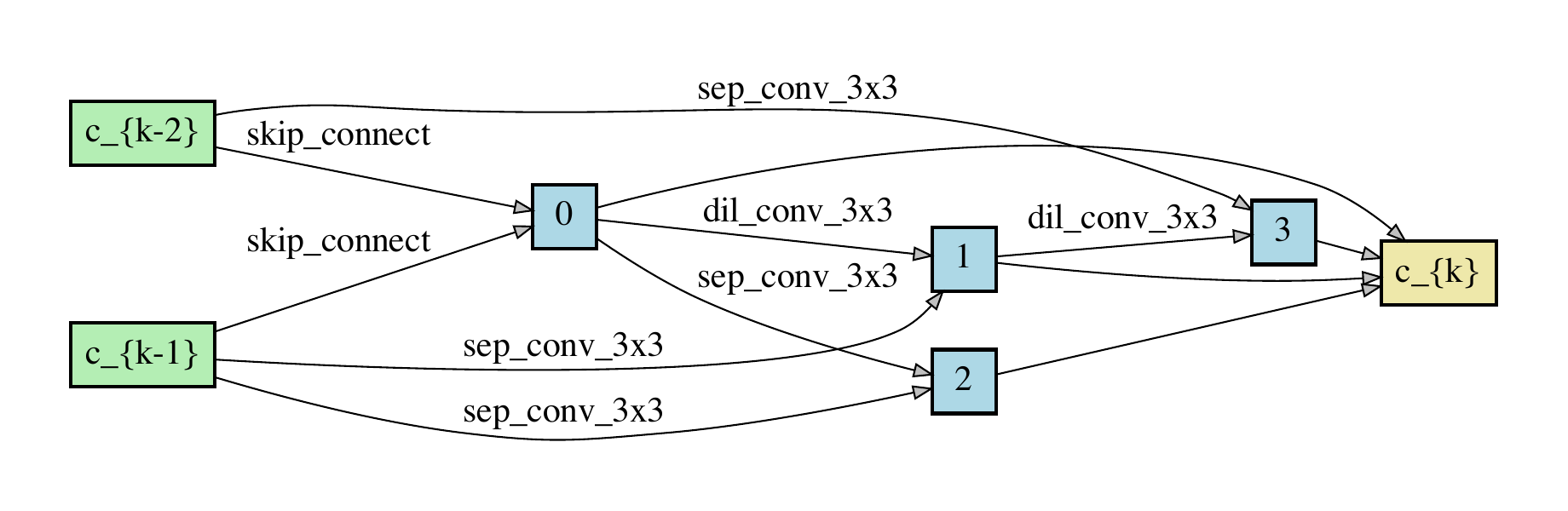}
		\includegraphics[width=0.45\columnwidth]{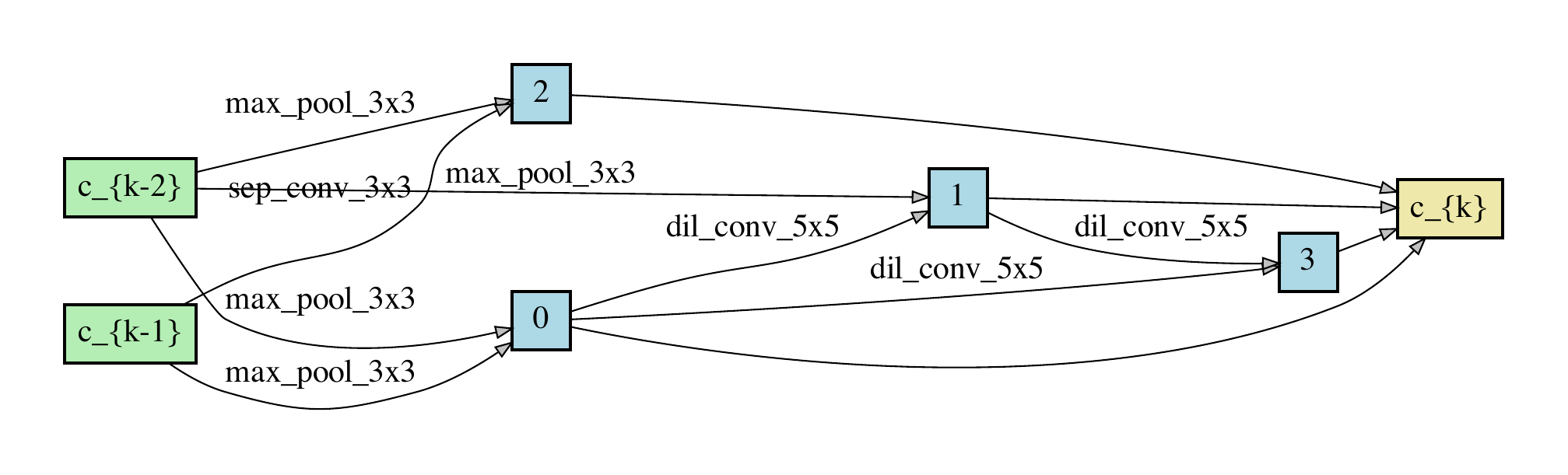}
		\caption{S1}
		\label{fig:rome-v2-cifar100-rdarts-ss-s1}
	\end{subfigure}
	
	\begin{subfigure}{0.98\columnwidth}
		\includegraphics[width=0.45\columnwidth]{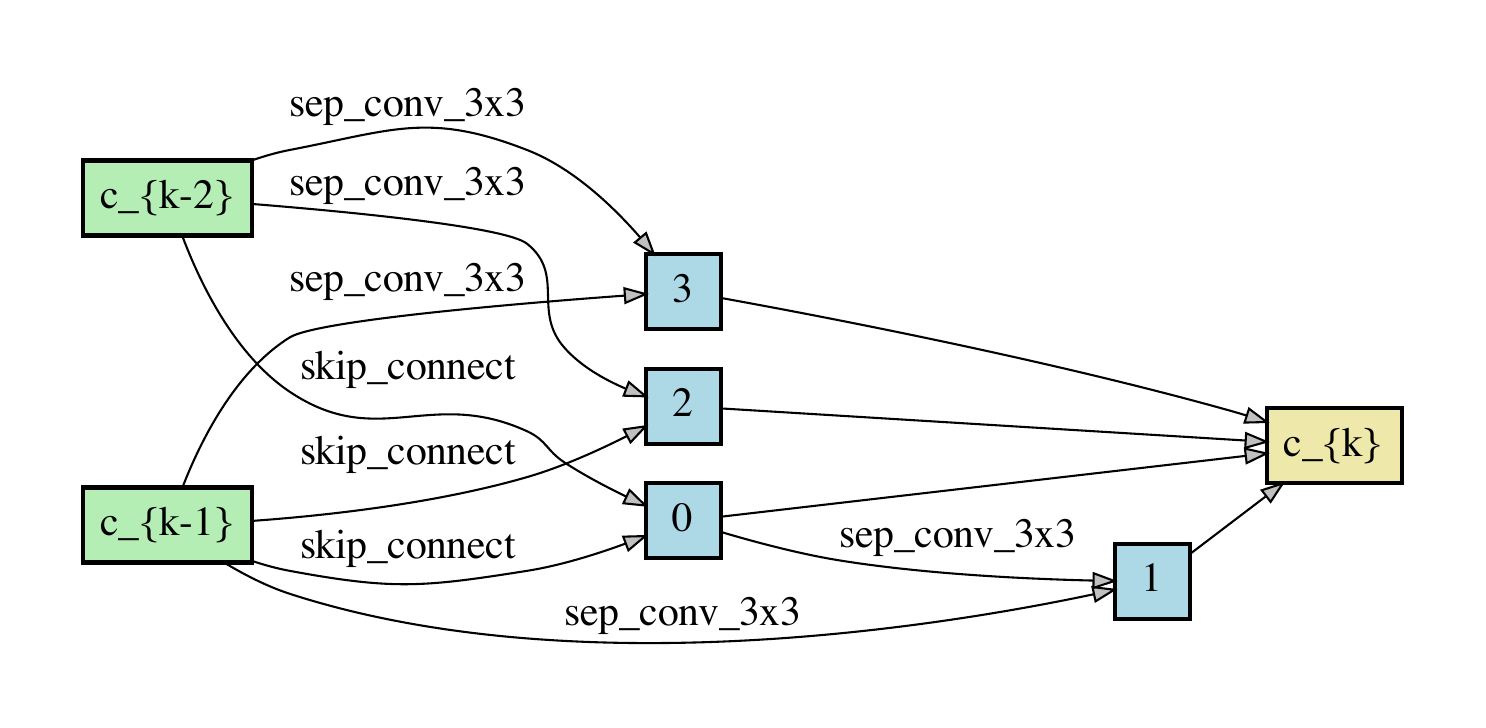}
		\includegraphics[width=0.45\columnwidth]{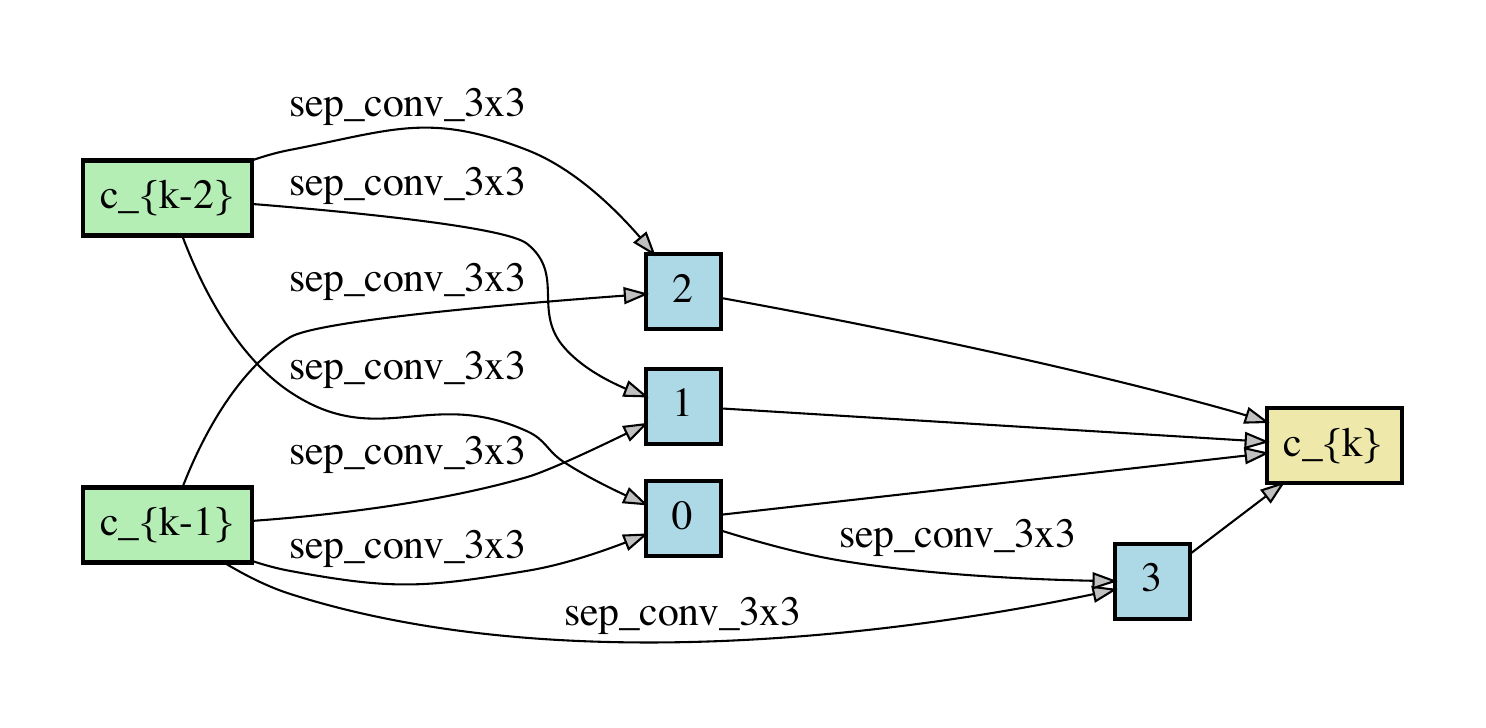}
		\caption{S2}
		\label{fig:rome-v2-cifar100-rdarts-ss-s2}
	\end{subfigure}
	
	\begin{subfigure}{0.98\columnwidth}
		\includegraphics[width=0.45\columnwidth]{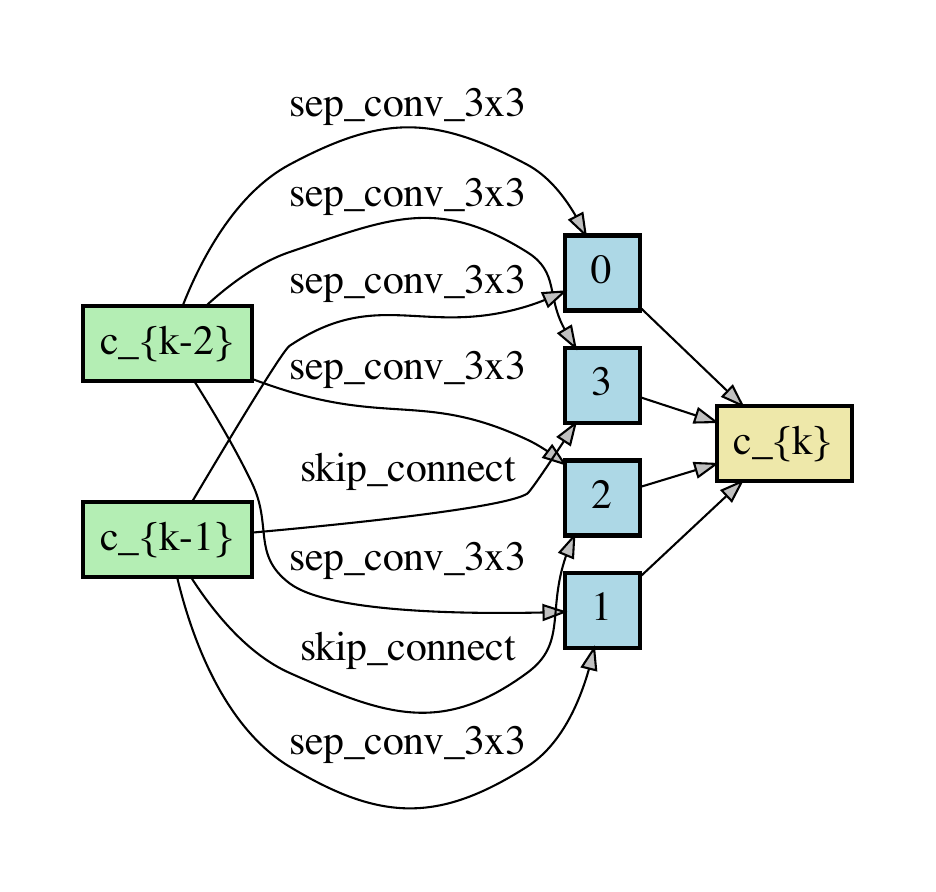}
		\includegraphics[width=0.45\columnwidth]{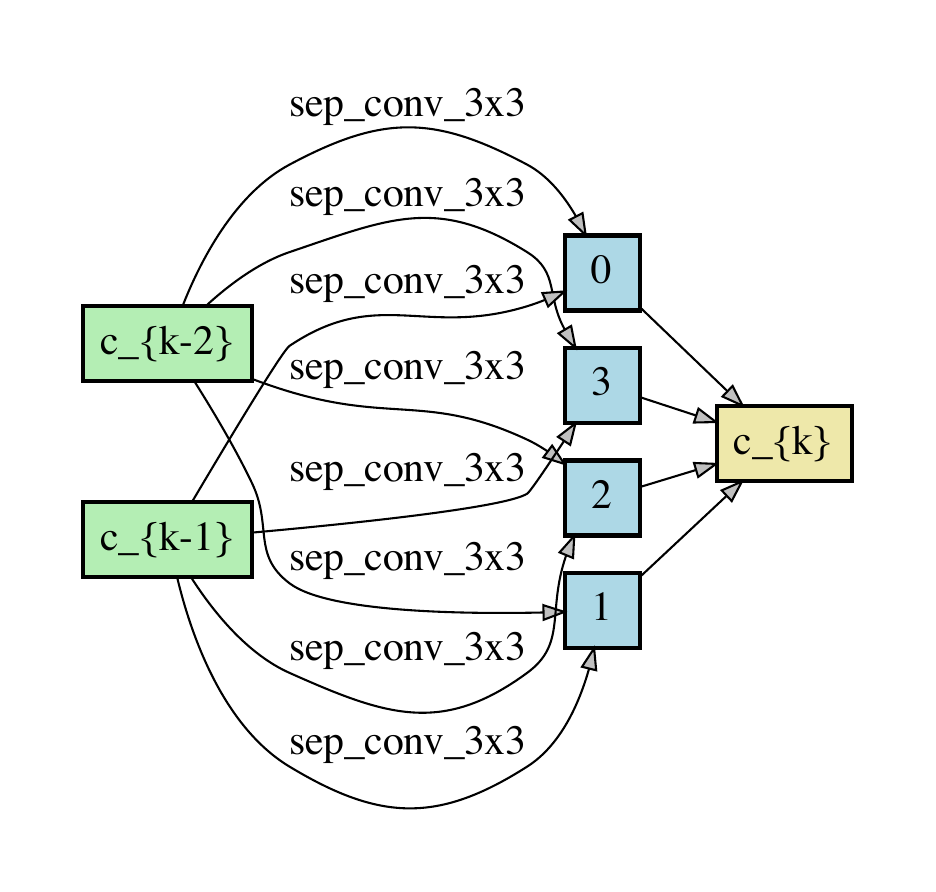}
		\caption{S3}
		\label{fig:rome-v2-cifar100-rdarts-ss-s3}
	\end{subfigure}
	
	\begin{subfigure}{0.98\columnwidth}
		\includegraphics[width=0.45\columnwidth]{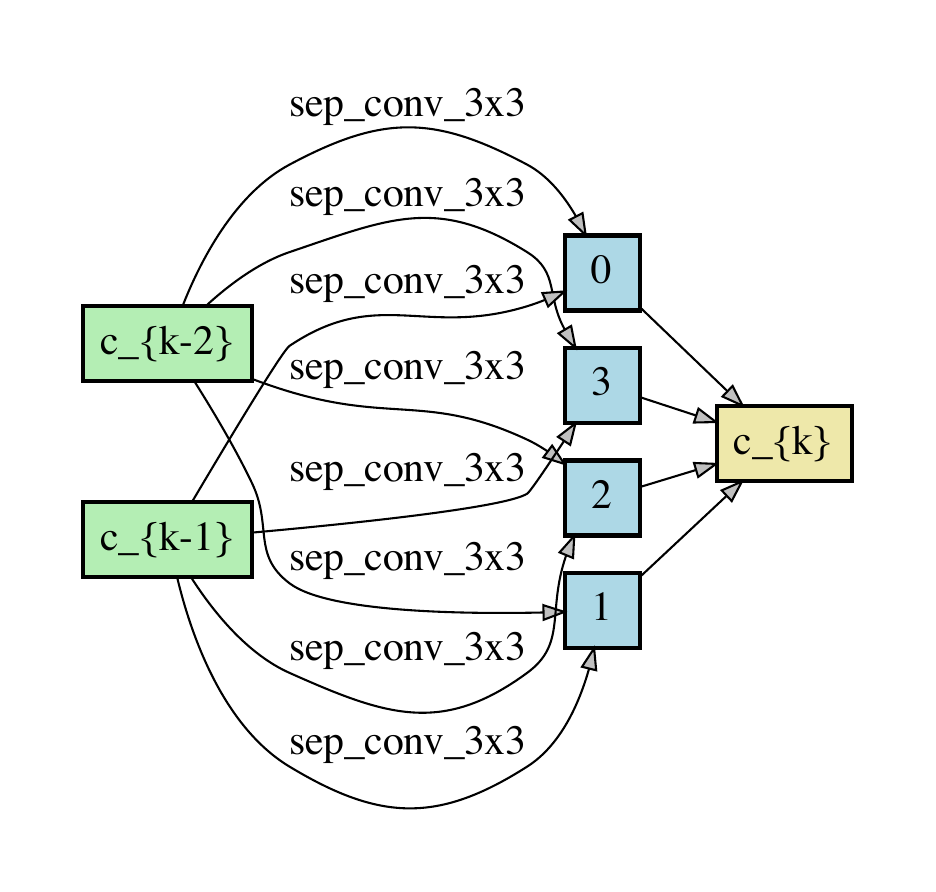}
		\includegraphics[width=0.45\columnwidth]{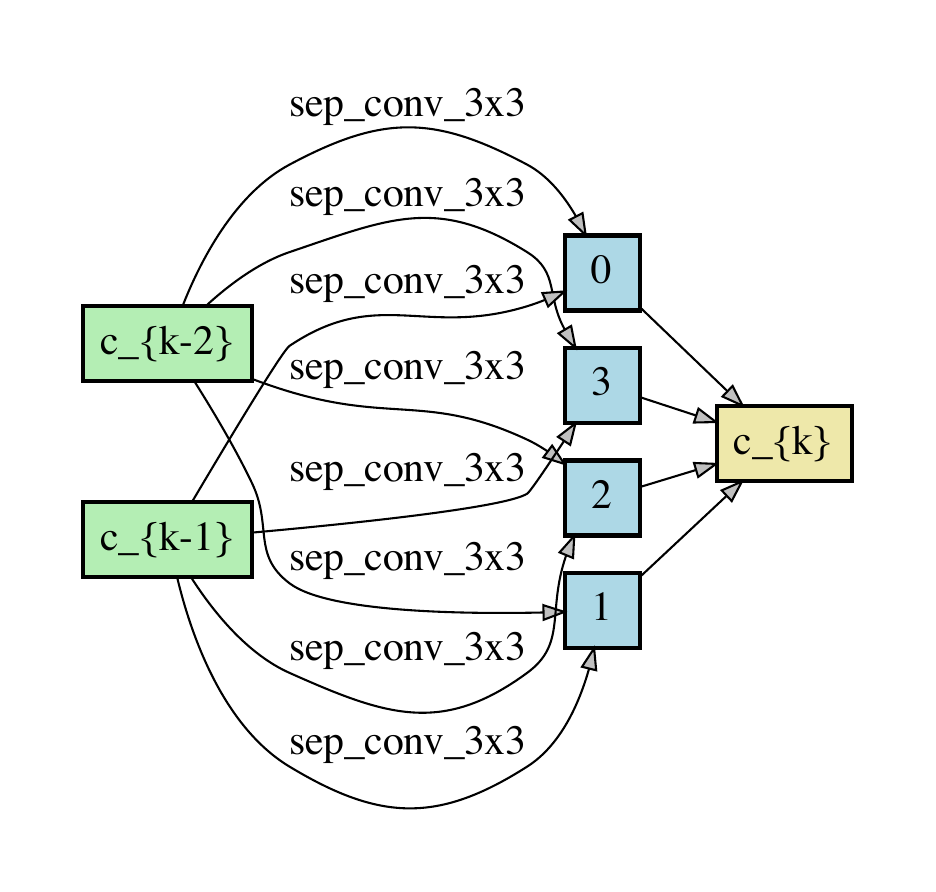}
		\caption{S4}
		\label{fig:rome-v2-cifar100-rdarts-ss-s4}
	\end{subfigure}
	
	\caption{ROME-V2 best cells (paired in normal and reduction) on CIFAR100 in reduced search spaces of RobustDARTS.}
	\label{fig:rome-v2-cifar100-rdarts-ss}
\end{figure}

\begin{figure}[ht]
	\centering
	\begin{subfigure}{0.98\columnwidth}
		\includegraphics[width=0.45\columnwidth]{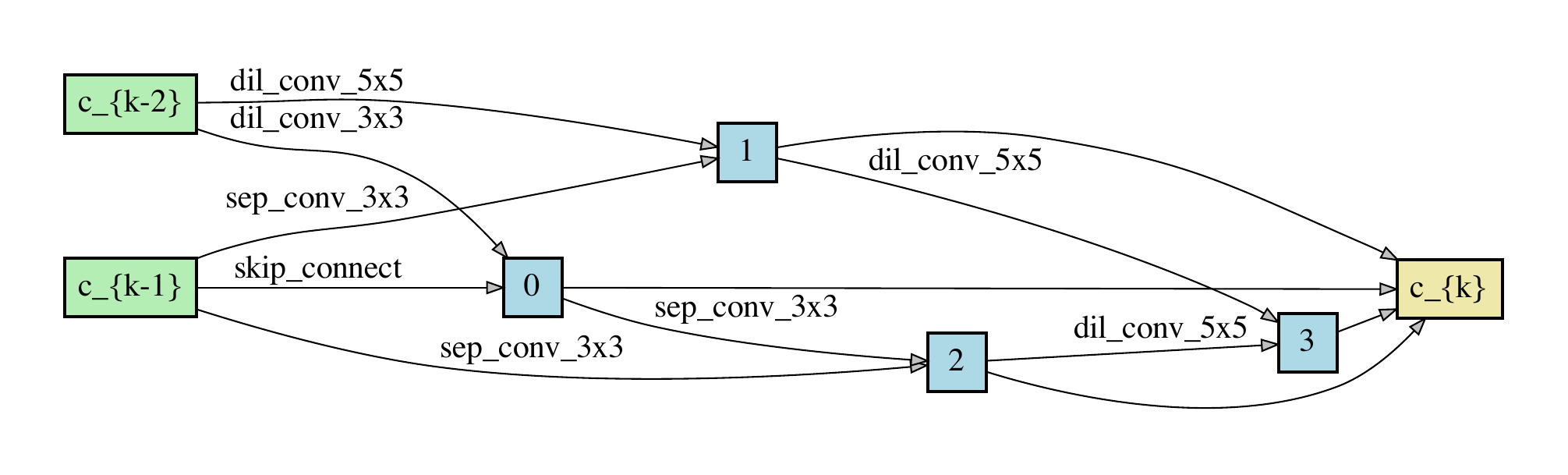}
		\includegraphics[width=0.45\columnwidth]{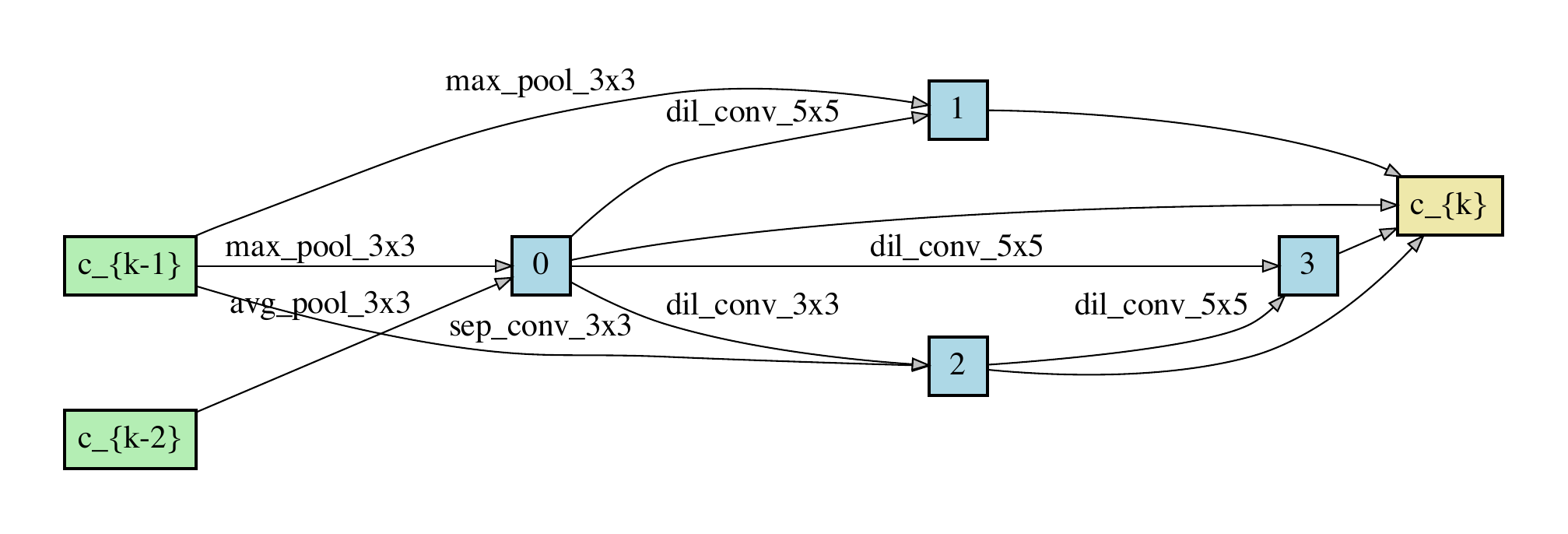}
		\caption{S1}
		\label{fig:rome-v2-svhn-rdarts-ss-s1}
	\end{subfigure}
	
	\begin{subfigure}{0.98\columnwidth}
		\includegraphics[width=0.45\columnwidth]{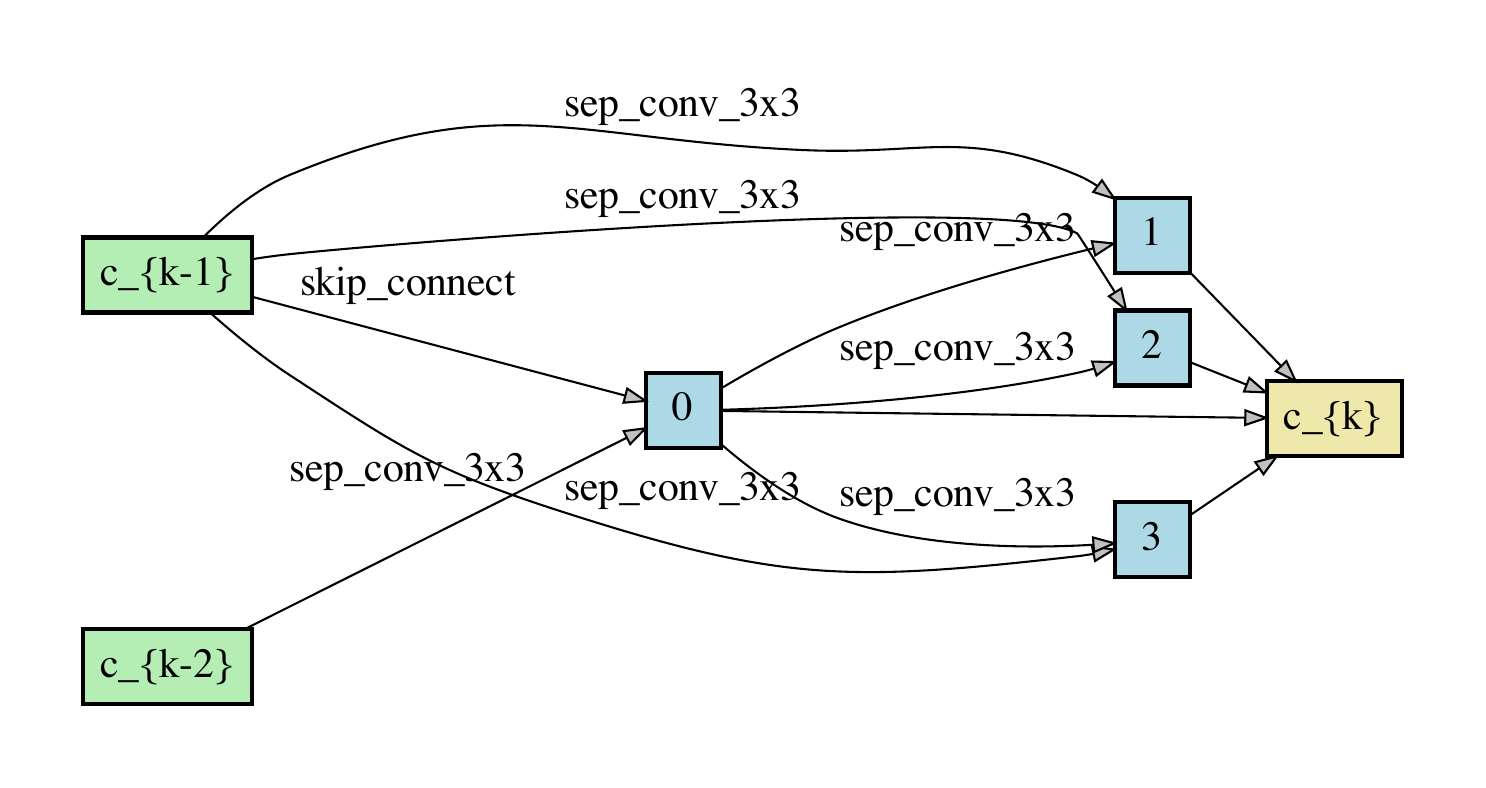}
		\includegraphics[width=0.45\columnwidth]{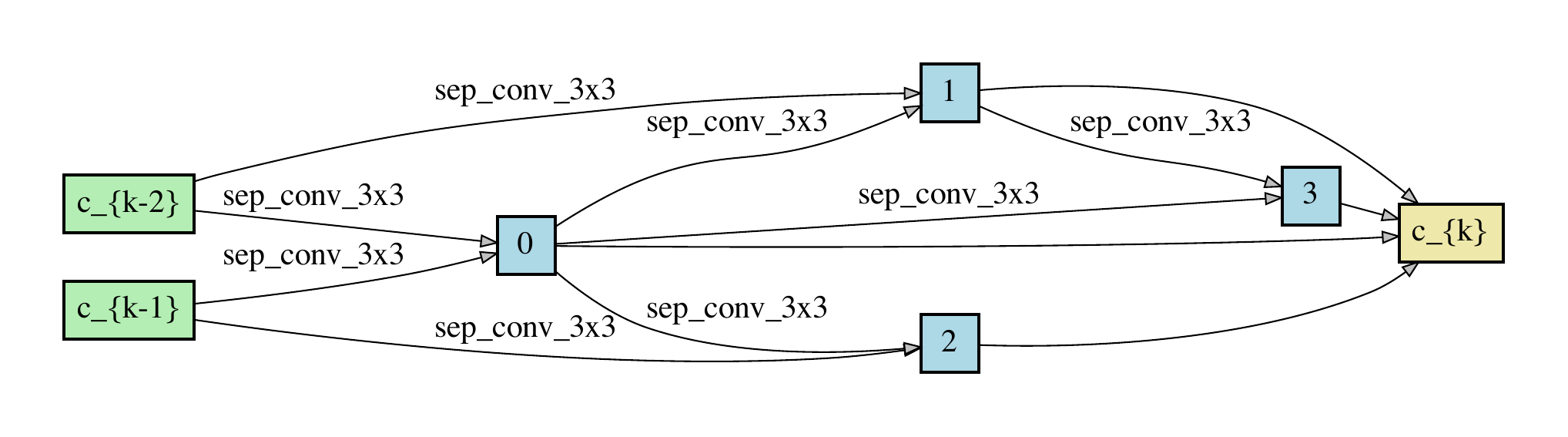}
		\caption{S2}
		\label{fig:rome-v2-svhn-rdarts-ss-s2}
	\end{subfigure}
	
	\begin{subfigure}{0.98\columnwidth}
		\includegraphics[width=0.45\columnwidth]{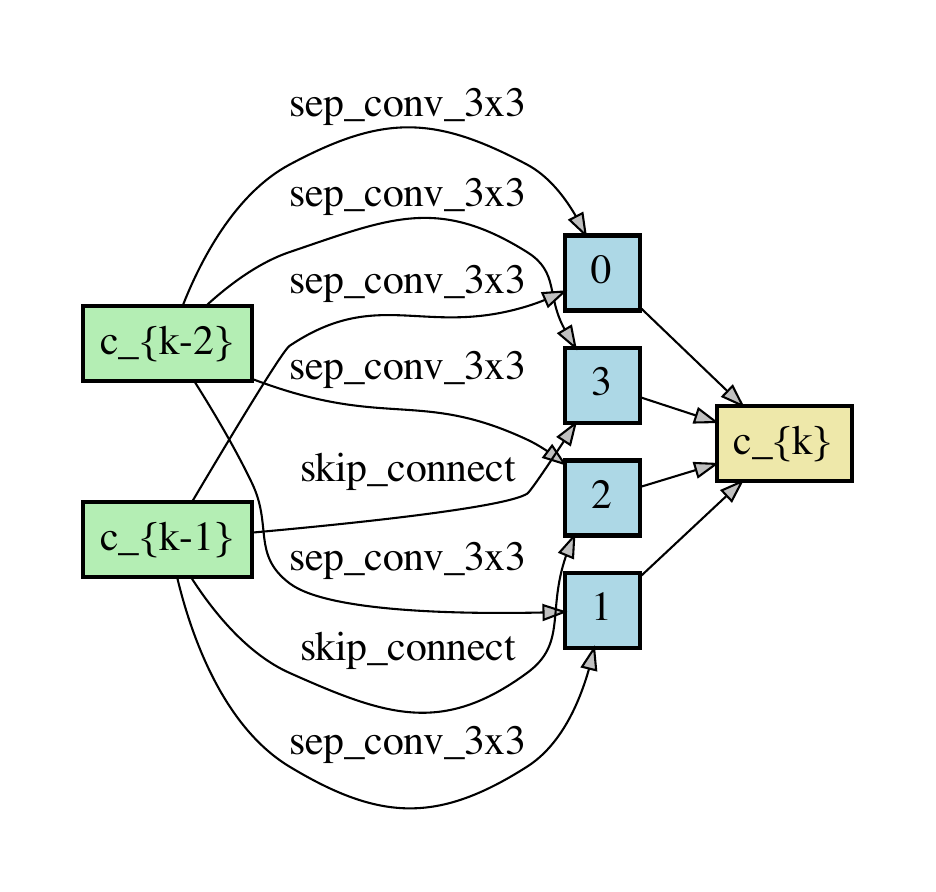}
		\includegraphics[width=0.45\columnwidth]{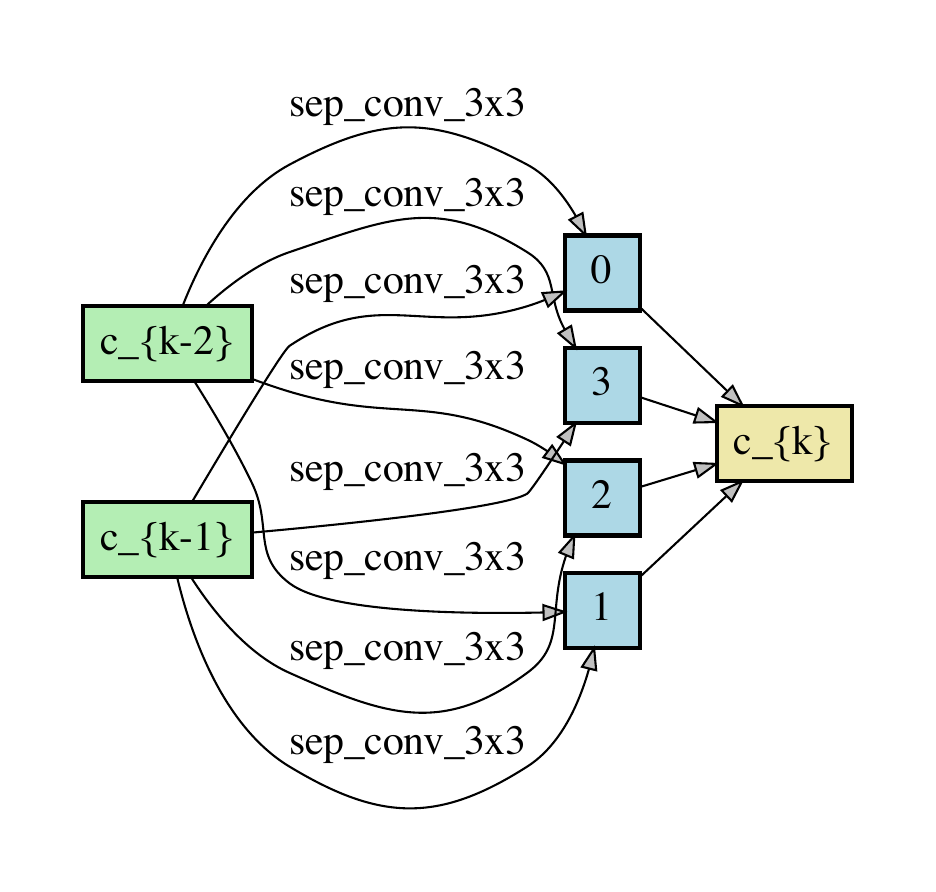}
		\caption{S3}
		\label{fig:rome-v2-svhn-rdarts-ss-s3}
	\end{subfigure}
	
	\begin{subfigure}{0.98\columnwidth}
		\includegraphics[width=0.45\columnwidth]{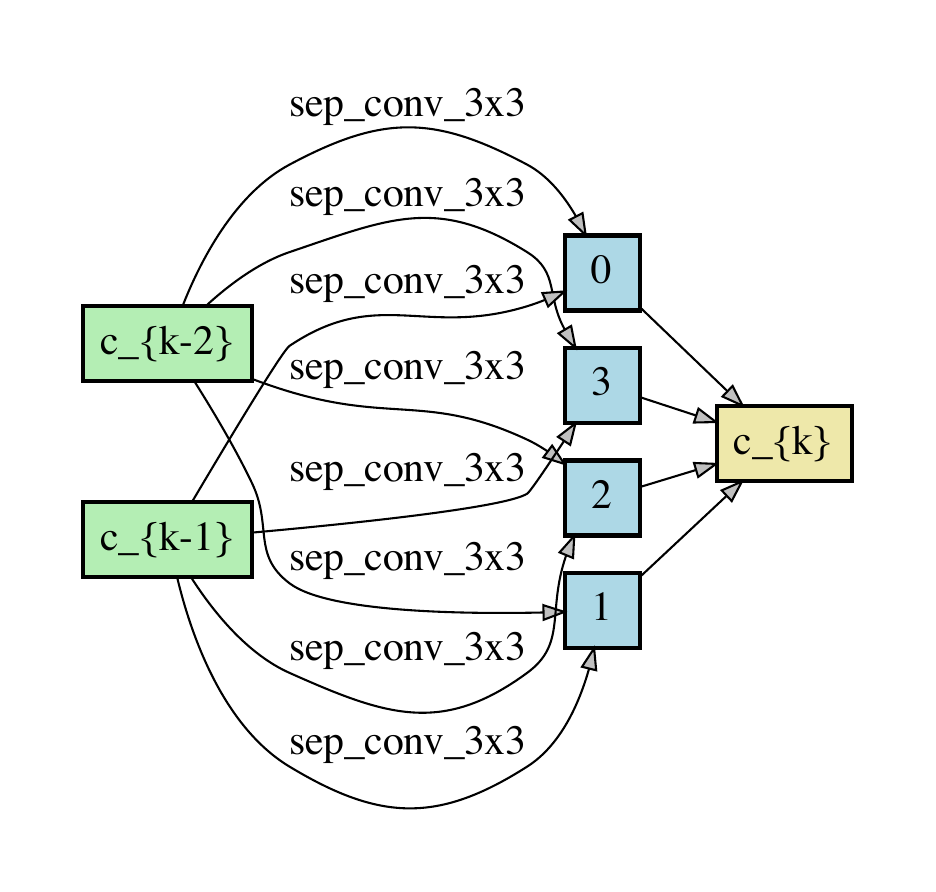}
		\includegraphics[width=0.45\columnwidth]{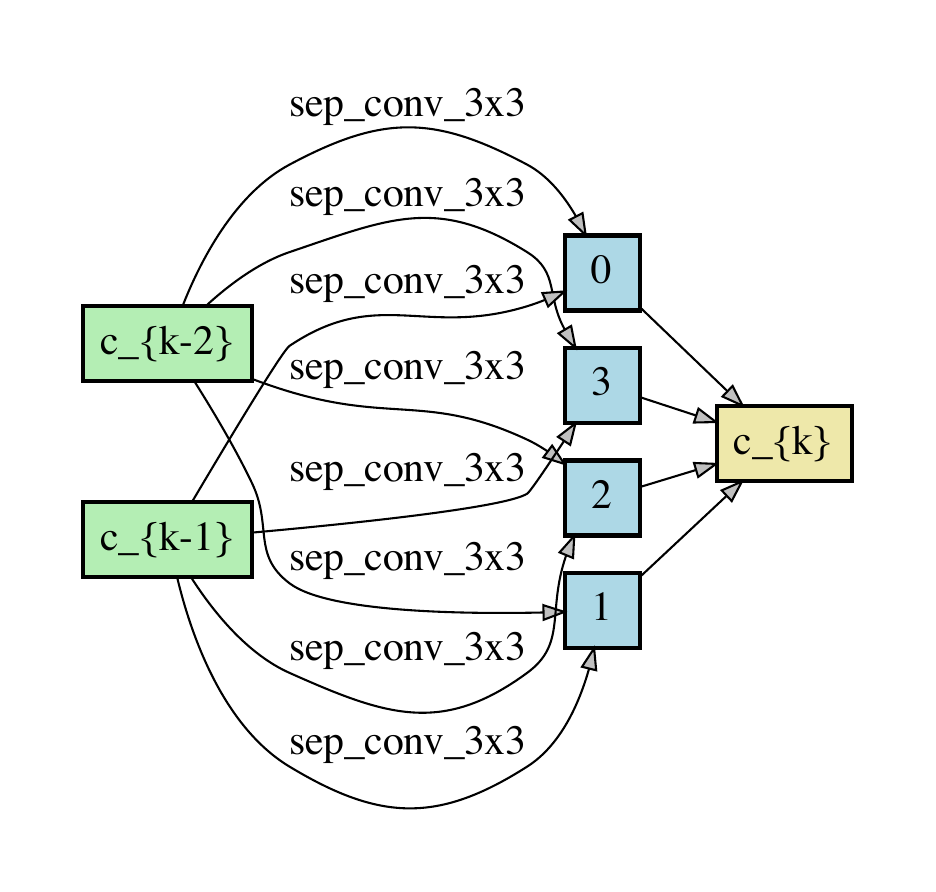}
		\caption{S4}
		\label{fig:rome-v2-svhn-rdarts-ss-s4}
	\end{subfigure}
	
	\caption{ROME-V2 best cells (paired in normal and reduction) on SVHN in reduced search spaces of RobustDARTS.}
	\label{fig:rome-v2-svhn-rdarts-ss}
\end{figure}

\begin{figure}[ht]
	\centering
	\begin{subfigure}{0.98\columnwidth}
		\includegraphics[width=0.9\columnwidth]{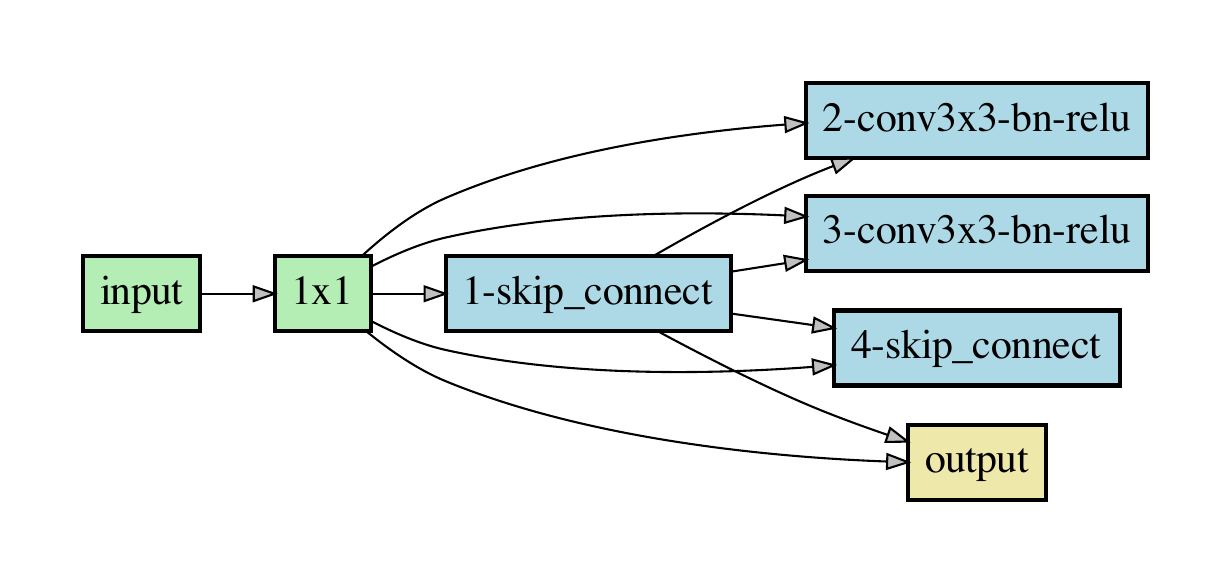}
		\caption{S1}
		\label{fig:1shot1-gdas-fail-rest-s1}
	\end{subfigure}
	
	\begin{subfigure}{0.98\columnwidth}
		\includegraphics[width=0.9\columnwidth]{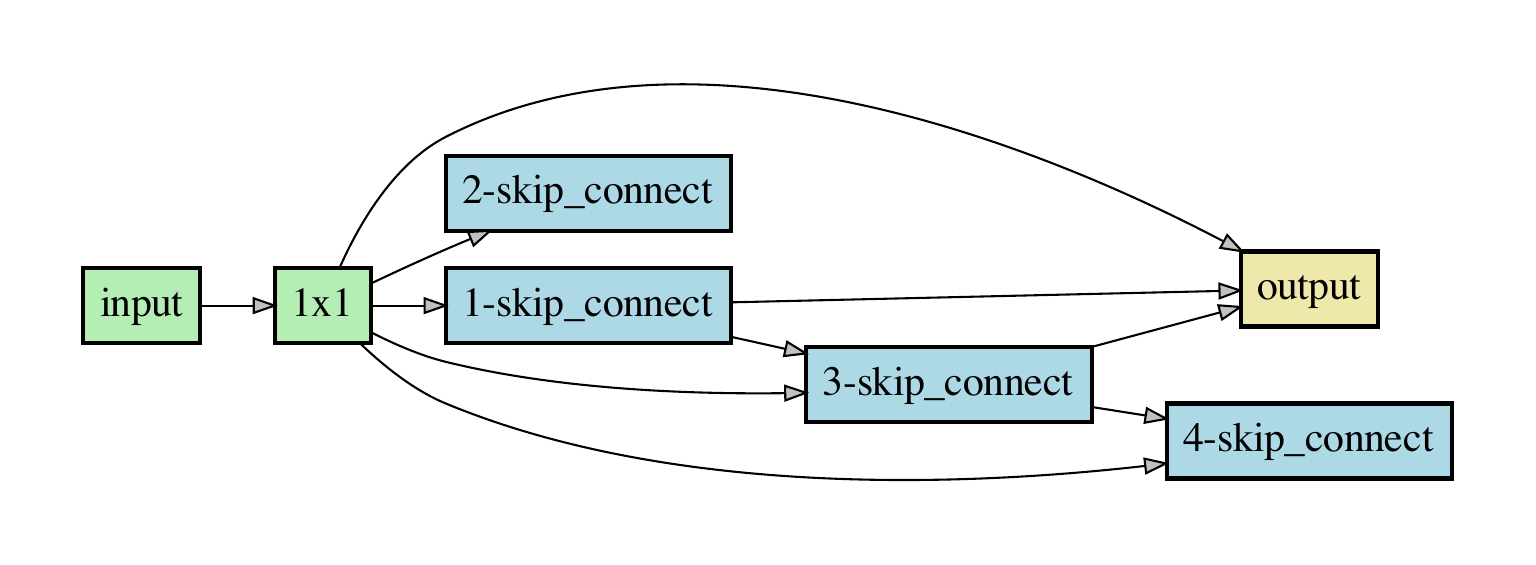}
		\caption{S2}
		\label{fig:1shot1-gdas-fail-rest-s2}
	\end{subfigure}
	
	\begin{subfigure}{0.98\columnwidth}
		\includegraphics[width=0.9\columnwidth]{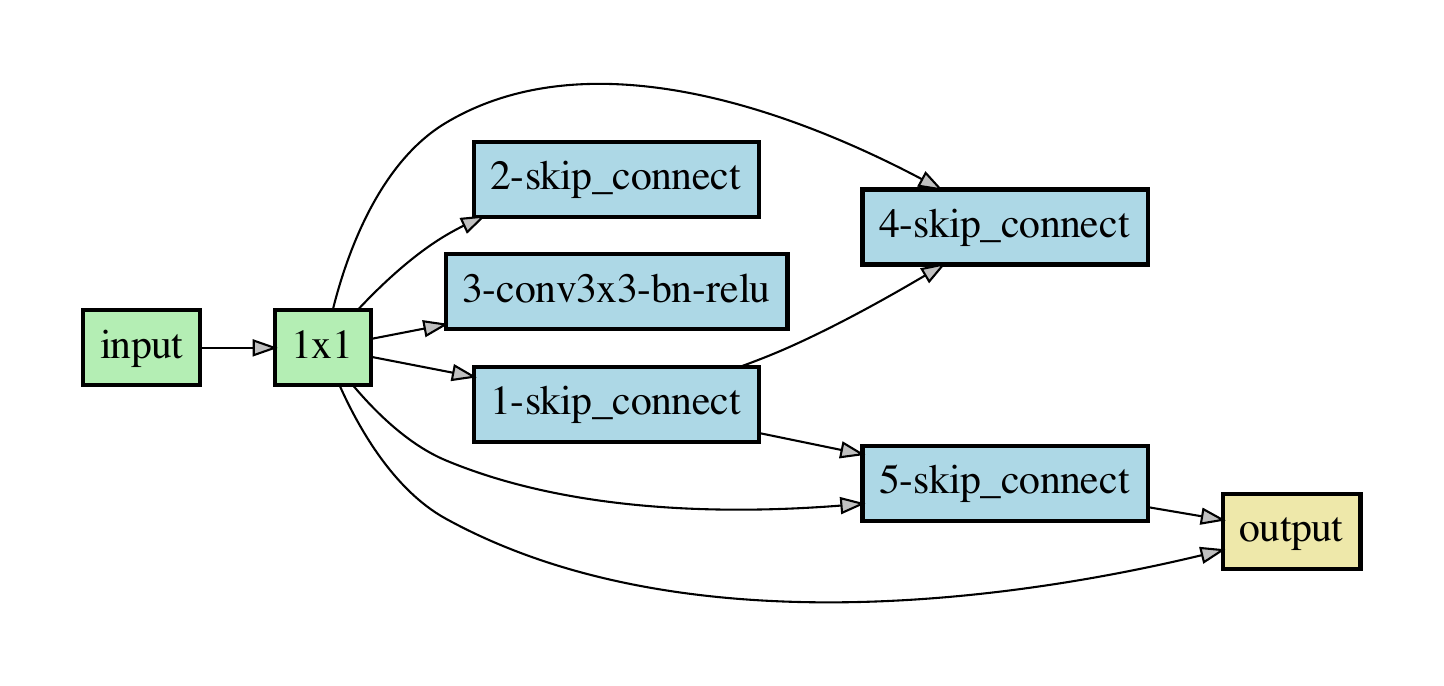}
		\caption{S3}
		\label{fig:1shot1-gdas-fail-rest-s3}
	\end{subfigure}
	
	\caption{GDAS fails on NAS-Bench-1Shot1 \cite{zela2020nas} when searching on CIFAR-10 in all three search spaces when skip connection are added into choices. In each MixedOp, we have three choices: \{maxpool3x3, conv3x3-bn-relu, skip-connect\}.}
	\label{fig:1shot1-gdas-fail-rest}
\end{figure}

\begin{figure}[ht]
	\centering
	\begin{subfigure}{0.98\columnwidth}
		\includegraphics[width=0.9\columnwidth]{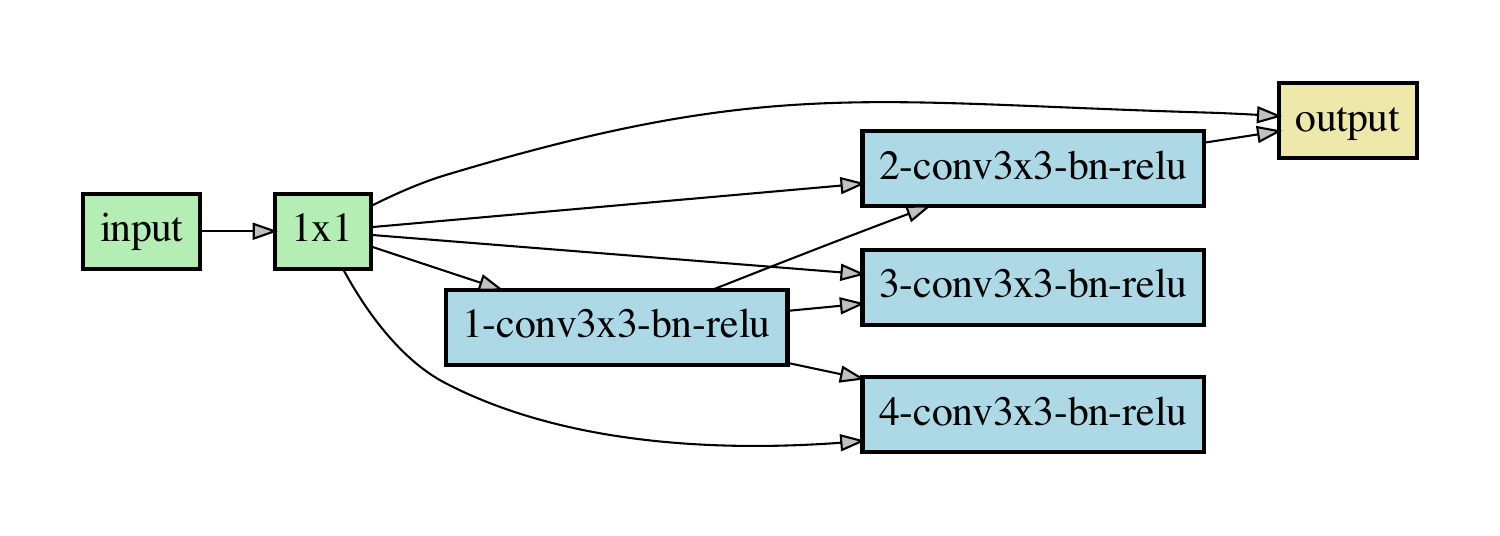}
		\caption{S1}
		\label{fig:1shot1-rome-best-s1}
	\end{subfigure}
	
	\begin{subfigure}{0.98\columnwidth}
		\includegraphics[width=0.9\columnwidth]{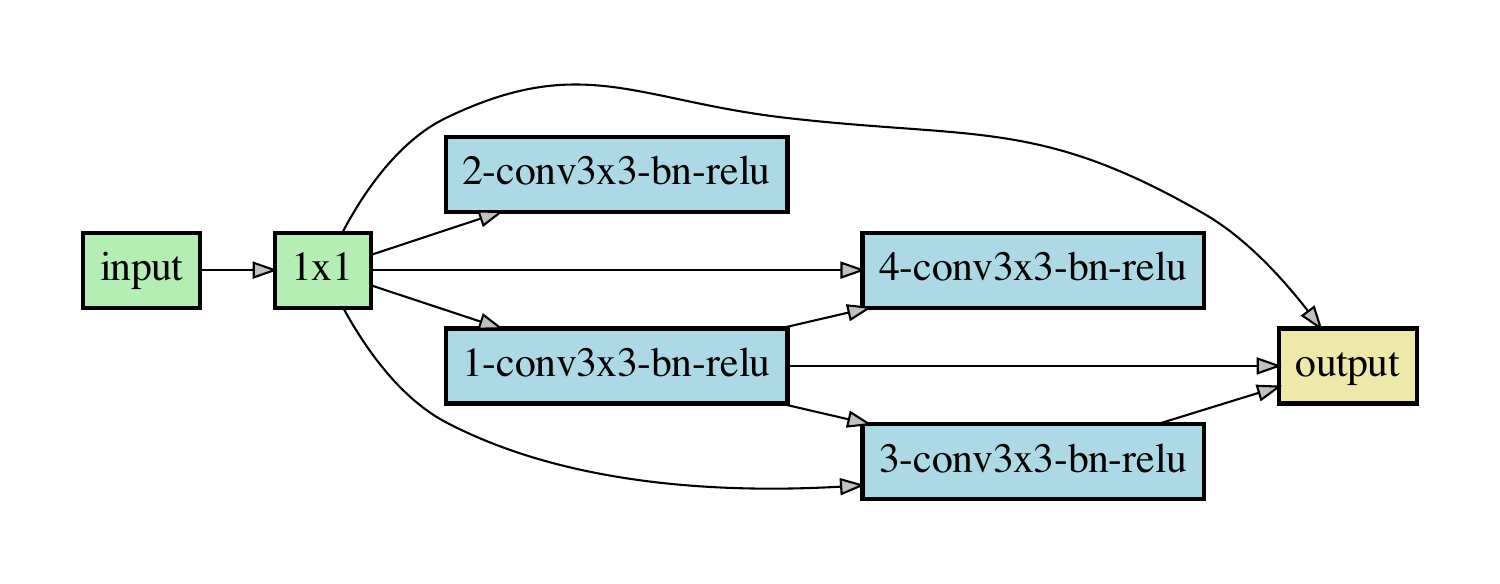}
		\caption{S2}
		\label{fig:1shot1-rome-best-s2}
	\end{subfigure}
	
	\begin{subfigure}{0.98\columnwidth}
		\includegraphics[width=0.9\columnwidth]{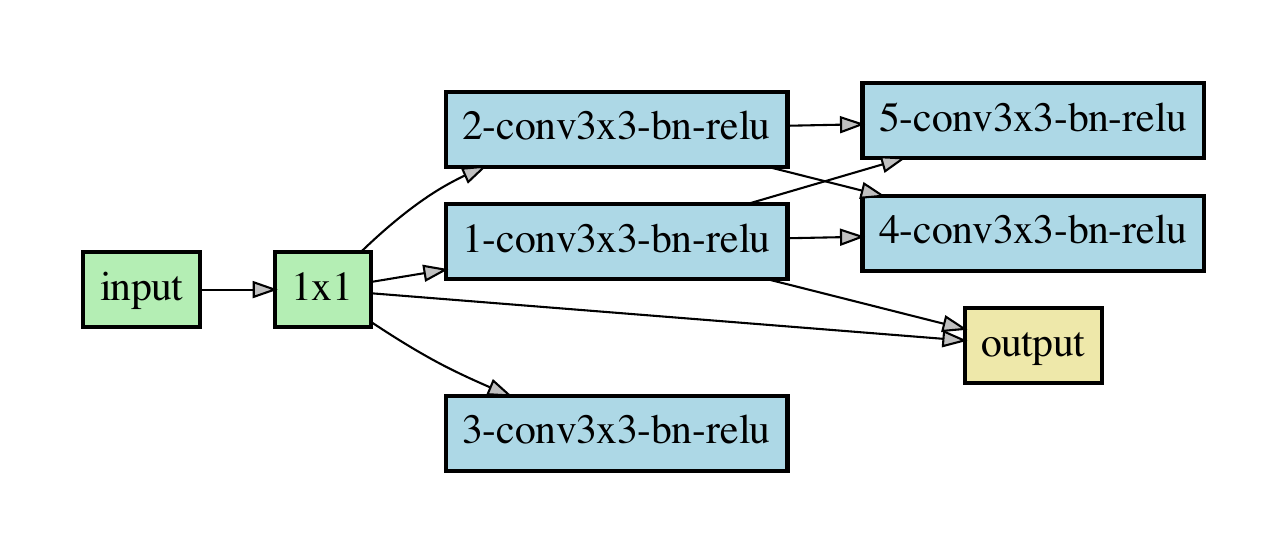}
		\caption{S3}
		\label{fig:1shot1-rome-best-s3}
	\end{subfigure}
	
	\caption{ROME-V2 resolves the aggregation of skip connections  on NAS-Bench-1Shot1 \cite{zela2020nas}. Notice intermediate nodes concatenate their outputs as the input for the output node, while some have loose ends and don't feed to the output node.}
	\label{fig:1shot1-rome-best}
\end{figure}

Genotypes of the discovered architectures by ROME are illustrated in Fig.~\ref{fig:rome-v1-cifar10-best-geno} - Fig.~\ref{fig:1shot1-rome-best}.

\end{document}